\documentclass{article}
\usepackage{preprint,times}
\finalcopy

\usepackage{amsmath,amsfonts,bm}

\def\eqref#1{equation~\ref{#1}}

\def\1{\bm{1}}

\DeclareMathAlphabet{\mathsfit}{\encodingdefault}{\sfdefault}{m}{sl}
\SetMathAlphabet{\mathsfit}{bold}{\encodingdefault}{\sfdefault}{bx}{n}

\DeclareMathOperator*{\argmax}{arg\,max}
\DeclareMathOperator*{\argmin}{arg\,min}

\usepackage{url}

\usepackage[utf8]{inputenc}
\usepackage[T1]{fontenc}
\usepackage[hidelinks]{hyperref}
\usepackage{booktabs}
\usepackage{amsfonts}
\usepackage{nicefrac}
\usepackage{microtype}
\usepackage{xcolor}

\usepackage[utf8]{inputenc}
\usepackage[small]{caption}
\usepackage{graphicx}
\usepackage{amsmath}
\usepackage{amsthm}
\usepackage{paralist}
\usepackage{wrapfig}

\usepackage{natbib}

\usepackage[linesnumbered,ruled,vlined]{algorithm2e}

\usepackage{algpseudocode}
\algrenewcommand{\algorithmiccomment}[1]{\hfill// #1}

\usepackage{subcaption}
\usepackage{float}
\usepackage{needspace}

\title{Dynamic Hyperparameter Importance for Efficient Multi-Objective Optimization}

\author{%
  Daphne Theodorakopoulos$^{1,2}$ \quad Marcel Wever$^{1,2}$ \quad Marius Lindauer$^{1,2}$\\
  $^1$Institute of Artificial Intelligence (LUH$|$AI), Leibniz University Hannover\\
  $^2$L3S Research Center\\
  \texttt{\{d.theodorakopoulos, m.wever, m.lindauer\}@uni-hannover.de}
}

\usepackage[inline]{enumitem}
\usepackage{xspace}
\newcommand{\pymoo}{PyMOO\xspace}
\newcommand{\yahpo}{YAHPO-Gym\xspace}

\newcommand{\carps}{CARP-S benchmark suite\xspace}

\begin{document}

\maketitle

\begin{abstract}
Choosing a suitable ML model is a complex task that can depend on several objectives, e.g., accuracy, fairness, or energy consumption.
In practice, this requires trading off multiple, often competing, objectives through multi-objective optimization (MOO).
However, existing MOO methods typically treat all hyperparameters as equally important, disregarding that hyperparameter importance (HPI) can vary significantly across objectives.
We propose a novel dynamic optimization approach that prioritizes the most influential hyperparameters based on varying objective trade-offs during the search, thereby accelerating empirical convergence. 
We advance prior work on HPI for MOO from post-analysis to direct, dynamic integration within the optimization, using HyperSHAP, a recent HPI method that attributes performance gains to hyperparameters via Shapley values. For this, we leverage the objective weightings naturally produced by the MOO algorithm ParEGO and reduce the configuration space by fixing the unimportant hyperparameters, allowing the search to focus on the important ones.
We evaluate our method on synthetic tasks from \pymoo and moderate-dimensional HPO tasks from \yahpo.
For HPO, integrating HPI yields 62\% improvement in final Pareto front quality, while on synthetic data, the improvement is 71\% over standard ParEGO.\footnote{The code is available at \url{https://github.com/LUH-AI/hpi_parego}.}
\end{abstract}

\section{Introduction}

Hyperparameter optimization (HPO) is a critical step in maximizing the performance of machine learning models~\citep{bergstra-jmlr12a,snoek-nips12a,levesque-uai16a,feurer-automlbook19a,bischl-dmkd23a}. Traditionally, HPO has been conducted with a single performance objective in mind, often predictive accuracy. However, real-world applications rarely depend solely on accuracy. Instead, they must balance multiple, often conflicting requirements. For example, large-scale deployments demand low inference latency, while embedded and edge devices require strict limits on memory and energy usage. In socially sensitive applications, fairness constraints may be mandated by regulation~\citep{schmucker-metalearn20a,weerts-jair24a}. This motivates the need to frame HPO as a multi-objective optimization (MOO) problem, where the goal is not a single best configuration but an approximation of the Pareto front, a set of optimal trade-offs across objectives. From that, a developer can make an informed decision about the final model to deploy. We focus on the common setting of HPO for Deep Learning with moderate-dimensional configuration spaces, where fewer trials can save substantial compute due to the high cost of model training.

\begin{figure}[t]
  \centering
  \includegraphics[width=0.8\columnwidth]{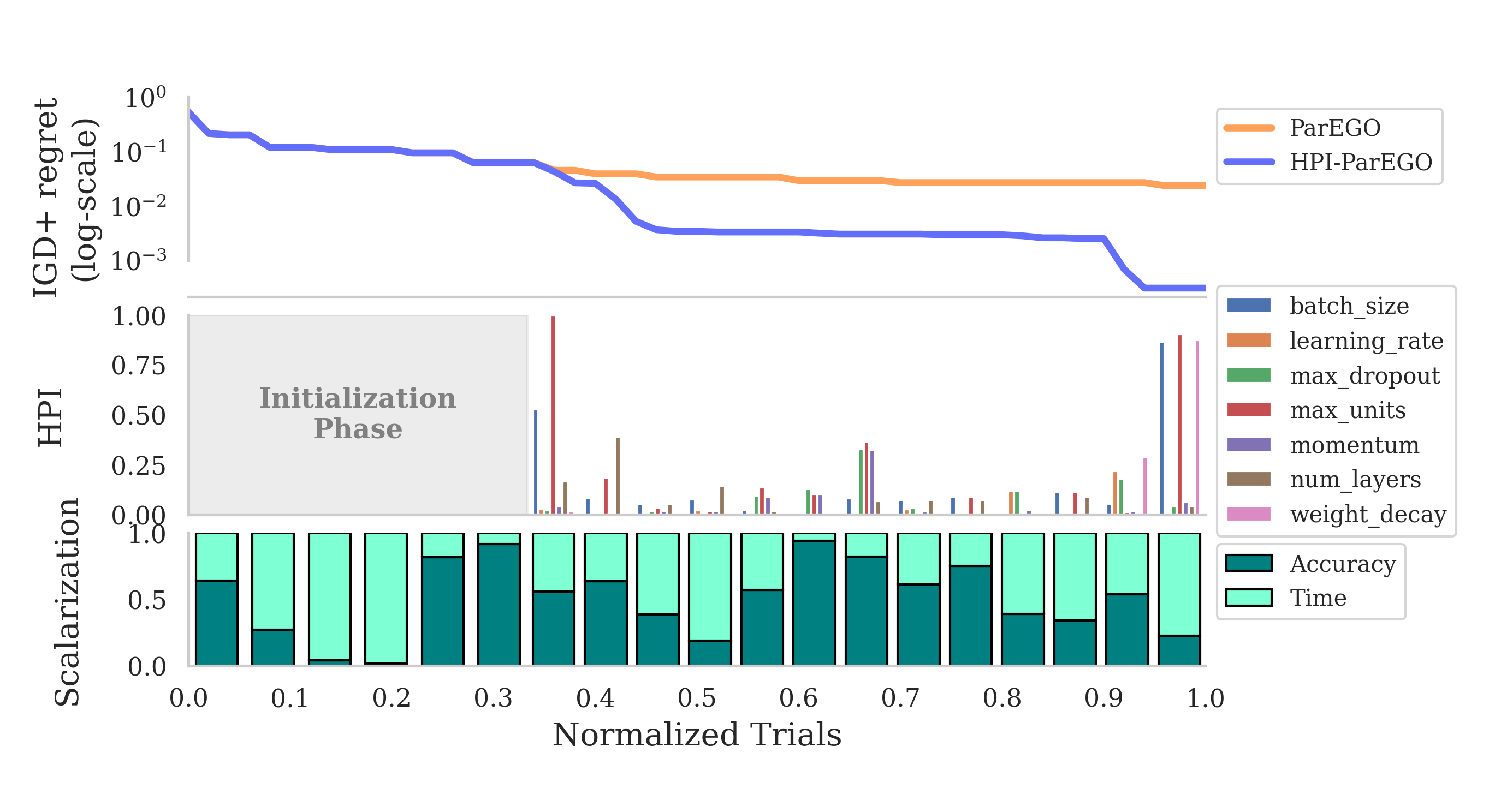}
  \caption{HPO task \texttt{lcbench\_126026}: accuracy–time trade-off. Top: ParEGO vs. our variant with dynamic HPI on IGD+ regret over trials. Middle: HPI evolution across scalarizations; incl. initialization phase with no reduction of the configuration space. Bottom: Scalarizations sampled by ParEGO.}
  \label{fig:teaser}
\end{figure}

Compared to single-objective HPO, multi-objective variants are considerably more complex and computationally demanding, making efficiency even more critical. While existing MOO methods~\citep{elsken-iclr19a,zhao-iclr21a,moraleshernandez-air22a} assume uniform Hyperparameter Importance (HPI), single-objective studies show that HPI varies by task~\citep{bergstra-jmlr12a,hutter-icml14a,rijn-kdd18a,wever-aaai26a}. Leveraging this variability in HPI can improve optimization efficiency~\citep{hutter-icml14a,wang-pmlr25a,wever-aaai26a}.

Recent work shows that HPI varies across different objective trade-offs in multi-objective settings~\citep{theodorakopoulos-ecai24a}. We hypothesize that optimizers with dynamic scalarization, such as ParEGO~\citep{knowls-evoco06a}, can exploit this by adapting their search toward the most relevant hyperparameters. Our approach achieves this by dynamically reducing the configuration space for the current scalarization.
Reducing the configuration space to the effective dimensionality enables the surrogate to learn more efficiently and accurately from limited data~\citep{wang-jair16a,eriksson-neurips19a}, potentially also mitigating the curse of dimensionality~\citep{bellman-book57a}.
Figure~\ref{fig:teaser} displays the superior performance of our optimizer against the baseline in an illustrative example.

\paragraph{Contributions}
\begin{enumerate}
    \item
    We present the first efficient MOO method that dynamically incorporates HPI into the optimization loop of scalarization-based MOO approaches. HPI-ParEGO adaptively selects and focuses on the most influential hyperparameters based on the current objective scalarization.
    \item
    We provide insights into the performance of our method with synthetic and HPO tasks, focusing on moderate-dimensional settings from \pymoo~\citep{blank-ieeeaccess20a} and \yahpo~\citep{pfisterer-automl22a}. Our HPI-based hyperparameter reduction demonstrates superior performance in single-objective optimization and outperforms standard ParEGO on \pymoo and all our MOO baselines on \yahpo in convergence speed and Pareto front quality. 
    \item
    We demonstrate the impact of key design choices through an ablation study, which includes the amount of randomness, HPI thresholds, when to draw a new scalarization, configuration space reduction, including hyperparameter interactions, the HyperSHAP surrogate model, and the informativeness of HyperSHAP vs. random hyperparameter sets.
\end{enumerate}

\section{Background and Related Work}
\textbf{Multi-objective optimization (MOO).}
MOO simultaneously optimizes two or more potentially conflicting objectives. It produces a \textit{set} of Pareto-optimal solutions, instead of a single solution, where no objective can be improved without degrading another \citep{pareto-manual71a}. 
This set, known as the Pareto front, represents the trade-off surface among objectives: 
\begin{equation}
\begin{aligned}
& \min_{\boldsymbol{\lambda} \in \Lambda} \ \mathbf{f}(\boldsymbol{\lambda}) = \left(f_1(\boldsymbol{\lambda}), \dots, f_m(\boldsymbol{\lambda})\right) \\
& \text{subject to: } \neg \exists \boldsymbol{\lambda}' \in \Lambda \text{ such that } f_i(\boldsymbol{\lambda}') \leq f_i(\boldsymbol{\lambda}) \ \forall i, \\
& \text{ with } f_j(\boldsymbol{\lambda}') < f_j(\boldsymbol{\lambda}) \text{ for some } j
\end{aligned}
\label{eq:moo}
\end{equation}
where \( \boldsymbol{\lambda} \in \Lambda \) represents a candidate configuration from the configuration space \( \Lambda \) and \( f_j(\boldsymbol{\lambda}) \) denotes the \( j \)-th objective function to be minimized.

MOO is often tackled using either evolutionary algorithms or Bayesian optimization \citep{moraleshernandez-air22a}. A well-known evolutionary approach is NSGA-II~\citep{deb-tec02a}, which handles the objectives and maintains solution diversity through non-dominated sorting and crowding distance.
ParEGO \citep{knowls-evoco06a} is a prominent Bayesian method for HPO that facilitates standard surrogate-based optimization by converting MOO problems into a series of single-objective problems via random scalarization weight sampling:
\begin{equation}
\min_{\boldsymbol{\lambda} \in \Lambda} \left\{ \max_{j=1,\ldots,m} \left[ w_j \cdot f_j(\boldsymbol{\lambda}) \right] + \rho \sum_{j=1}^m w_j \cdot f_j(\boldsymbol{\lambda}) \right\}
\label{eq:parego}
\end{equation}
where \( \boldsymbol{w} = (w_1, \ldots, w_m) \) is a weight vector sampled randomly, with \( w_j \geq 0 \) and \( \sum_{j=1}^m w_j = 1 \), and \( \rho \) is a small positive scalar (e.g., 0.05) that encourages diversity.
While efficient, ParEGO and similar algorithms typically treat all hyperparameters equally, ignoring that some may have more impact on the current objective trade-off. Our work embeds dynamic HPI estimation into ParEGO to prioritize the most relevant hyperparameters during optimization.

\textbf{Hyperparameter Importance.}
In single-objective HPO, a variety of HPI methods exist. These are typically applied post-hoc, using the results of completed optimization runs. Surrogate models trained on such optimization data often model performance as a function of hyperparameter configurations. Notable techniques include fANOVA \citep{hutter-icml14a} assessing HPI by decomposing performance variance, as well as forward selection \citep{hutter-lion13a} and Local Parameter Importance \citep{biedenkapp-lion18a}.
Shapley values \citep{shapley-book53a} have also been applied to attribute HPI \citep{adachi-aistats24a,rodemann-arxiv24a,wever-aaai26a}. As a local approach, ablation path analysis~\citep{fawcett-heu16a,biedenkapp-aaai17a} measures hyperparameter contributions by comparing the default configuration to optimized ones. 
Building on these foundations, \citet{theodorakopoulos-ecai24a} extended HPI estimation to the multi-objective setting. Their approach scalarized objective values from the Pareto front and applied fANOVA and ablation analysis to retrospectively evaluate HPI. Our work advances this line of research by directly integrating HPI into the optimization, enabling dynamic and adaptive search guidance. Additionally, we use HyperSHAP as an HPI measure, as it quantifies performance gains.

\textbf{Dynamic Configuration Space Reductions.}
Several HPO methods dynamically reduce the configuration space to improve efficiency. 
\citet{wistuba-pkdd15a} prune unpromising regions using prior experiments or early evaluations, \citet{lee-iasc22a} shrink or expand the space as the budget evolves, and Optuna~\citep{akiba-kdd19a} allows users to adjust it during optimization.
SEBO~\citep{liu-aistats23a} and BONSAI~\citep{daulton-arxiv26a} promote sparsity by resetting low-impact hyperparameters to their defaults for individual configurations, but do not explicitly model HPI to reduce the configuration space globally.
ExperienceThinking~\citep{wang-kbs21a}, SAASBO~\citep{eriksson-uai21a}, and VS-BO~\citep{shen-automl23a} focus the search on influential hyperparameters, whereas GSOS~\citep{wang-pmlr25a} leverages HPI to group hyperparameters and optimize them in order of importance. 
However, these methods are designed for single-objective settings; they neither account for HPI varying across objective trade-offs nor explicitly quantify interactions.
PriMO~\citep{arxiv-basu25a} incorporates human priors per objective in the acquisition function of MOO, but it relies on user-defined beliefs, and their priors cover only the extremes of the Pareto front, whereas our approach automatically adapts the search using HPI and explores a broader region of the Pareto front.
These efforts reflect a growing interest in adaptive HPO techniques that tailor the configuration space to improve search efficiency. 
These methods either reduce the space without an explicit HPI estimate (i.e., user-defined priors) or estimate relevance for a single objective. In scalarization-based MOO, in contrast, the target itself changes with every new weight vector~\citep{theodorakopoulos-ecai24a}, so the relevant subspace must be re-identified for each scalarization. To the best of our knowledge, no prior work incorporates dynamic HPI estimation into multi-objective HPO.

\begin{figure*}[b]
  \centering
  \includegraphics[width=0.9\textwidth]{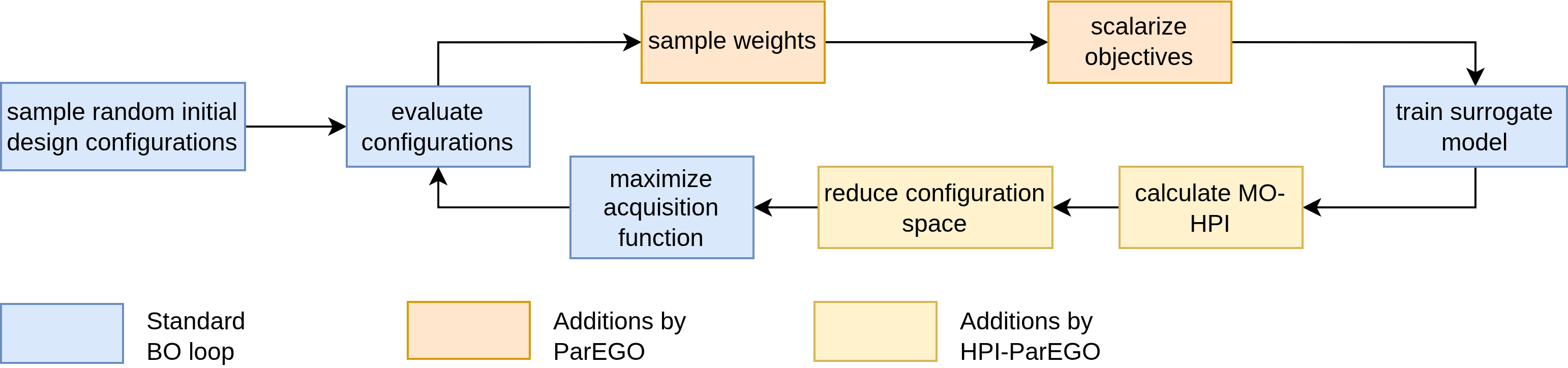}
  \caption{Algorithm overview: Blue boxes represent the standard Bayesian optimization (BO) loop, orange boxes additions made by ParEGO, and yellow boxes those introduced by our proposed HPI-ParEGO.}
  \label{fig:overview}
\end{figure*}

\section{Dynamic Hyperparameter Reduction}\label{sec:method}
We extend the ParEGO algorithm by incorporating HPI estimation into each optimization step. Our method dynamically identifies the most influential hyperparameters under the current objective scalarization and uses this information to guide the search (see Figure~\ref{fig:overview}). Section~\ref{sub:hpi_parego} outlines the approach; Section \ref{sec:calc_mo_hpi} details HPI integration and configuration space reduction; and Section~\ref{sec:threshold} presents the thresholding strategy for how aggressively the space is reduced.

\subsection{General Algorithm: HPI-ParEGO}\label{sub:hpi_parego}
Our method builds on ParEGO~\citep{knowls-evoco06a} (Equation~\ref{eq:parego}).\footnote{Our approach only assumes that the objectives are scalarized during the optimization, and thus is applicable to approaches similar to ParEGO~\citep{zhang-tec07a,bradford-jgo18a} with minor modifications.} By varying the weights across iterations, ParEGO gradually explores diverse tradeoffs along the Pareto front.
ParEGO (Algorithm~\ref{alg:hpi_parego}) starts from an initial design of sampled configurations (Line~1). At each iteration, scalarization weights $\boldsymbol{w}$ are drawn uniformly from the unit simplex (Lines~3-4), determining the emphasis on each objective. The evaluated configurations $\boldsymbol{\lambda}_i$ and their costs $f(\boldsymbol{\lambda}_i)$ in the optimization history $\mathcal{H}$ are then scalarized with $\boldsymbol{w}$, and a surrogate model $\hat{f}: \boldsymbol{\lambda} \mapsto f(\boldsymbol{\lambda}_i)$ is fitted to approximate performance under the current scalarization (Line~5).

\begin{algorithm}[t]
\small
\caption{HPI-ParEGO (\textcolor{blue}{Blue highlighting the additional parts of our approach})}
\label{alg:hpi_parego}
\DontPrintSemicolon
\KwIn{Objective functions $f(\cdot) = (f_1(\cdot), \ldots , f_m(\cdot))$, configuration space $\Lambda$, acquisition function $\alpha$, budget $B$, initial size $N_{\text{init}}$, initial weights $\boldsymbol{w} \sim \Delta^{m-1}$, threshold $\tau$, random chance $r$, weight update every $u$ iterations}
\KwOut{Approximated Pareto Set}

Initialize history $\mathcal{H} = \{ (\boldsymbol{\lambda}_i, f(\boldsymbol{\lambda}_i)) \}_{i=1}^{N_{\text{init}}}$ with an initial design (e.g., random search)\;

\For{$i = N_{\text{init}}+1$ \KwTo $B$}{
    \uIf{\textcolor{blue}{$i \bmod u = 0$}}{
        Update scalarization weights $\boldsymbol{w} \sim \Delta^{m-1}$ \tcp*{For the original ParEGO $u=1$}
    }
    Train surrogate model $\hat{f}_{\boldsymbol{w}}$ on $\mathcal{H}$ using the scalarized objective value of ParEGO (see Equation \ref{eq:parego})\;
    \uIf{\textcolor{blue}{$1- r$ chance}}{
        \textcolor{blue}{Compute HPI $h(\Lambda^j)$ for each hyperparameter $\Lambda^j$ \tcp*{See Section~\ref{sec:calc_mo_hpi}}}
        \textcolor{blue}{Select important hyperparameters (HPs) by HPI threshold $\tau$ based on $h(\Lambda^j)$\;}
        \textcolor{blue}{Reduce configuration space $\Lambda' \subset \Lambda$ by fixing unimportant HPs}\;
        Select configurations: $\boldsymbol{\lambda}_i \in \argmax_{\lambda \in \Lambda'} \alpha(\boldsymbol{\lambda} \mid \mathcal{H}, \hat{f}_{\boldsymbol{w}})$\;
    }
    \Else{
        \textcolor{blue}{Select random configuration $\boldsymbol{\lambda}_i$ from original configuration space $\boldsymbol{\lambda}_i \in \Lambda$}
    }
    Evaluate $f(\boldsymbol{\lambda}_i)$ and add result to history: $\mathcal{H} \leftarrow \mathcal{H} \cup \{ (\boldsymbol{\lambda}_i, f(\boldsymbol{\lambda}_i)) \}$\;
    \textcolor{blue}{Update threshold $\tau$ \tcp*{See Section~\ref{sec:threshold}}}
    
}
\Return Pareto Set based on $\mathcal{H}$\;
\end{algorithm}

We extend the ParEGO-loop by incorporating HPI (Lines 7-9). Using a surrogate model $\hat{g}$, we estimate how important each hyperparameter $\Lambda^{j}$ is for the current scalarized objective, yielding an importance score $h(\Lambda^{j})$. We then select the smallest subset of hyperparameters that collectively account for at least a fraction $\tau$ (Line 8) of the total performance gain, retaining only the most influential dimensions $\Lambda^\prime \subset \Lambda$ (Line~9) and concentrating the optimization budget where changes are most likely to improve performance. The remaining hyperparameters are fixed to a constant value (i.e., the current incumbent configuration in our implementation\footnote{Concretely, we define the incumbent as the observed configuration in $\mathcal{H}$ minimizing the current scalarized objective. Using the incumbent as the reference point has two advantages: (i) HyperSHAP quantifies the improvement over the incumbent and thus pushes the optimizer to even better configurations; and (ii) the reference point may change over time, and thus it mitigates the risk of a local optimum.
}).
Because both the importance estimates and the scalarization weights jointly determine the reduced space, changing the weights too frequently would discard useful structure before it can be fully exploited. We therefore update $\boldsymbol{w}$ only every $u$ iterations (Line~3), giving the optimizer time to capitalize on a stable reduced space. 
Note that HPI is re-estimated at every iteration as the surrogate improves with new evaluations.

Next, the acquisition function $\alpha$, balancing exploration and exploitation, e.g., expected improvement~\citep{jones-jgo98a} or upper confidence bound \citep{srinivas-icml10a}, is optimized over $\Lambda'$ to propose new candidates (Line~10), which are evaluated and added to the history $\mathcal{H}$ (Line~13). The threshold $\tau$ may then be updated based on the optimization's progress (Line 14).

Following established practice in Bayesian optimization (e.g., \citep{falkner-icml18a,lindauer-jmlr22a}), we interleave,
with a random chance $r$, a random configuration drawn from the full space $\Lambda$ (Line~12). 
This guards against premature convergence to early HPI estimates and ensures that hyperparameters deemed unimportant under a scalarization can still be explored under future tradeoffs.

\subsection{Calculating MO-HPI and Reducing the Configuration Space} \label{sec:calc_mo_hpi}
We dynamically estimate HPI during optimization to focus on the most promising dimensions. Since HPI is re-evaluated on every iteration, the choice of HPI method is critical.
We adopt HyperSHAP~\citep{wever-aaai26a}, which frames hyperparameter optimization as a cooperative game~\citep{fudenberg-mit91a} and applies Shapley values to quantify tunability \citep{probst-jmlr19a}. Concretely, let $\lambda^\ast$ denote a reference configuration and $\hat{g}$ the surrogate model trained on the run history $\mathcal{H}$. In our implementation, $\hat{g}$ is a Gaussian Process. HyperSHAP treats each hyperparameter $\Lambda^j$ as a player and computes its Shapley value $h(\Lambda^j)$, measuring the expected marginal gain in $\hat{g}$ from tuning $\Lambda^{j}$ away from $\lambda^\ast$, averaged over all possible subsets of co-tuned hyperparameters. Because the Shapley values are defined relative to a reference configuration rather than by variance decomposition, they capture directed performance \text{gains} and naturally account for higher-order interactions. In the multi-objective setting, we leverage ParEGO's scalarization to reduce this to a single-objective tunability problem at each iteration.
By the efficiency axiom of Shapley values, the individual contributions sum to the total gain over the reference $\sum_{\Lambda^j\in\Lambda}h(\Lambda^j) = \hat{g}(\lambda^\ast_\Lambda) - \hat{g}(\lambda^\ast)$, where $\lambda^\ast_\Lambda$ denotes the configuration obtained by tuning all hyperparameters. We exploit this property to reduce the configuration space by selecting the smallest subset that accounts for at least a $\tau$-fraction of the total gain
\[
\argmin_{\Lambda' \subseteq \Lambda} |\Lambda'| \quad \text{subject to} \quad \frac{\sum_{\Lambda^{j} \in \Lambda'} h(\Lambda^{j})}{\sum_{\Lambda^{k} \in \Lambda} h(\Lambda^{k})} \geq \tau.
\]
In practice, we solve this greedily by ordering hyperparameters by their Shapley values and selecting the top contributors until their cumulative share exceeds $\tau$. Hyperparameters contributing less than 5\% of the total gain are not selected. If an alternative HPI method is used, a simpler selection rule, such as retaining all hyperparameters above the $\tau$-quantile, can be substituted.

We consider only the most influential hyperparameters and reduce optimization complexity by constructing a reduced configuration space $\Lambda' \subset \Lambda$, by diminishing the influence of less important hyperparameters. Based on the importance scores $h(\Lambda^j)$, we retain only the most relevant hyperparameters $\Lambda' \subset \Lambda$ and fix the rest to constants (we use the incumbent). We fix unselected conditional hyperparameters that are inactive in the incumbent to their defaults, and for selected conditional hyperparameters, we add their parents to the selected set. Each reduction starts from the original configuration space $\Lambda$.
In rare cases, this leads to a too-narrow configuration space, such that no new configurations can be selected; we then fall back to the original space $\Lambda$ for that iteration.\footnote{If no hyperparameter was considered important, we fall back to the original space. If only one hyperparameter was selected, we add another hyperparameter at random. We do this to maintain some diversity in configurations and to avoid encouraging an early, overconfident surrogate model to tune only one hyperparameter.}  

\subsection{Dynamic Threshold}\label{sec:threshold}
The threshold $\tau$ determining the selected hyperparameters changes over the course of the optimization.
Since both surrogate models (for optimization and HPI estimation) require a sufficient number of trials to provide reliable estimates, we disable hyperparameter reduction during the initial third of trials (initialization phase).
This allows broad exploration and reliable estimation of HPI.
Then we apply a threshold of $\tau = 0.8$, corresponding to 80\% of the cumulative contribution measured by the Shapley values, to focus the search on the most relevant hyperparameters and accelerate progress. 

\section{Empirical Evaluation} 
In this section, we begin by describing our experimental setup and then evaluate the proposed HPI-ParEGO along the following research questions:
\begin{inparaenum}[(i)]
    \item How well does the HPI-selection work for single-objective optimization? (Sec.~\ref{sub:soo})
    \item Is our HPI-based approach more effective for MOO than the standard ParEGO algorithm on well-understood synthetic benchmark functions? (Sec.~\ref{sub:pymoos})
    \item How do the design choices of HPI-ParEGO impact its performance? (Sec.~\ref{sec:ablation_study})
    \item How well do the results translate to real-world HPO benchmarks? (Sec.~\ref{sub:yahpo})    
\end{inparaenum}

\subsection{Experiment Setup}\label{sec:exp_setup}

\textbf{Datasets.} 
We conduct our experiments on synthetic functions from PyMOO~\citep{blank-ieeeaccess20a} and HPO tasks from \yahpo \citep{pfisterer-automl22a}, spanning a range of configuration space dimensionalities. From \pymoo, we select the \texttt{ZDT} suite \citep{zitzler-ec00a} (30 dimensions for \texttt{ZDT1-3}, 10 for the others), excluding \texttt{ZDT5}, which is defined over a discrete bitstring domain and therefore less amenable to conversion into an HPO problem. From \yahpo, we select 13 tasks each from \texttt{LCBench}~\citep{zimmer-tpami21a} (7 dimensions) and \texttt{rbv2\_ranger} (8 dimensions)\footnote{Fidelity hyperparameters are fixed to the highest fidelity.}, chosen because different hyperparameters matter for different objective trade-offs (see Appendix~\ref{app:task_selection} for details). Of these, three tasks per scenario serve as an initial ablation study, with the remaining 20 reserved for the final evaluation to avoid biasing it.
Additionally, we evaluate on 10 randomly selected tasks from each scenario, \texttt{LCBench} and \texttt{rbv2\_ranger}, to show our algorithm also performs well on tasks where the property is not given.

\textbf{Baselines.}
Our primary baseline is the ParEGO implementation in the HPO framework SMAC3 \citep{lindauer-jmlr22a}, without our HPI modifications. It uses a random forest as the surrogate model and expected improvement as the acquisition function, a combination that has shown strong performance in prior HPO benchmark studies~\citep{eggensperger-neuripsdbt21a}.
The acquisition function is optimized by sorted random search.
The surrogate model is retrained every two evaluations, scalarization weights are updated every ten iterations ($u = 10$), and a random configuration is evaluated with a random chance $r=10\%$.
The default configuration of the configuration space is always evaluated first.
We also compare against several established MOO algorithms: multi-objective TPE \citep{ozaki-gecco20a}, NSGA-II \citep{deb-tec02a}, and EHVI~\citep{daulton-neurips20a}, all implemented in Optuna \citep{akiba-kdd19a}\footnote{For EHVI, we use Optuna's BoTorchSampler with default settings.}, and DE from Nevergrad \citep{rapin-nevergrad18a}.

\textbf{Evaluation Metrics.}
We track the normalized best-so-far performance using two common multi-objective indicators: inverted generational distance plus (IGD+) and hypervolume (HV).
IGD+ quantifies the average distance from each point in a reference Pareto front to its closest point in the solution set~\citep{ishibuchi-emco15a}.
The reference Pareto front pools the costs across optimizers and seeds and is normalized per task to a \mbox{[0,1]} scale.
The HV indicator is the volume of objective space dominated by the current Pareto front, bounded by a reference point set to the componentwise maximum of all observed cost vectors; we negate the HV to minimize it consistently.
For both indicators, we apply per-task min-max normalization, yielding normalized regrets on a \mbox{[0,1]} scale where lower is better.
Convergence plots show these regrets over normalized trials (i.e., function evaluations), averaged first across tasks and then across seeds, with shaded regions indicating $\pm 1$ standard error.
For the ablation study, we report the mean area under the curve (AUC) of the per-seed convergence curves for each indicator, each averaged across tasks as a summary measure.
We mark differences as significant (*) if a paired Wilcoxon signed-rank test~\citep{wilcoxon-biometrics45a} with Holm correction~\citep{holm-sjs79a} over all variants yields $p < 0.05$.

\textbf{Implementation Details.}
Our HPI-ParEGO variant inherits all settings from the ParEGO baseline and augments them with our HPI extension, ensuring that any performance differences are attributable to the proposed method.
All experiments are executed through the \carps ~\citep{benjamins-tmlr26a}, using 10 random seeds for the main evaluation and 5 for the ablation study. More details on implementation, hardware, and per-task trial counts are provided in Appendices \ref{app:ablation}, \ref{app:hardware}, and \ref{app:n_trials}.

\needspace{8\baselineskip}
\subsection{Single Objective Optimization}\label{sub:soo}
\begin{wrapfigure}{r}{0.45\columnwidth}
    \vspace{-20pt}
  \includegraphics[width=\linewidth]{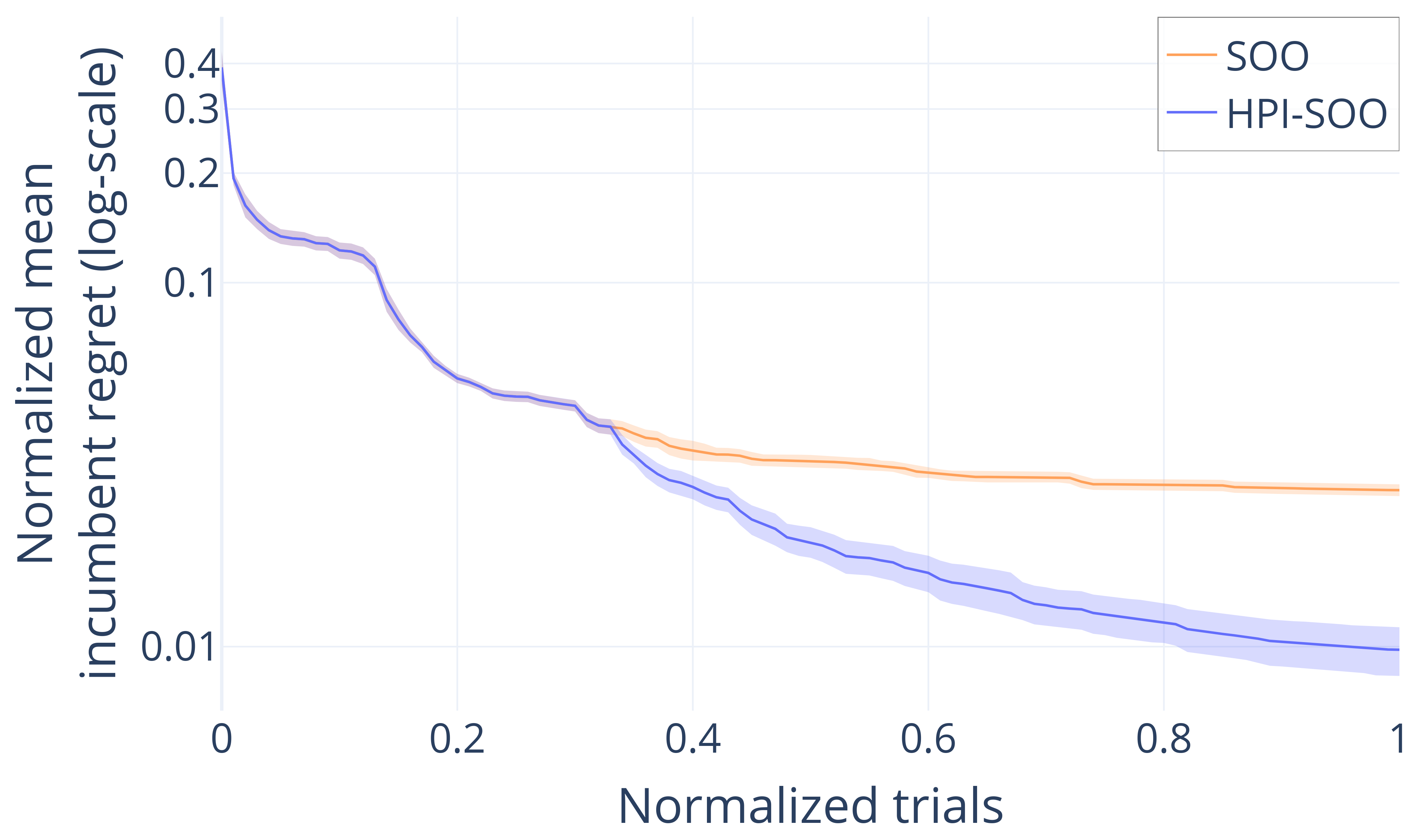}
  \caption{The HPI-based optimizer for single-objective optimization compared to baseline HPO on the selected \texttt{LCBench} tasks.}
  \label{fig:soo}
\end{wrapfigure}

We first evaluate whether HPI-based configuration space reduction improves convergence speed in the simpler single-objective optimization (SOO) setting. To this end, we replace ParEGO with SMAC's default HPO optimizer while keeping all other settings identical to those used in the MOO experiments to ensure a fair comparison.
We target validation accuracy on the selected \texttt{LCBench} tasks and measure performance via normalized mean incumbent regret.
As shown in Figure \ref{fig:soo}, the HPI-reduced configuration space consistently outperforms the full configuration space baseline in SOO, lending confidence that the benefits of HPI transfer to the MOO setting studied in the remainder of this paper.

\subsection{HPI-ParEGO on PyMOO}\label{sub:pymoos}
\begin{figure*}[t]
    \centering
    \begin{subfigure}[t]{0.5\textwidth}
      \centering
      \includegraphics[width=\columnwidth]{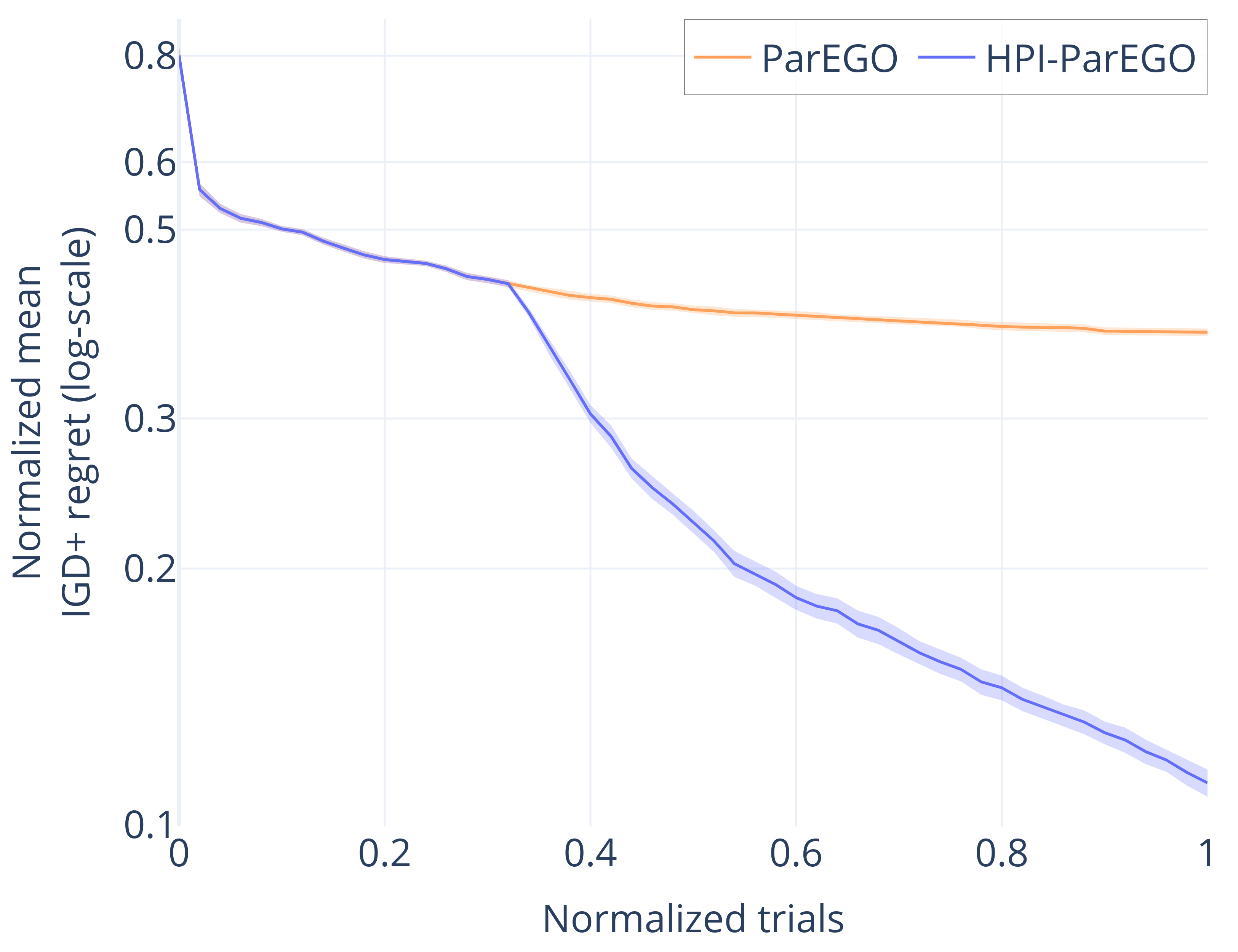}
      \caption{The HPI-ParEGO optimizer compared to the ParEGO optimizer on IGD+ regret.}
      \label{fig:pymoo_mo}
    \end{subfigure}%
    ~ 
    \begin{subfigure}[t]{0.5\textwidth}
          \centering
          \includegraphics[width=\columnwidth]{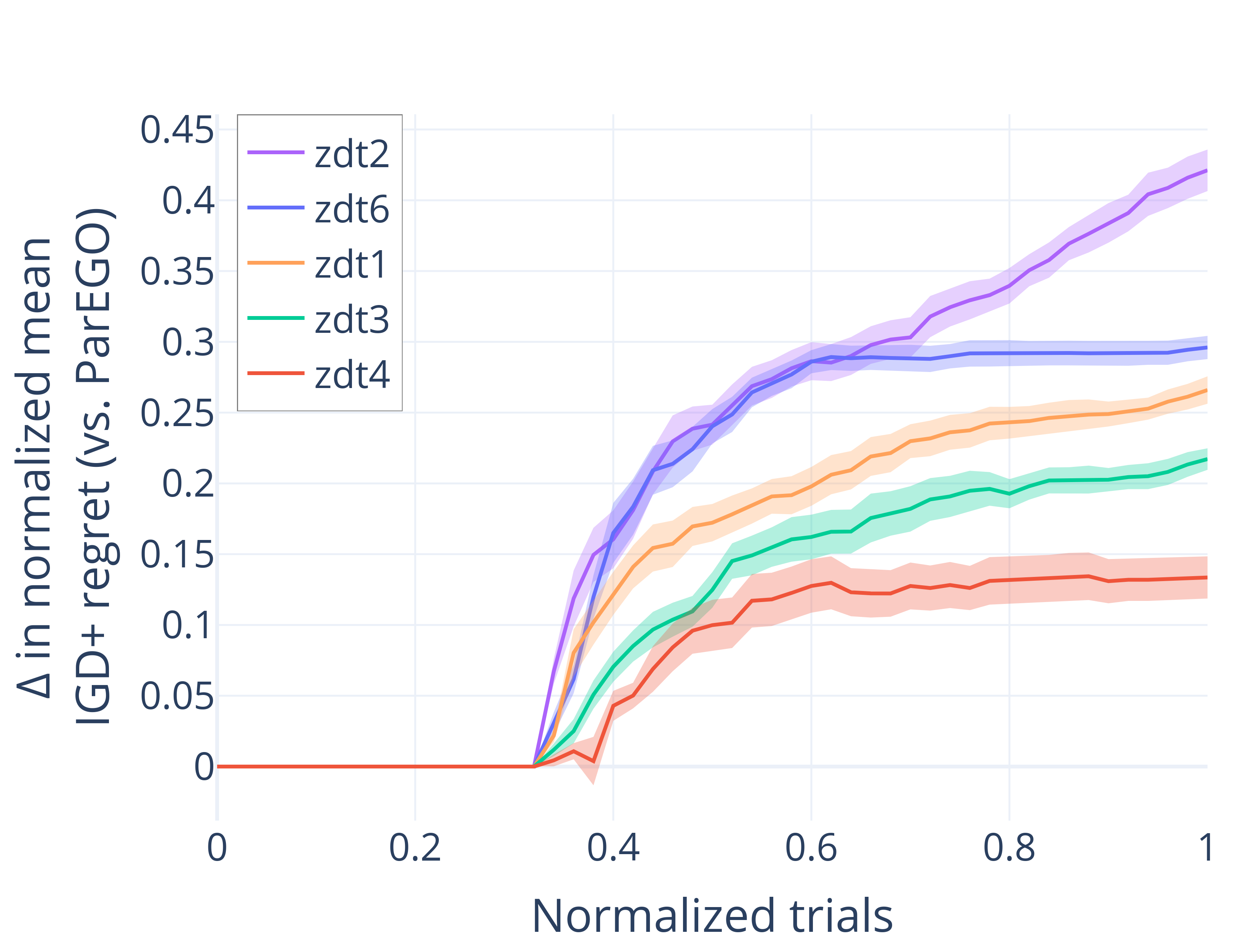}
          \caption{Difference in mean IGD+ regret with standard error between ParEGO and HPI-ParEGO. Values $> 0$ indicate a better performance of HPI-ParEGO.}
          \label{fig:pymoo_task_diff}
    \end{subfigure}
    \caption{Results on the \pymoo tasks.}
\end{figure*}

We next evaluate our approach on the selected \pymoo tasks (Figure \ref{fig:pymoo_mo}), which are established synthetic MOO benchmarks that allow for detailed analysis. HPI-ParEGO begins to outperform the ParEGO baseline at roughly $33\%$ of the trial budget, precisely when the HPI-based configuration space reduction begins (see Subsection~\ref{sec:threshold}). This indicates that the most important hyperparameters are identified early.
Per-task breakdowns (Figure~\ref{fig:pymoo_task_diff}) confirm that HPI-ParEGO substantially outperforms ParEGO across all \pymoo functions.
Together, these results demonstrate that HPI-ParEGO handles well-known synthetic benchmarks effectively and accelerates convergence throughout the trial budget. We further examine its behavior as dimensionality increases in Appendix~\ref{app:dim}.
Appendix~\ref{app:pymoo} shows the HV and IGD+ results compared to all baselines. We note that, by construction, ParEGO can have slight advantages on HV compared to other baselines, and IGD+ is seen as a fairer metric. Appendix~\ref{app:pymoo} shows that on these artificial benchmarks, an HV-based acquisition function (EHVI) efficiently exploits the fairly simple Pareto fronts across both metrics. However, as we show later, this advantage does not generalize to more complex real-world problems.

\subsection{Ablation Study} \label{sec:ablation_study}
\begin{figure}[t]
\centering
\begin{subfigure}[c]{0.53\columnwidth}
    \centering
    \resizebox{\linewidth}{!}{
        \begin{tabular}{ll}
        \toprule
        \textbf{Component} & \textbf{Values} \\
        \midrule
        HPI method: & HyperSHAP, random \\
        & subselection, MCTS-VS \\
        &criterion~\citep{song-neurips22a} \\
        Fixing strategy: & incumbent, default, random \\
        Threshold $\tau$-schedule: & Initialization (0,0.8,0.8),\\
        & Consolidation (0.8,0.8,0), \\
        & Constant-0.8,\\ 
        & Symmetric-0.8 (0,0.8,0)\\
        Random chance $r$: & 0.0, 0.1, 0.2 \\
        Weight update iterations $u$: & 1, 6, 10, 20 \\
        Include interactions & true, false \\
        HyperSHAP surrogate model & RF, GP\\
        \bottomrule
        \end{tabular}
    }
    \caption{Explored settings of the ablation study.}
    \label{tab:ablation}
\end{subfigure}%
\hfill
\begin{subfigure}[c]{0.43\columnwidth}
    \centering
    \includegraphics[width=\linewidth]{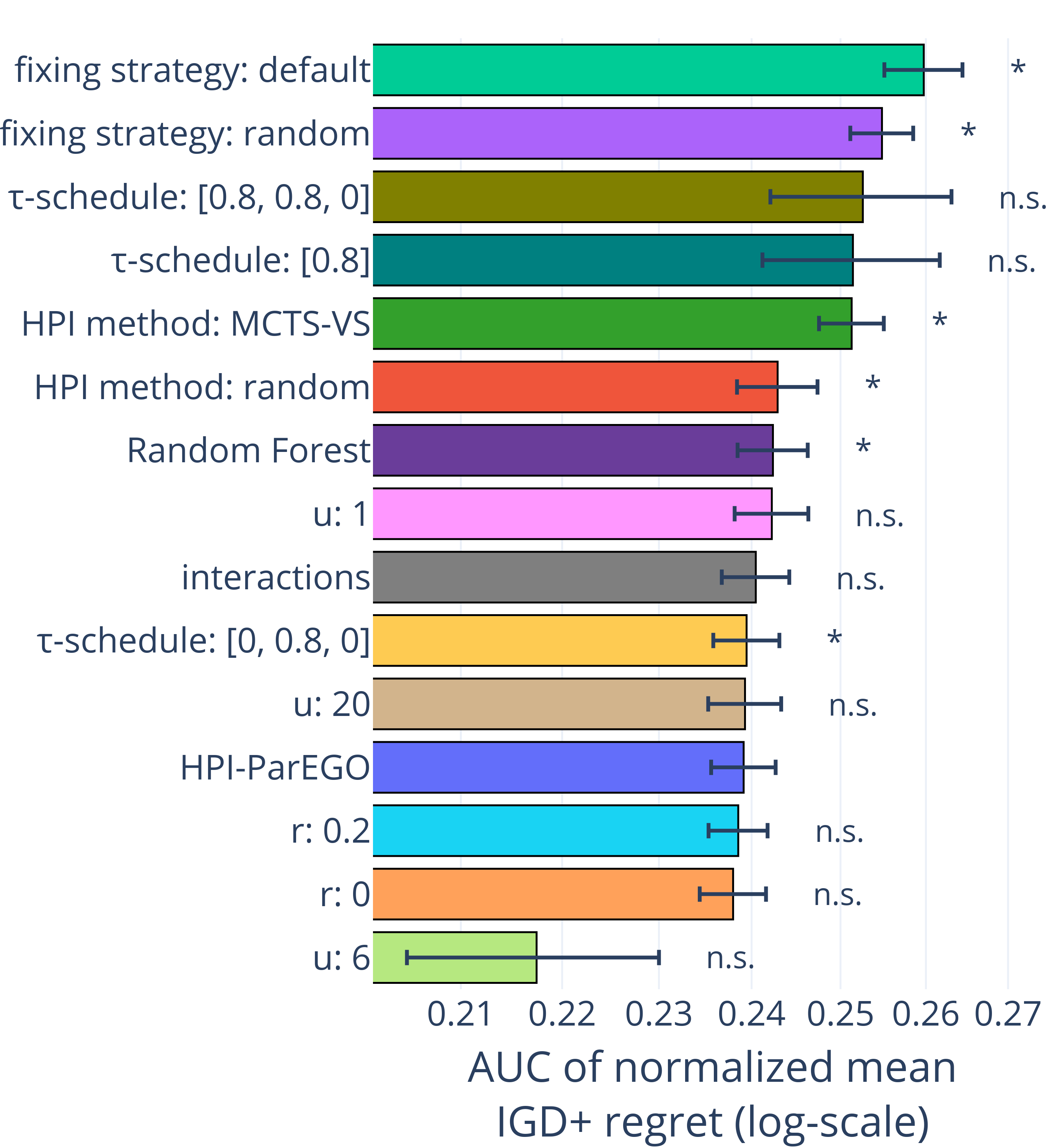}
    \caption{AUC of IGD+ regret for ablating over different algorithm components.}
    \label{fig:ablation}
\end{subfigure}
\caption{Ablation study results.}
\end{figure}

To address our third research question of how the design choices of HPI-ParEGO affect its performance, we conduct a systematic ablation study, in which we \textit{vary one component at a time} while holding all others fixed to the configuration of our proposed method (Table~\ref{tab:ablation}). The study is carried out on six held-out tasks, three from \texttt{LCBench} and three from \texttt{rbv2\_ranger}, which are excluded from all subsequent evaluations to guard against overfitting to these design decisions. Full per-variation results with IGD+ regret and HV regret are provided in Appendix~\ref{app:ablation}.
Figure~\ref{fig:ablation} summarizes the findings by reporting the AUC and statistical significance as each component is varied away from the HPI-ParEGO default.

Most variations are not statistically significant.
The most impactful design choice is the strategy for fixing unimportant hyperparameters: setting them to their current incumbent values yields a substantial advantage over using defaults or random values.
This is intuitive, as HyperSHAP measures tunability relative to the incumbent; fixing low-importance hyperparameters to that reference preserves a strong configuration while concentrating the optimization budget on high-impact dimensions.
Replacing HyperSHAP selection with random selection leads to a consistent performance drop, confirming that HPI provides a more informative criterion.
Appendix~\ref{app:random_hpi} shows that the gap is a lot stronger on the 10 selected \texttt{LCBench} tasks.
Ablating the reduction schedule shows that the initialization phase is important for good performance.
We tested considering interactions between hyperparameters, which did not perform better than HPI-ParEGO, likely because such interactions are even more difficult for a surrogate model to learn than HPI~\citep{wever-aaai26a}. 
To calculate HyperSHAP, we compared two surrogate models: the random forest (RF) used by SMAC for optimization and a Gaussian Process (GP) trained on the current run history. GP showed significantly higher performance, likely because it is better at estimating small differences near the incumbent.
Finally, the occasional evaluation of the unreduced space and the frequency of weight updates each contribute a small but positive effect.
Interestingly, the comparison with ground-truth importance shows only a small gap, indicating that HyperSHAP with the GP estimates HPI very well (see Appendix~\ref{app:ground_truth}).

\begin{figure*}[t]
    \centering
    \begin{subfigure}[t]{0.5\textwidth}
      \includegraphics[width=\columnwidth]{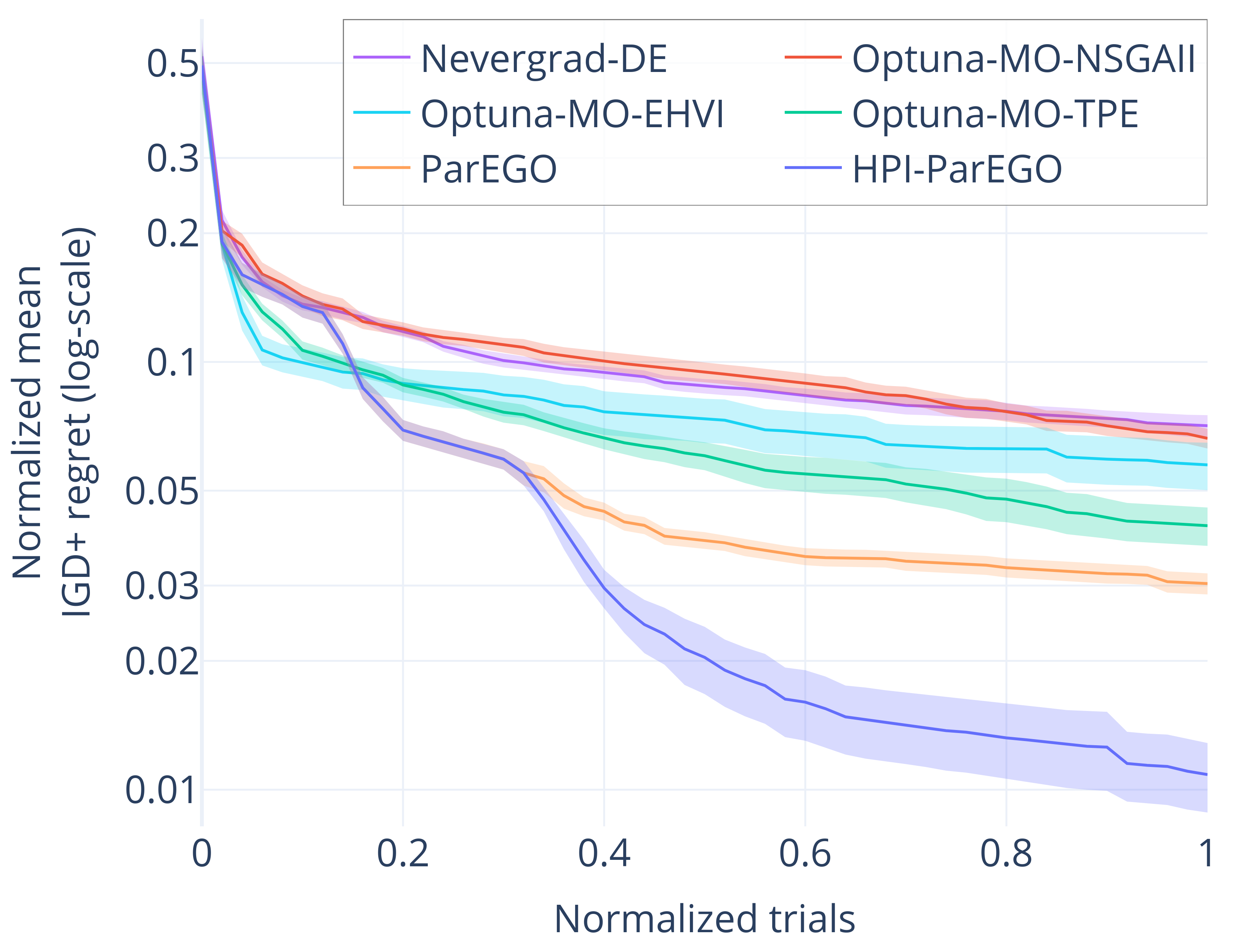}
      \caption{Results on the selected \texttt{LCBench} tasks.}
    \end{subfigure}%
    ~ 
    \begin{subfigure}[t]{0.5\textwidth}
      \centering
      \includegraphics[width=\columnwidth]{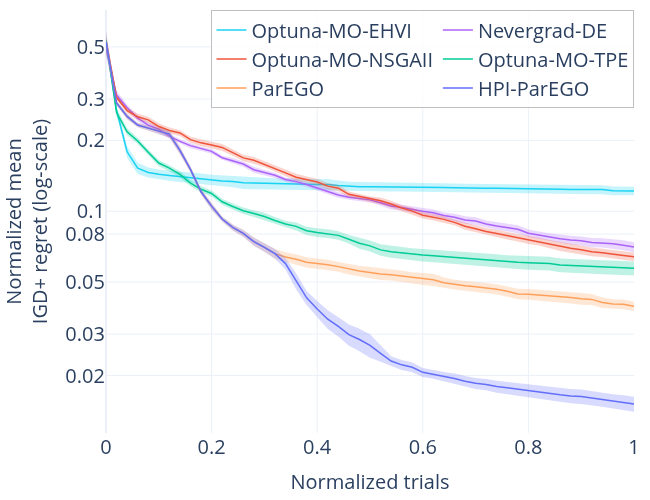}
      \caption{Results on the selected \texttt{rbv2\_ranger} tasks.}
    \end{subfigure}
    \caption{Results of HPI-ParEGO compared to the baseline optimizers measured in IGD+ regret.}
\label{fig:yahpo}
\end{figure*} 

\subsection{HPI-ParEGO on HPO Benchmarks}\label{sub:yahpo}
Lastly, we evaluate on real-world HPO problems, comparing HPI-ParEGO against ParEGO and the additional baselines on the selected \yahpo tasks (Figure~\ref{fig:yahpo}). For both tasks, HPI-ParEGO emerges as the best optimizer, overtaking EHVI and MO-TPE at around 15-20\% of the trial budget, with a large gap to ParEGO. While learning an accurate HPI can be more challenging on real HPO benchmarks, HPI-ParEGO learned it well, even on \texttt{rbv2\_ranger}, where conditional configuration spaces add further complexity. Interestingly, EHVI stagnates very quickly on both scenarios, showing that EHVI's strong results on \pymoo (Appendix~\ref{app:pymoo}) cannot be reproduced on complex real-world tasks. Appendix~\ref{app:yahpo_random_tasks} shows the results of HPI-ParEGO on randomly selected tasks of \texttt{LCBench} and \texttt{rbv2\_ranger}, with HPI-ParEGO still outperforming ParEGO. In summary, HPI-ParEGO performs strongly across both benchmark suites. HV regret results and per-task differences between HPI-ParEGO and ParEGO for each scenario are shown in Appendix \ref{app:task_diff}. Appendix~\ref{app:qnparego} applies the algorithm to qNParEGO~\citep{daulton-neurips20a}, another scalarization-based optimizer, which outperforms the standard version of qNParEGO.

\subsection{Computational Overhead}\label{sec:overhead}
Calculating HPI every iteration adds some computational overhead. On \pymoo, we measured it as $\sim58$ seconds per trial (s/trial) and on \yahpo $\sim9$ s/trial relative to ParEGO. Since we use surrogate benchmarks, we do not know the evaluation cost. This overhead is fixed and independent of the evaluation cost, which in HPO settings ranges from minutes to hours. Under a pessimistic 58 s/trial overhead, HPI-ParEGO remains favorable whenever it matches ParEGO's performance with fewer trials (e.g., 400 vs. 500) and the per-trial evaluation cost exceeds 232 s $(400\times(58+c) < 500\times c$, where $c$ denotes evaluation cost). This is normally true for HPO on modern DL.

\section{Conclusion}\label{sec:discussion}
In this work, we proposed a novel hyperparameter optimization method based on ParEGO that incorporates dynamic HPI into the optimization. Our method leverages ParEGO's scalarization mechanism to identify the most influential hyperparameters under varying trade-offs between objectives. Through extensive experiments on PyMOO and \yahpo, spanning synthetic and real-world multi-objective tasks, we demonstrated that HPI-ParEGO consistently outperforms standard ParEGO and most of our baseline optimizers in low- to moderate-dimensional tasks.
This supports our hypothesis: Many hyperparameters contribute little to performance under specific objective trade-offs, and dynamically filtering them during optimization helps the optimizer focus on the relevant subspace.
Calculating HPI adds a bit of computational overhead, but it is fixed on a given configuration space and negligible relative to typical evaluation costs in real HPO settings, making the method practical for HPO on modern DL (see Section~\ref{sec:overhead} for a detailed analysis).

A full theoretical understanding of when importance-guided filtering benefits MOO remains open. Our preliminary understanding suggests that in applications with many unimportant hyperparameters, HPI-ParEGO is more effective at quickly identifying them. However, a highly variable HPI with many slightly important hyperparameters is harder to learn with a reduced optimization budget.
Nevertheless, our results clearly show that HPI should not remain a post-hoc diagnostic tool, but can be used during optimization, as previous work has done for SOO \citep{wang-kbs21a}, even in MOO. This opens up a new class of resource-efficient MOO methods that can be important for Green AI and Green AutoML, emphasizing lower compute costs and reduced energy consumption without sacrificing accuracy~\citep{schwartz-arxiv19a,tornede-jair23a}.
By focusing on the most relevant hyperparameters, our method reduces computational costs and makes HPO more accessible, though practitioners should carefully consider objective selection to avoid encoding biases or misuse.

\subsection{Limitations} 
Our method relies on an accurate surrogate model for reliable HPI estimation, which can be a bottleneck in early iterations~\citep{wever-aaai26a} and data-scarce scenarios, despite our mitigation via interleaving random configurations and delayed reduction (via the initialization phase).
However, the $\tau$-schedule assumes a reasonably set budget; for very few trials, the $\tau$-schedule might not work well because the reduction occurs too early.
Moreover, the performance and surrogate model quality may degrade as the dimensionality of the configuration space increases, especially since HyperSHAP can incur excessive overhead, making it too slow in practice for very large-scale HPO problems modeling full ML pipelines, e.g., \citet{kotthoff-jmlr17a,feurer-jmlr22a}.
Our scalability analysis (Appendix~\ref{app:dim}) is limited to synthetic benchmarks; whether the same holds for high-dimensional real-world HPO tasks remains untested.
Furthermore, we selected YAHPO-Gym tasks with varying HPI across trade-offs to test the mechanism under the conditions it targets. This may favor our method, but we chose the tasks before any optimization was done. Moreover, we showed that our method also performs well on randomly selected tasks.

\subsection{Future Work} 
There are several opportunities for future work to overcome these limitations: (i)~exploration of more robust and uncertainty-aware HPI estimation techniques, for example, using stochastic Shapley values~\citep{neurips-chau23a}; (ii) meta-learning of HPI priors from previous HPO tasks; or (iii) reducing $\tau$ in a data-driven way.
Furthermore, a common approach to reduce the high cost of model training is to use multi-fidelity optimization~\citep{swersky-arxiv14a,falkner-icml18a,li-mlsys20a,bohdal-iclr23a}, but it is an open problem how much information, e.g., on HPI, can be leveraged from cheap training runs. Results from \citet{zimmer-tpami21a} suggest that HPI might be stable enough at least on simple DNNs.
Moreover, our method is currently tailored to scalarization-based optimizers such as ParEGO, which limits generality. Extending the method to other MOO paradigms, such as nondominated-sorting-based evolutionary algorithms, requires rethinking how HPI is applied. 
Finally, we do not cover many-objective optimization with more than three objectives, which requires large optimization budgets, but the sequential nature and ParEGO's surrogate model are better suited to constrained optimization budgets. Therefore, extensions of our idea to NSGA-III \citep{deb-ieeetvec-14a} and the population-based contexts are promising future work.

\subsection*{AI use statement}
In this work, we used generative AI tools to aid and polish writing and to draft sections of the paper.
We have not used generative AI tools for retrieval and discovery, or research ideation and execution. Generating synthetic datasets and proving mathematical claims are not applicable to this work. We have reviewed all AI-assisted work. All authors verified the content and text for correctness. We take responsibility for the final content of this work, including text, claims or artifacts produced with the aid of generative AI.

\subsection*{Reproducibility statement}
HPI-ParEGO is specified in Algorithm~\ref{alg:hpi_parego} and Section~\ref{sec:method}, and the experimental setup is in Section~\ref{sec:exp_setup} and Appendices ~\ref{app:task_selection}--\ref{app:ablation}. All benchmarks and frameworks are publicly available. Anonymized code to reproduce all experiments is provided as supplementary material.

\newpage
\bibliography{references}

@inproceedings{adachi-aistats24a,
  author    = {M. Adachi and B. Planden and D. A. Howey and M. A. Osborne and S. Orbell and N. Ares and K. Muandet and S. L. Chau},
  title     = {Looping in the Human: Collaborative and Explainable {Bayesian} Optimization},
  booktitle = {Proc. of {AISTATS}'24},
  year      = {2024}
}

@inproceedings{akiba-kdd19a,
  title     = {Optuna: A Next-Generation {H}yperparameter {O}ptimization Framework},
  author    = {T. Akiba and S. Sano and T. Yanase and T. Ohta and M. Koyama},
  pages     = {2623--2631},
  booktitle = {Proc. of {KDD}'19},
  year      = {2019}
}

@article{arxiv-basu25a,
  title   = {Multi-Objective Hyperparameter Optimization in the Age of Deep Learning},
  author  = {Basu, S. and Hutter, F. and Stoll, D.},
  journal = {arXiv preprint arXiv:2511.08371},
  year    = {2025}
}

@inproceedings{balandat-neurips20a,
  title     = {BoTorch: A Framework for Efficient Monte-Carlo {Bayesian} Optimization},
  author    = {M. Balandat and B. Karrer and D. Jiang and S. Daulton and B. Letham and A. Wilson and E. Bakshy},
  year      = {2020},
  booktitle = {Proc. of {N}eur{IPS}'20}
}

@book{bellman-book57a,
  title     = {Dynamic Programming},
  author    = {Bellman, R. and Rand Corporation and Karreman Mathematics Research Collection},
  isbn      = {9780691079516},
  lccn      = {57005444},
  series    = {Rand Corporation Research Study},
  year      = {1957},
  publisher = {Princeton University Press}
}

@article{benjamins-tmlr26a,
  title   = {carps: A Framework for Comparing N Hyperparameter Optimizers on M Benchmarks},
  author  = {C. Benjamins and H. Graf and S. Segel and D. Deng and T. Ruhkopf and L. Hennig and S. Basu and N. Mallik and E. Bergman and D. Chen and F. Clément and A. Tornede and M. Feurer and K. Eggensperger and F. Hutter and C. Doerr and M. Lindauer},
  year    = {2026},
  journal = {Transactions of Machine Learning Research (TMLR)}
}

@article{bergstra-jmlr12a,
  title    = {Random Search for Hyper-Parameter Optimization},
  author   = {J. Bergstra and Y. Bengio},
  year     = {2012},
  journal  = {Journal of Machine Learning Research},
  volume   = {13},
  pages    = {281--305}
}

@inproceedings{biedenkapp-aaai17a,
  title     = {Efficient Parameter Importance Analysis via Ablation with Surrogates},
  author    = {A. Biedenkapp and M. Lindauer and K. Eggensperger and C. Fawcett and H. Hoos and F. Hutter},
  pages     = {773--779},
  booktitle = {Proc. of {AAAI}'17},
  year      = {2017}
}

@inproceedings{biedenkapp-lion18a,
  title     = {{CAVE}: Configuration Assessment, Visualization and Evaluation},
  author    = {A. Biedenkapp and J. Marben and M. Lindauer and F. Hutter},
  booktitle = {Proc. of {LION}'18},
  year      = {2018}
}

@article{bischl-dmkd23a,
  title     = {Hyperparameter Optimization: Foundations, Algorithms, Best Practices, and Open Challenges},
  author    = {B. Bischl and M. Binder and M. Lang and T. Pielok and J. Richter and S. Coors and J. Thomas and T. Ullmann and M. Becker and A.{-}L. Boulesteix and D. Deng and M. Lindauer},
  year      = {2023},
  journal   = {Wiley Interdisciplinary Reviews: Data Mining and Knowledge Discovery},
  publisher = {Wiley Online Library},
  pages     = {e1484}
}

@article{blank-ieeeaccess20a,
  author  = {Blank, J. and Deb, K.},
  journal = {IEEE Access},
  title   = {{PyMOO}: Multi-Objective Optimization in {Python}},
  year    = {2020},
  volume  = {8},
  pages   = {89497-89509}
}

@inproceedings{bohdal-iclr23a,
  title     = {{PASHA}: Efficient {HPO} and {NAS} With Progressive Resource Allocation},
  author    = {O. Bohdal and L. Balles and M. Wistuba and B. Ermis and C. Archambeau and G. Zappella},
  booktitle = {Proc. of {ICLR}'23},
  year      = {2023}
}

@article{bradford-jgo18a,
  author  = {E. Bradford and A. Schweidtmann and A. Lapkin},
  title   = {Efficient Multiobjective Optimization Employing {Gaussian} Processes, Spectral Sampling and a Genetic Algorithm},
  journal = {J. Glob. Optim.},
  volume  = {71},
  number  = {2},
  pages   = {407--438},
  year    = {2018}
}

@article{daulton-arxiv26a,
  title   = {{BONSAI}: Bayesian Optimization with Natural Simplicity and Interpretability},
  author  = {Daulton, S. and Eriksson, D. and Balandat, M. and Bakshy, E.},
  journal = {arXiv},
  year    = {2026}
}

@inproceedings{daulton-neurips20a,
  title     = {Differentiable Expected Hypervolume Improvement for Parallel {M}ulti-{O}bjective {B}ayesian Optimization},
  author    = {S. Daulton and M. Balandat and E. Bakshy},
  pages     = {9851--9864},
  booktitle = {Proc. of {N}eur{IPS}'20},
  year      = {2020}
}

@article{deb-ieeetvec-14a,
  author  = {Deb, K. and Jain, H.},
  journal = {IEEE Transactions on Evolutionary Computation},
  title   = {An Evolutionary Many-Objective Optimization Algorithm Using Reference-Point-Based Nondominated Sorting Approach, {Part I}: Solving Problems With Box Constraints},
  year    = {2014},
  volume  = {18},
  number  = {4},
  pages   = {577-601}
}

@article{deb-tec02a,
  author  = {K. Deb and A. Pratap and S. Agarwal and T. Meyarivan},
  journal = {IEEE Transactions on Evolutionary Computation},
  title   = {A Fast and Elitist Multiobjective Genetic Algorithm: {NSGA-II}},
  year    = {2002},
  volume  = {6},
  number  = {2},
  pages   = {182-197}
}

@inproceedings{eggensperger-neuripsdbt21a,
  title     = {{HPOBench}: A Collection of Reproducible Multi-Fidelity Benchmark Problems for {HPO}},
  author    = {K. Eggensperger and P. M{\"u}ller and N. Mallik and M. Feurer and R. Sass and A. Klein and N. Awad and M. Lindauer and F. Hutter},
  booktitle = {Proc. of {N}eur{IPS}'21 Datasets and Benchmarks Track},
  year      = {2021}
}

@inproceedings{elsken-iclr19a,
  title     = {Efficient Multi-Objective {N}eural {A}rchitecture {S}earch via Lamarckian Evolution},
  author    = {T. Elsken and J. Metzen and F. Hutter},
  booktitle = {Proc. of {ICLR}'19},
  year      = {2019}
}

@inproceedings{eriksson-neurips19a,
  title     = {Scalable Global Optimization via Local {Bayesian} Optimization},
  author    = {D. Eriksson and M. Pearce and J. Gardner and R. Turner and M. Poloczek},
  booktitle = {Proc. of {N}eur{IPS}'19},
  year      = {2019}
}

@inproceedings{eriksson-uai21a,
  title     = {High-dimensional {Bayesian} optimization with sparse axis-aligned subspaces},
  author    = {D. Eriksson and M. Jankowiak},
  pages     = {493--503},
  booktitle = {Proc. of {UAI}'21},
  year      = {2021}
}

@inproceedings{falkner-icml18a,
  title     = {{BOHB}: Robust and Efficient {H}yperparameter {O}ptimization at Scale},
  author    = {S. Falkner and A. Klein and F. Hutter},
  pages     = {1437--1446},
  booktitle = {Proc. of {ICML}'18},
  year      = {2018}
}

@article{fawcett-heu16a,
  title   = {Analysing differences between algorithm configurations through ablation},
  author  = {C. Fawcett and H. Hoos},
  year    = {2016},
  journal = {Journal of Heuristics},
  volume  = {22},
  number  = {4},
  pages   = {431--458}
}

@incollection{feurer-automlbook19a,
  title     = {Hyperparameter {O}ptimization},
  author    = {M. Feurer and F. Hutter},
  pages     = {3 -- 38},
  chapter   = {1},
  year      = {2019},
  booktitle = {Automated Machine Learning: Methods, Systems, Challenges},
  publisher = {Springer},
  note      = {Available for free at \url{http://automl.org/book}},
  editor    = {F. Hutter and L. Kotthoff and J. Vanschoren}
}

@article{feurer-jmlr22a,
  title   = {{Auto-Sklearn} 2.0: Hands-free {AutoML} via Meta-Learning},
  author  = {M. Feurer and K. Eggensperger and S. Falkner and M. Lindauer and F. Hutter},
  year    = {2022},
  journal = {Journal of Machine Learning Research},
  volume  = {23},
  number  = {261},
  pages   = {1--61},
  editor  = {M. Schoenauer}
}

@book{fudenberg-mit91a,
  title     = {Game Theory},
  author    = {Fudenberg, D. and Tirole, J.},
  year      = {1991},
  publisher = {MIT press}
}

@article{holm-sjs79a,
  author    = {S. Holm},
  title     = {A Simple Sequentially Rejective Multiple Test Procedure},
  journal   = {Scandinavian Journal of Statistics},
  volume    = {6},
  number    = {2},
  pages     = {65--70},
  year      = {1979},
  publisher = {JSTOR}
}

@inproceedings{hutter-icml14a,
  title     = {An Efficient Approach for Assessing Hyperparameter Importance},
  author    = {F. Hutter and H. Hoos and K. Leyton-Brown},
  pages     = {754--762},
  booktitle = {Proc. of {ICML}'14},
  year      = {2014}
}

@inproceedings{hutter-lion13a,
  title     = {Identifying Key Algorithm Parameters and Instance Features using Forward Selection},
  author    = {F. Hutter and H. Hoos and K. Leyton-Brown},
  pages     = {364--381},
  booktitle = {Proc. of {LION}'13},
  year      = {2013}
}

@inproceedings{ishibuchi-emco15a,
  author    = {Ishibuchi, H. and Masuda, H. and Tanigaki, Y. and Nojima, Y.},
  title     = {Modified Distance Calculation in Generational Distance and Inverted Generational Distance},
  booktitle = {Evolutionary Multi-Criterion Optimization},
  year      = {2015},
  publisher = {Springer International Publishing},
  pages     = {110--125}
}

@article{jones-jgo98a,
  title    = {Efficient Global Optimization of Expensive Black Box Functions},
  author   = {D. Jones and M. Schonlau and W. Welch},
  year     = {1998},
  journal  = {Journal of Global Optimization},
  volume   = {13},
  pages    = {455--492}
}

@article{knowls-evoco06a,
  title   = {{ParEGO}: a hybrid algorithm with on-line landscape approximation for expensive multiobjective optimization problems},
  author  = {J. D. Knowles},
  year    = {2006},
  journal = {{IEEE} Transactions on Evolutionary Computation},
  volume  = {10},
  number  = {1},
  pages   = {50--66}
}

@article{kotthoff-jmlr17a,
  author  = {L. Kotthoff and C. Thornton and H. Hoos and F. Hutter and K. Leyton{-}Brown},
  title   = {{Auto-WEKA 2.0}: Automatic model selection and hyperparameter optimization in {WEKA}},
  journal = {Journal of Machine Learning Research},
  volume  = {18},
  pages   = {25:1--25:5},
  year    = {2017}
}

@article{lee-iasc22a,
  author  = {J. Lee and S. Ahn and H. Kim and J. Ruth Lee},
  title   = {Dynamic Hyperparameter Allocation Under Time Constraints for {Automated Machine Learning}},
  journal = {Intelligent Automation \& Soft Computing},
  volume  = {31},
  year    = {2022},
  number  = {1},
  pages   = {255--277}
}

@inproceedings{levesque-uai16a,
  title     = {Bayesian {Hyperparameter} {Optimization} for {Ensemble} {Learning}},
  author    = {J. Lévesque and C. Gagné and R. Sabourin},
  pages     = {437--446},
  booktitle = {Proc. of {UAI}'16},
  year      = {2016}
}

@inproceedings{li-mlsys20a,
  title     = {A System for Massively Parallel Hyperparameter Tuning},
  author    = {L. Li and K. Jamieson and A. Rostamizadeh and E. Gonina and J. Ben{-}tzur and M. Hardt and B. Recht and A. Talwalkar},
  booktitle = {Proc. of {MLSys}'20},
  year      = {2020}
}

@article{lindauer-jmlr22a,
  title   = {{SMAC3}: A Versatile Bayesian Optimization Package for {H}yperparameter {O}ptimization},
  author  = {M. Lindauer and K. Eggensperger and M. Feurer and A. Biedenkapp and D. Deng and C. Benjamins and T. Ruhkopf and R. Sass and F. Hutter},
  year    = {2022},
  journal = {Journal of Machine Learning Research},
  volume  = {23},
  number  = {54},
  pages   = {1--9}
}

@inproceedings{liu-aistats23a,
  title        = {Sparse {Bayesian} Optimization},
  author       = {Liu, S. and Feng, Q. and Eriksson, D. and Letham, B. and Bakshy, E.},
  booktitle    = {International Conference on Artificial Intelligence and Statistics},
  pages        = {3754--3774},
  year         = {2023},
  organization = {PMLR}
}

@article{moraleshernandez-air22a,
  author  = {A. Morales-Hernández and I. Van Nieuwenhuyse and S. Gonzalez},
  title   = {A survey on multi-objective hyperparameter optimization algorithms for Machine Learning},
  year    = {2022},
  journal = {Artificial Intelligence Review},
  volume  = {56},
  pages   = {8043--8093}
}

@inproceedings{neurips-chau23a,
  author    = {Chau, S. L. and Muandet, K. and Sejdinovic, D.},
  title     = {Explaining the Uncertain: Stochastic {Shapley} Values for {Gaussian} Process Models},
  year      = {2023},
  publisher = {Curran Associates Inc.},
  address   = {Red Hook, NY, USA},
  booktitle = {Proceedings of the 37th International Conference on Neural Information Processing Systems},
  articleno = {2209},
  numpages  = {27},
  location  = {New Orleans, LA, USA},
  series    = {NIPS '23}
}

@inproceedings{ozaki-gecco20a,
  title     = {Multiobjective Tree-Structured Parzen Estimator for Computationally Expensive Optimization Problems},
  author    = {Y. Ozaki and Y. Tanigaki and S. Watanabe and M. Onishi},
  pages     = {533–541},
  booktitle = {Proc. of {GECCO}'20},
  year      = {2020}
}

@book{pareto-manual71a,
  author     = {V. Pareto},
  title      = {Manual of Political Economy},
  year       = {1971},
  publisher  = {Augustus M. Kelley},
  address    = {New York},
  translator = {Ann S. Schwier and Alfred N. Page},
  note       = {Originally published in Italian as \emph{Manuale di Economia Politica}, 1906}
}

@inproceedings{pfisterer-automl22a,
  author    = {F. Pfisterer and L. Schneider and J. Moosbauer and M. Binder and B. Bischl},
  title     = {{YAHPO Gym} -- An Efficient Multi-Objective Multi-Fidelity Benchmark for Hyperparameter Optimization},
  booktitle = {Proc. of {AutoML} Conf'22},
  year      = {2022},
  publisher = {PMLR}
}

@article{probst-jmlr19a,
  title   = {Tunability: Importance of Hyperparameters of Machine Learning Algorithms},
  author  = {P. Probst and A. Boulesteix and B. Bischl},
  year    = {2019},
  journal = {Journal of Machine Learning Research},
  volume  = {20},
  number  = {53},
  pages   = {1--32}
}

@misc{rapin-nevergrad18a,
  title  = {Nevergrad - A Gradient-Free Optimization Platform},
  author = {Rapin, J. and Teytaud, O.},
  year   = {2018}
}

@inproceedings{rijn-kdd18a,
  title     = {Hyperparameter Importance Across Datasets},
  author    = {J. van Rijn and F. Hutter},
  pages     = {2367--2376},
  booktitle = {Proc. of {KDD}'18},
  year      = {2018}
}

@article{rodemann-arxiv24a,
  author  = {J. Rodemann and F. Croppi and P. Arens and Y. Sale and J. Herbinger and B. Bischl and E. H{\"{u}}llermeier and T. Augustin and C. Walsh and G. Casalicchio},
  title   = {Explaining {Bayesian} Optimization by {Shapley} Values Facilitates Human-{AI} Collaboration},
  journal = {CoRR},
  volume  = {abs/2403.04629},
  year    = {2024}
}

@inproceedings{schmucker-metalearn20a,
  author    = {R. Schmucker and M. Donini and V. Perrone and M. Zafar and C. Archambeau},
  title     = {Multi-Objective Multi-Fidelity Hyperparameter Optimization with Application to Fairness},
  booktitle = {{MetaLearn'20}},
  year      = {2020}
}

@article{schwartz-arxiv19a,
  title   = {Green {AI}},
  author  = {R. Schwartz and J. Dodge and N. A. Smith and O. Etzioni},
  year    = {2019},
  journal = {arXiv:1907.10597v3 [cs.CY]}
}

@inbook{shapley-book53a,
  title     = {A Value for {N}-Person Games},
  booktitle = {Contributions to the Theory of Games (AM-28), Volume II},
  author    = {L. S. Shapley},
  publisher = {Princeton University Press},
  address   = {Princeton},
  pages     = {307--318},
  year      = {1953}
}

@inproceedings{shen-automl23a,
  title     = {Computationally Efficient High-Dimensional {Bayesian} Optimization via Variable Selection},
  author    = {Shen, Y. and Kingsford, C.},
  year      = {2023},
  booktitle = {Proc. of {AutoML} Conf'23},
  publisher = {PMLR}
}

@inproceedings{snoek-nips12a,
  title     = {Practical {B}ayesian Optimization of Machine Learning Algorithms},
  author    = {J. Snoek and H. Larochelle and R. Adams},
  pages     = {2960--2968},
  booktitle = {Proc. of {N}eur{IPS}'12},
  year      = {2012}
}

@inproceedings{song-neurips22a,
  title     = {Monte carlo tree search based variable selection for high dimensional bayesian optimization},
  author    = {L. Song and K. Xue and X. Huang and C. Qian},
  pages     = {28488--28501},
  booktitle = {Proc. of {N}eur{IPS}'22},
  year      = {2022}
}

@inproceedings{srinivas-icml10a,
  title     = {Gaussian Process Optimization in the Bandit Setting: No Regret and Experimental Design},
  author    = {N. Srinivas and A. Krause and S. Kakade and M. Seeger},
  pages     = {1015--1022},
  booktitle = {Proc. of {ICML}'10},
  year      = {2010}
}

@article{swersky-arxiv14a,
  title   = {Freeze-Thaw {B}ayesian Optimization},
  author  = {K. Swersky and J. Snoek and R. Adams},
  year    = {2014},
  journal = {arXiv:1406.3896 {[stats.ML]}}
}

@inproceedings{theodorakopoulos-ecai24a,
  title     = {Hyperparameter Importance Analysis for Multi-Objective {AutoML}},
  author    = {D. Theodorakopoulos and F. Stahl and M. Lindauer},
  pages     = {1100--1107},
  booktitle = {Proc. of {ECAI}'24},
  year      = {2024}
}

@article{tornede-jair23a,
  title   = {Towards Green {Automated Machine Learning}: Status Quo and Future Directions},
  author  = {T. Tornede and A. Tornede and J. Hanselle and F. Mohr and M. Wever and E. H{\"{u}}llermeier},
  year    = {2023},
  journal = {Journal of Artificial Intelligence Research},
  volume  = {77},
  pages   = {427--457}
}

@article{wang-jair16a,
  title   = {Bayesian Optimization in a Billion Dimensions via Random Embeddings},
  author  = {Z. Wang and F. Hutter and M. Zoghi and D. Matheson and N. de~Freitas},
  year    = {2016},
  journal = {Journal of Artificial Intelligence Research},
  volume  = {55},
  pages   = {361--387}
}

@article{wang-kbs21a,
  title   = {{ExperienceThinking}: Constrained Hyperparameter Optimization Based on Knowledge and Pruning},
  journal = {Knowledge-Based Systems},
  volume  = {223},
  pages   = {106602},
  year    = {2021},
  author  = {C. Wang and H. Wang and C. Zhou and H. Chen}
}

@inproceedings{wang-pmlr25a,
  title     = {Grouped Sequential Optimization Strategy - The Application of Hyperparameter Importance Assessment in Deep Learning},
  author    = {R. Wang and I. Nabney and M. Golbabaee},
  year      = {2025},
  booktitle = {Proceedings of Machine Learning Research}
}

@article{weerts-jair24a,
  title   = {Can Fairness be Automated? {Guidelines} and Opportunities for Fairness-aware {AutoML}},
  author  = {H. Weerts and F. Pfisterer and M. Feurer and K. Eggensperger and E. Bergman and N. Awad and J. Vanschoren and M. Pechenizkiy and B. Bischl and F. Hutter},
  year    = {2024},
  volume  = {79},
  journal = {Journal of Artificial Intelligence Research},
  pages   = {639--677}
}

@inproceedings{wever-aaai26a,
  author    = {M. Wever and M. Muschalik and F. Fumagalli and M. Lindauer},
  title     = {{HyperSHAP}: {Shapley} Values and Interactions for Hyperparameter Importance},
  booktitle = {Proc. of {AAAI}'26},
  year      = {2026}
}

@article{wilcoxon-biometrics45a,
  author    = {Wilcoxon, Frank},
  title     = {Individual comparisons by ranking methods},
  journal   = {Biometrics Bulletin},
  volume    = {1},
  number    = {6},
  pages     = {80--83},
  year      = {1945},
  publisher = {JSTOR}
}

@inproceedings{wistuba-pkdd15a,
  title     = {Hyperparameter Search Space Pruning - {A} New Component for Sequential Model-Based {H}yperparameter {O}ptimization},
  author    = {M. Wistuba and N. Schilling and L. Schmidt{-}Thieme},
  pages     = {104--119},
  booktitle = {Proc. of {ECML}/{PKDD}'15},
  year      = {2015}
}

@article{zhang-tec07a,
  author  = {Q. Zhang and H. Li},
  title   = {{MOEA/D}: A Multiobjective Evolutionary Algorithm Based on Decomposition},
  journal = {{IEEE} Trans. Evol. Comput.},
  volume  = {11},
  number  = {6},
  pages   = {712--731},
  year    = {2007}
}

@inproceedings{zhao-iclr21a,
  title     = {Multi-objective Optimization by Learning Space Partition},
  author    = {Y. Zhao and L. Wang and K. Yang and T. Zhang and T. Guo and Y. Tian},
  year      = {2021},
  booktitle = {International Conference on Learning Representations}
}

@article{zimmer-tpami21a,
  title   = {{Auto-Pytorch}: Multi-Fidelity MetaLearning for Efficient and Robust {AutoDL}},
  author  = {L. Zimmer and M. Lindauer and F. Hutter},
  year    = {2021},
  journal = {IEEE Transactions on Pattern Analysis and Machine Intelligence},
  volume  = {43},
  pages   = {3079--3090},
  issue   = {9}
}

@article{zitzler-ec00a,
  title     = {Comparison of multiobjective evolutionary algorithms: Empirical results},
  author    = {E. Zitzler and K. Deb and L. Thiele},
  year      = {2000},
  journal   = {Evolutionary computation},
  publisher = {MIT Press},
  volume    = {8},
  number    = {2},
  pages     = {173--195}
}
\bibliographystyle{preprint}

\newpage
\appendix

\section{Selection of \yahpo Tasks} \label{app:task_selection}
We selected the tasks for \texttt{LCBench} and \texttt{rbv2\_ranger} by randomly drawing 20 different weight vectors and applying HyperSHAP to the surrogate models provided by YAHPO-Gym \citep{pfisterer-automl22a}, combining multiple objectives as originally defined by \cite{knowls-evoco06a}. For \texttt{rbv2\_ranger}, we consider accuracy and memory as objectives; for \texttt{LCBench}, we run tasks based on validation accuracy and time. To this end, we assessed the tunability of these weightings relative to the default configuration as a baseline. Based on these results, we apply our hyperparameter selection method to identify a subset of hyperparameters to tune for a specific weight vector. Afterward, we selected the top 13 tasks for \texttt{LCBench} and \texttt{rbv2\_ranger} that maximize the number of distinct hyperparameter subsets across different weight vectors. Additionally, we paid attention to the fact that the subsets of hyperparameters actually differ across different weightings. Although these criteria may favor HPI-aware methods, tasks were fixed prior to all experiments.

Additionally, we selected 10 tasks from \texttt{LCBench} and 10 from \texttt{rbv2\_ranger} at random with random seed 0, excluding the tasks already selected previously.

\section{Hardware Setup}\label{app:hardware}
The computations were conducted on a high-performance computer with nodes equipped with 2$\times$ AMD Milan 7763 ($2\times 64$ cores) and 256 GiB of RAM each, running Red Hat Enterprise Linux Ootpa and Slurm. Individual runs took between 1 and 24 CPU-hours depending on the optimizer and benchmark, with 1 core and 16GB RAM allocated per run. 
Our main results were evaluated on 25 tasks $\times$ 6 optimizers $\times$ 10 seeds (1,500 runs), plus an additional 630 runs for the remaining experiments, totaling to 2,130 runs and an estimated 25,560 CPU-hours (at $\sim$12 CPU-hours/run on average).

\section{Number of Trials per Task} \label{app:n_trials}
In the \yahpo paper \citep{pfisterer-automl22a}, the number of trials is defined as $\text{number of trials} = 20 + 40 \sqrt{D}$ with $D$ being the number of dimensions.
For our \texttt{LCBench}, \texttt{rbv2\_ranger}, and \texttt{PyMOO}, we used five times as many trials.
We ran the \pymoo tasks \texttt{ZDT1-3} only three times as many trials, since the optimizer overhead per iteration is higher for these tasks. Table~\ref{tab:n_trials} lists the resulting number of trials per task (rounded up to the nearest multiple of 5).
\begin{table}[h!]
    \centering
    \begin{tabular}{lll}
        \toprule
        \textbf{Benchmark} & \textbf{Task/Scenario} & \textbf{Trials} \\
        \midrule
        \pymoo & \texttt{ZDT1-3} & 720  \\
        \pymoo & \texttt{ZDT4, ZDT6} & 735  \\
        \yahpo & \texttt{LCBench} & 630  \\
        \yahpo & \texttt{rbv2\_ranger} & 670  \\
        \bottomrule
    \end{tabular}
    \caption{Number of trials per task.}
    \label{tab:n_trials}
\end{table}

\section{Ablation Study Variants}\label{app:ablation}
\begin{figure}[h!]
    \centering
    \includegraphics[width=0.8\linewidth]{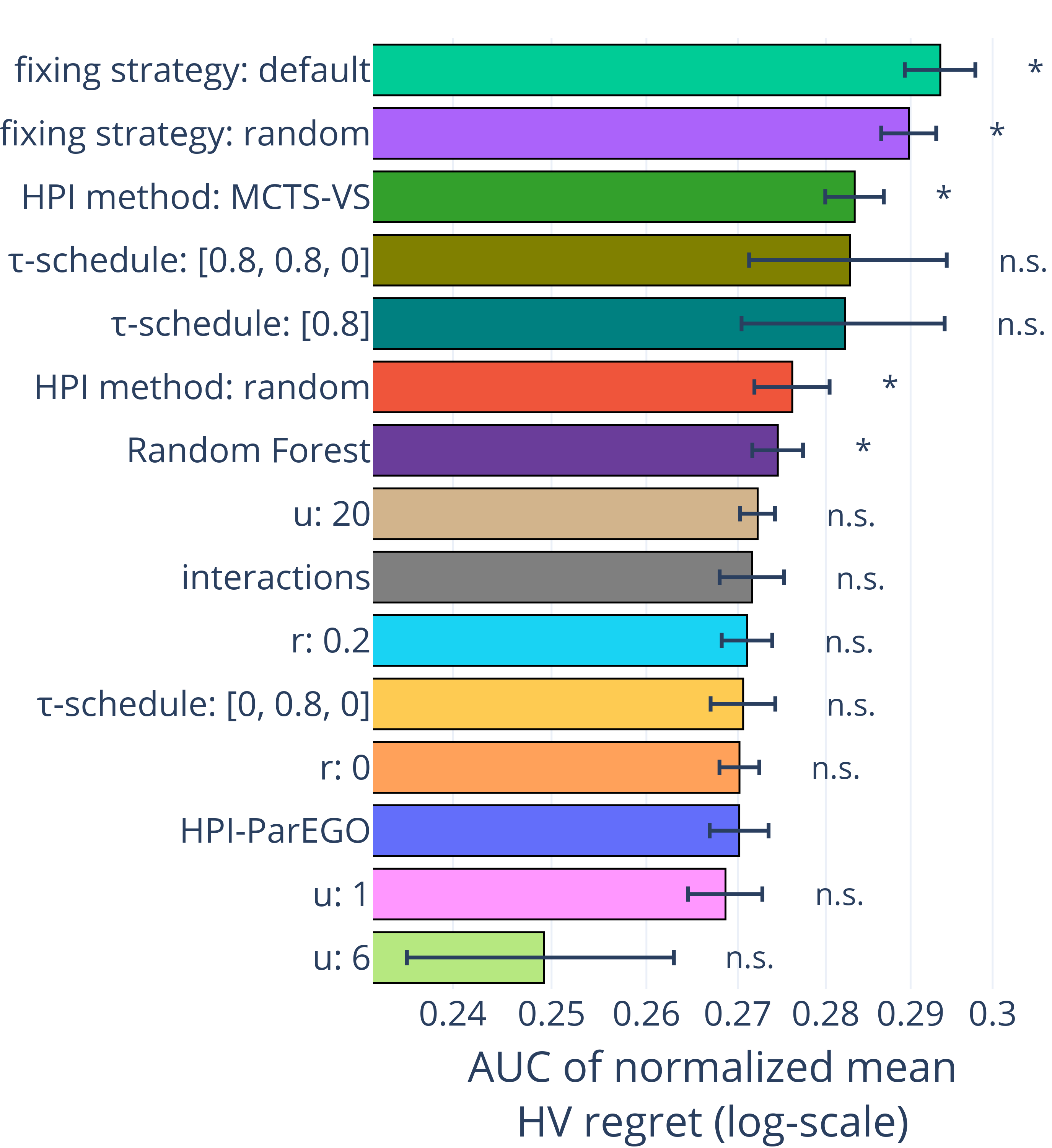}
    \caption{AUC of HV regret for ablating over different algorithm components.}
    \label{fig:ablation_hv}
\end{figure}

In an ablation study, we tried variants of our algorithm. All options are shown in Table~\ref{tab:ablation} in Section~\ref{sec:ablation_study} and will be described here in more detail. All comparisons were performed with the HPI-ParEGO optimizer setup (HPI method: HyperSHAP, fixing strategy: incumbent, $\tau$-schedule: initialization-0.8, $r$=0.1, $u$=10, no interactions, HyperSHAP surrogate: GP). For the ablation experiments, we used three tasks from \texttt{LCBench} and three tasks from \texttt{rbv2\_ranger}, which were not part of the final evaluation.
The summary of the results for HV regret is shown in Figure~\ref{fig:ablation_hv}.

\subsection{HPI Implementation Details}
To integrate HyperSHAP, we use the HyperSHAP library \citep{wever-aaai26a} version 0.0.8, which is available as a PyPI package and on GitHub (\url{https://github.com/automl/hypershap}). We use the tunability estimation with the index set to Shapley values of order 1: for each coalition of hyperparameters, HyperSHAP simulates HPO by sampling 2,000 configurations in which only the coalition varies from the reference configuration, and evaluates them with a surrogate model.
We enable fast mode, which simulates HPO only for coalitions of at most three hyperparameters. For more than 14 hyperparameters, the Shapley values are approximated with a budget of 10,000 coalitions.
We use another surrogate model for HyperSHAP than for hyperparameter optimization. We fit a Gaussian Process (GP) to the observed costs, scalarized with the current ParEGO weights. We do not use the BO surrogate model because it is a random forest, and GPs are better than random forests at capturing small differences around the incumbent. We use SMAC's BlackBox facade implementation of the GP (Matérn kernel, Hamming kernel for categorical hyperparameters) and refit it at every optimization step.

\newpage
\subsection{Comparison of the Hyperparameter Selection Methods}\label{app:random_hpi}
\begin{figure}[h!]
    \centering
    \begin{subfigure}[h]{\textwidth}
        \includegraphics[width=\columnwidth]{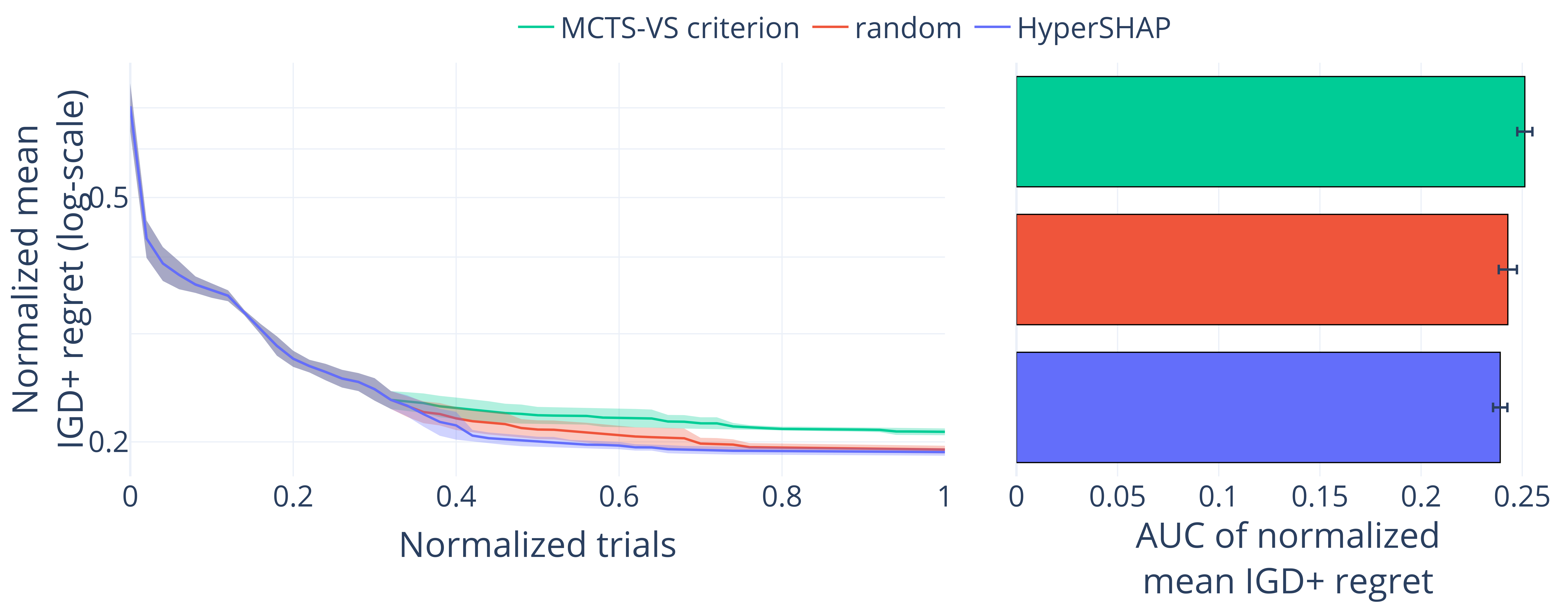}
        \caption{IGD+ regret.}
    \end{subfigure}%
    
    \begin{subfigure}[h]{\textwidth}
        \includegraphics[width=\columnwidth]{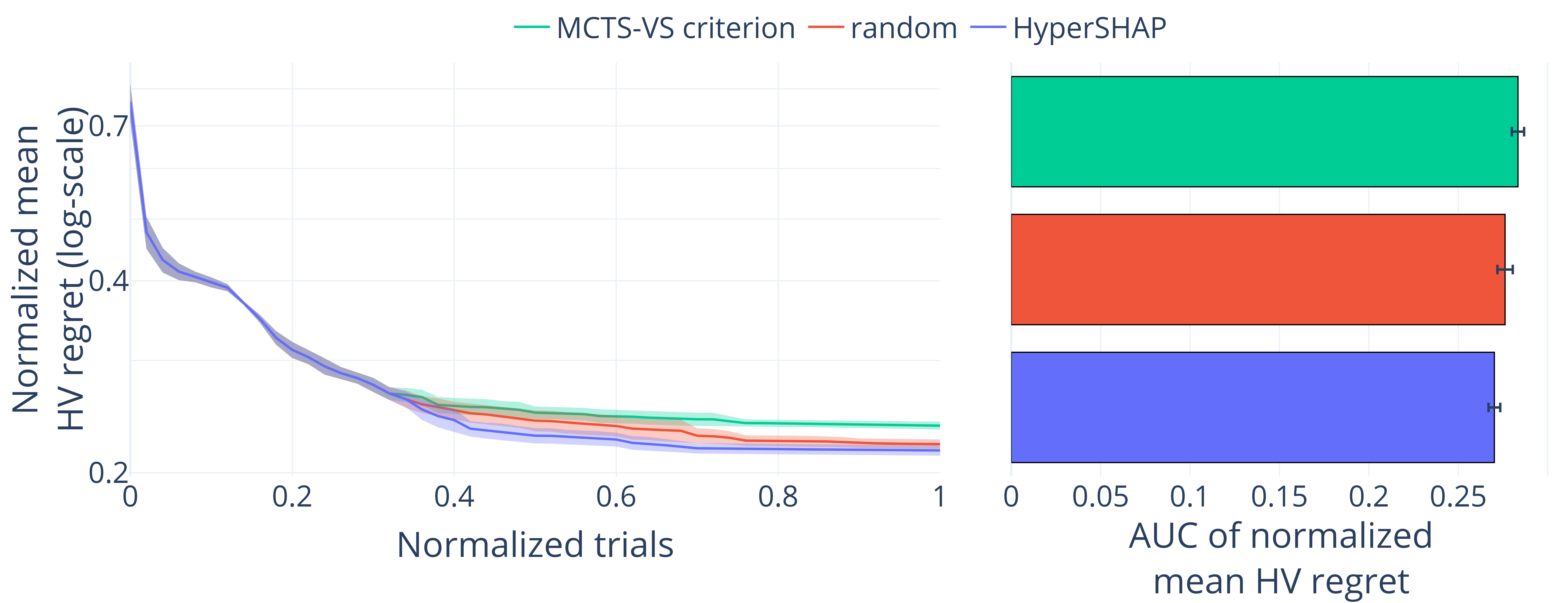}
        \caption{HV regret.}
    \end{subfigure}
    \caption{Ablation results for comparing subsets selected randomly, using the MCTS-VS criterion, or using HyperSHAP (HPI-ParEGO in blue).}
\label{fig:random_vs_hypershap}
\end{figure}

To investigate whether the performance gains of HPI-ParEGO are primarily driven by dimensionality reduction, we compare random selection of a set of important hyperparameters with HPI-based selection using HyperSHAP. The size of the random subset is determined by the threshold: $(1-\tau) \times |\Lambda|$, which is rounded up.
The third hyperparameter selection method that we try is the criterion used in MCTS-VS~\citep{song-neurips22a}.
It scores each hyperparameter by the mean scalarized cost of all evaluated configurations in which it was tuned, i.e., not fixed by the reduction, relative to the mean over all evaluated configurations.
Since ParEGO changes the scalarization, we re-scalarize all observed costs with the current weights before computing the scores, and then select hyperparameters with the same threshold rule as for HyperSHAP.
Unlike MCTS-VS, we do not use its tree search over subsets of hyperparameters, but only its importance criterion.
As shown in Figure~\ref{fig:random_vs_hypershap}, HyperSHAP-based selection consistently outperforms the other methods. 

\begin{figure}[h!]
    \centering
    \begin{subfigure}[h]{0.5\textwidth}
        \includegraphics[width=\columnwidth]{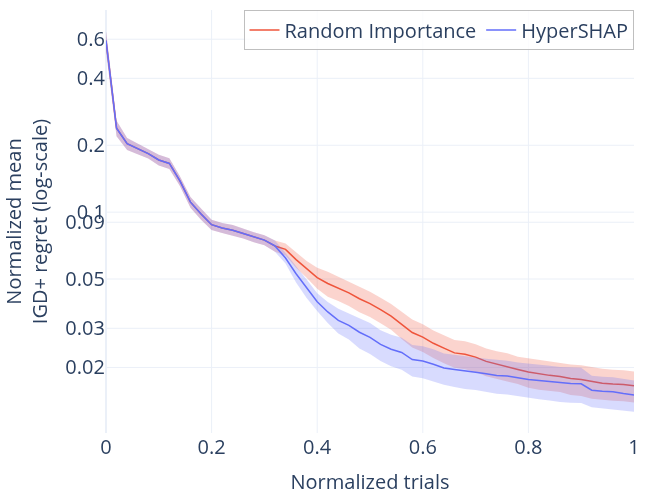}
        \caption{IGD+ regret.}
    \end{subfigure}%
    ~
    \begin{subfigure}[h]{0.5\textwidth}
        \includegraphics[width=\columnwidth]{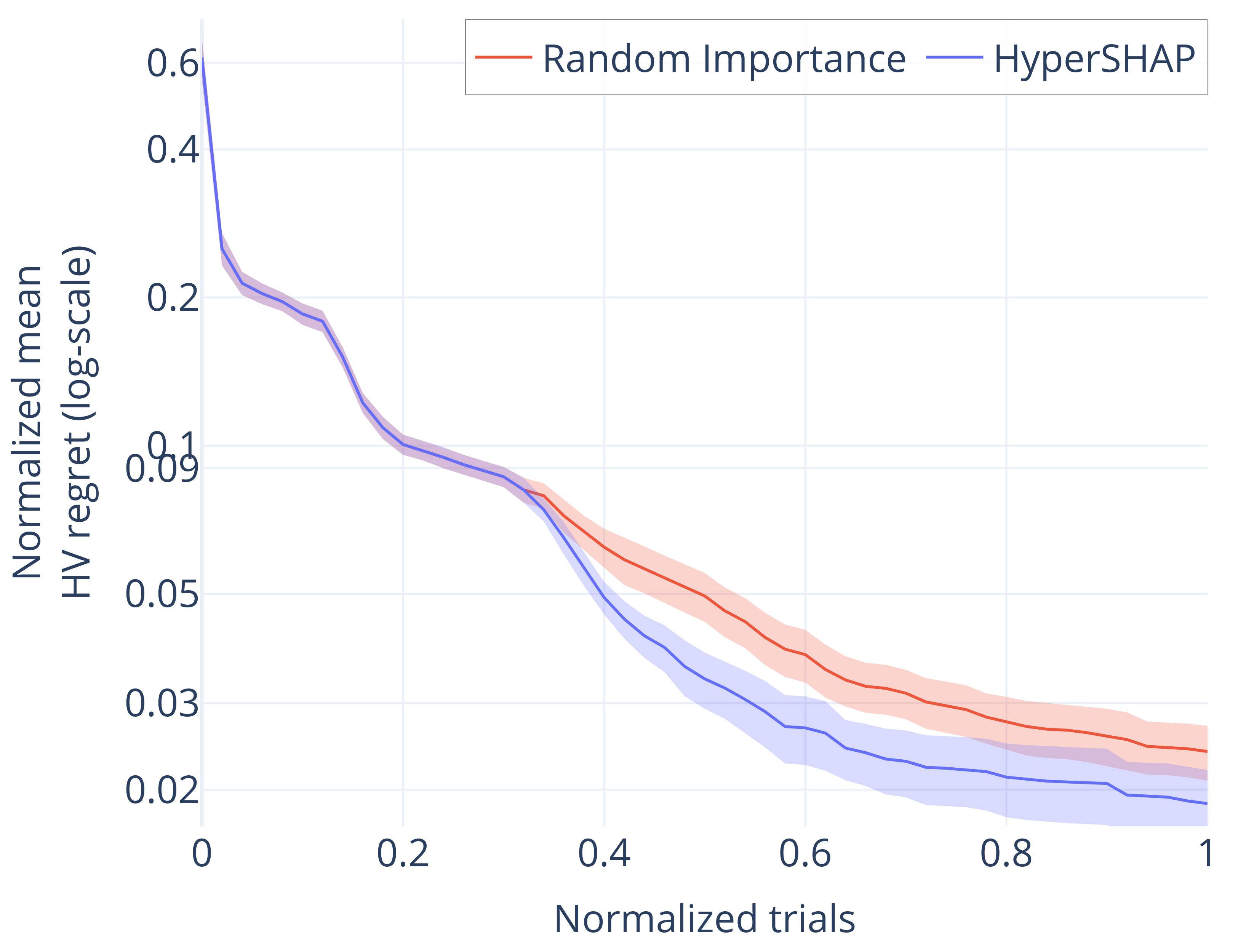}
        \caption{HV regret.}
    \end{subfigure}
    \caption{Ablation results for comparing randomly selected subsets against HPI-ParEGO on \texttt{LCBench}.}
\label{fig:lcbench_random}
\end{figure}

Since the difference to random selection is small, we also show results for that on the 10 selected \texttt{LCBench} tasks of section~\ref{sub:yahpo}.
Figure~\ref{fig:lcbench_random} shows HPI-based selection clearly outperforming random subset selection.
This suggests that HyperSHAP can identify relevant hyperparameters already early in the optimization process, even with limited data.

\newpage
\subsection{Hyperparameter Fixing Strategy}
\begin{figure}[h!]
    \centering
    \begin{subfigure}[h]{\textwidth}
        \includegraphics[width=\columnwidth]{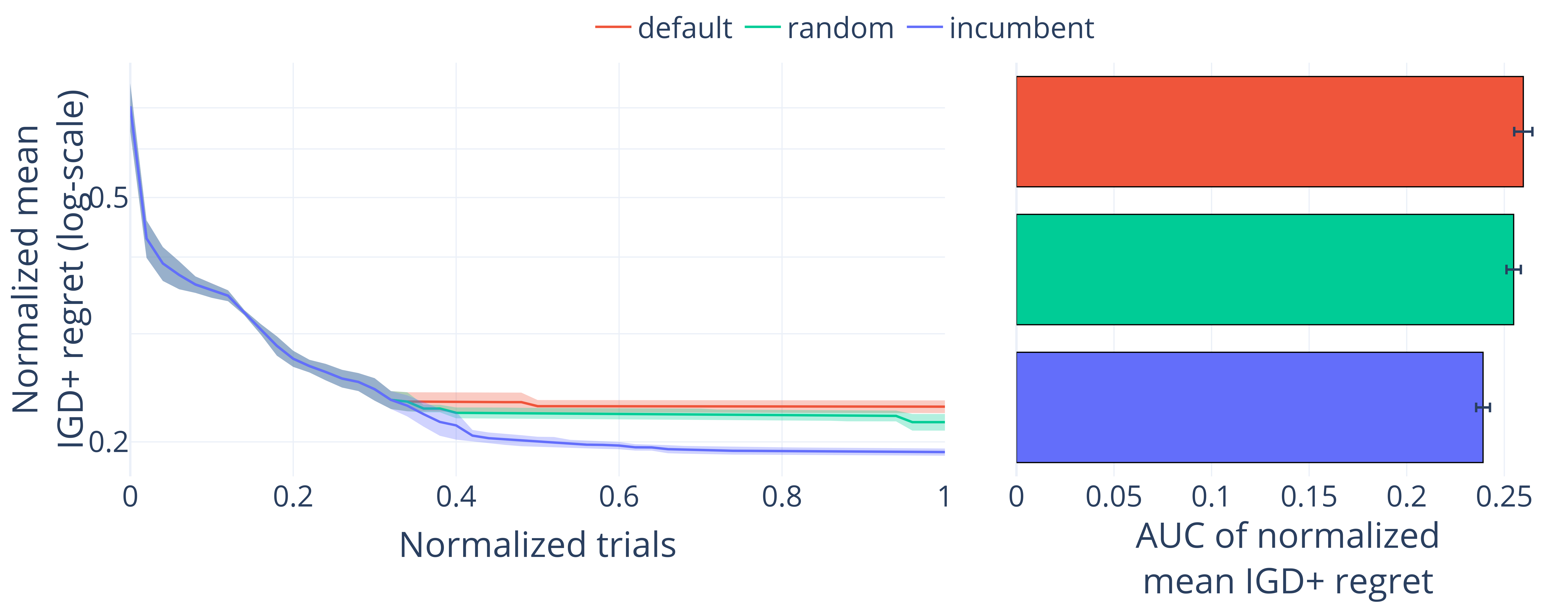}
        \caption{IGD+ regret.}
    \end{subfigure}%
    
    \begin{subfigure}[h]{\textwidth}
        \includegraphics[width=\columnwidth]{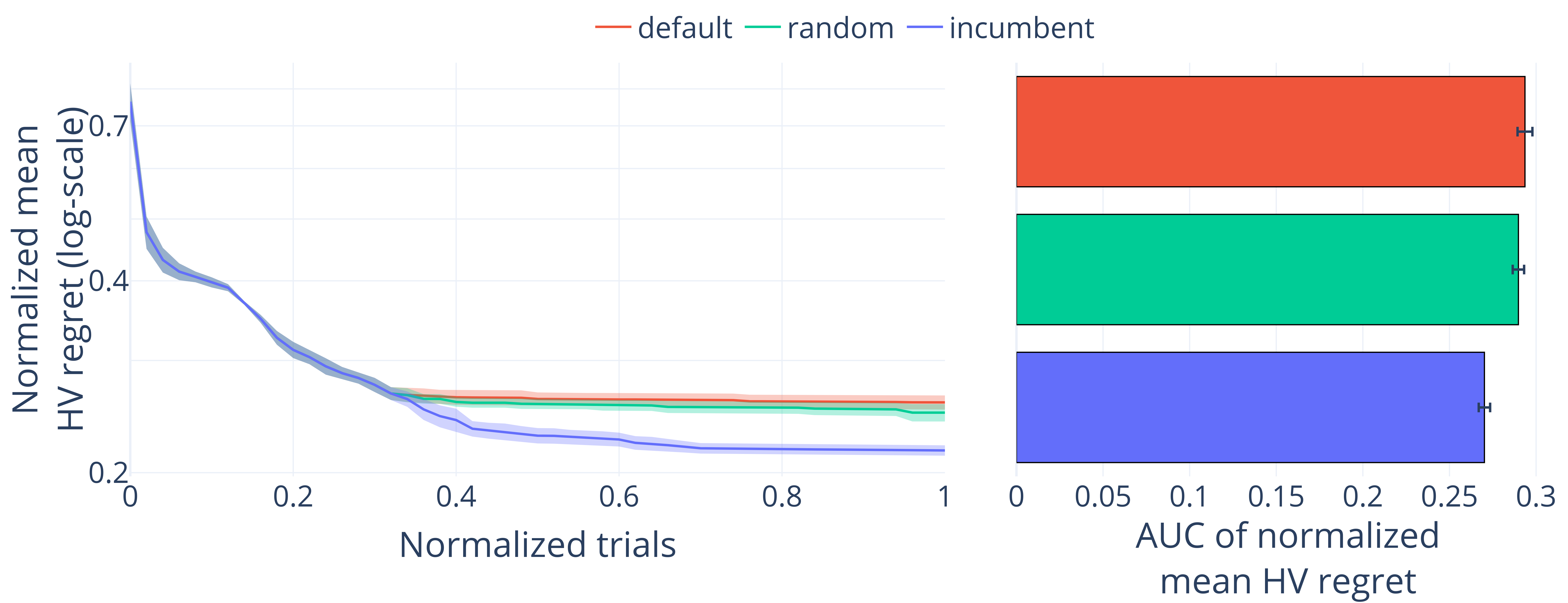}
        \caption{HV regret.}
    \end{subfigure}
    \caption{Ablation results for different constant values to reduce the configuration space (HPI-ParEGO in blue).}
\label{fig:cs_values}
\end{figure}

We tested fixing the constant values of the unimportant hyperparameters to the hyperparameter's default value, the configuration with the current incumbent value in the run history $\mathcal{H}$ for the given scalarization, or a randomly sampled value. Additionally, for HyperSHAP, that choice (incumbent, default, or random) was also used as a reference configuration. 
Figure~\ref{fig:cs_values} displays the difference between the approaches. A large improvement can be seen by using the current incumbent value.

\newpage
\subsection{Threshold $\tau$-Schedule}
Furthermore, we ablated over different $\tau$-schedules. 
The optimization trials are split into three parts.
In each part, the reduction can be either off or on, with a fixed value (0.8 in our implementation). 
The motivation for the initialization phase (no reduction in the first third) is to first learn which are important and improve the surrogate model. 
The motivation behind the consolidation phase (no reduction in the last third) is to allow the optimizer to tune the less important hyperparameters in the end as well. 
For the symmetric threshold, we combine both.

\begin{figure}[h!]
    \centering
    \begin{subfigure}[h]{\textwidth}
        \includegraphics[width=\columnwidth]{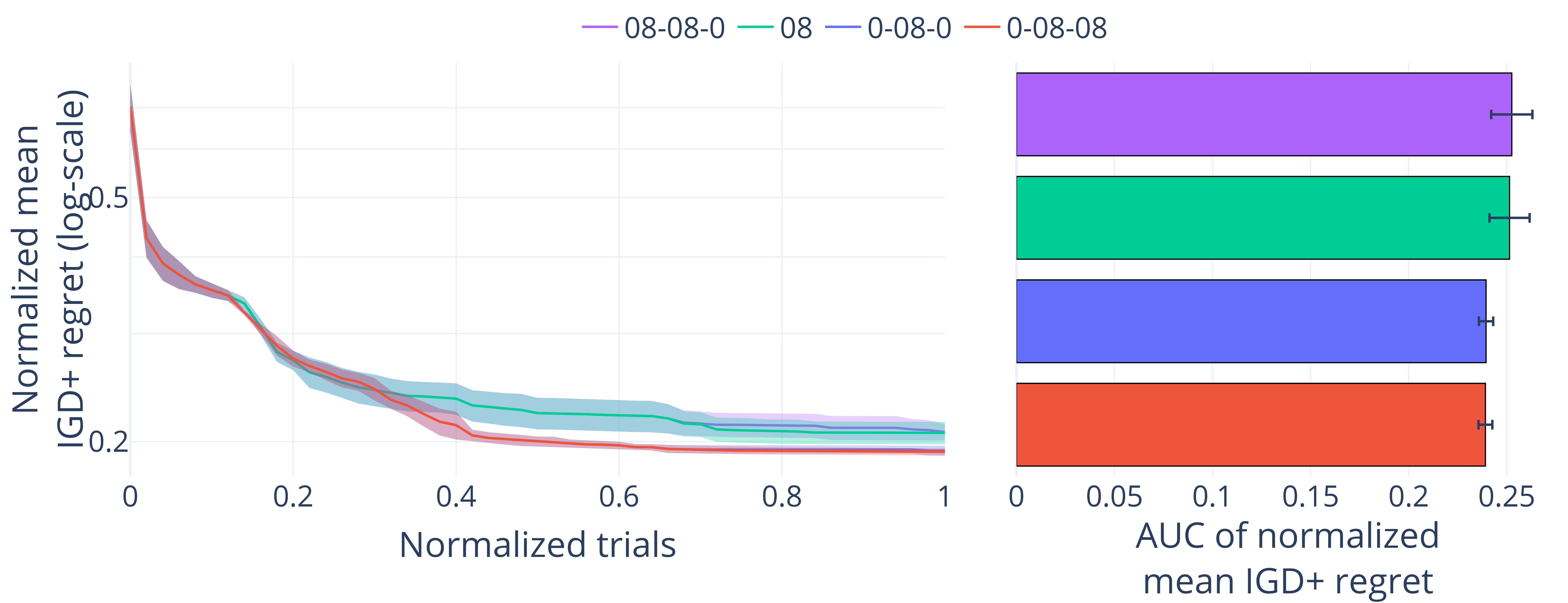}
        \caption{IGD+ regret.}
    \end{subfigure}%
    
    \begin{subfigure}[h]{\textwidth}
        \includegraphics[width=\columnwidth]{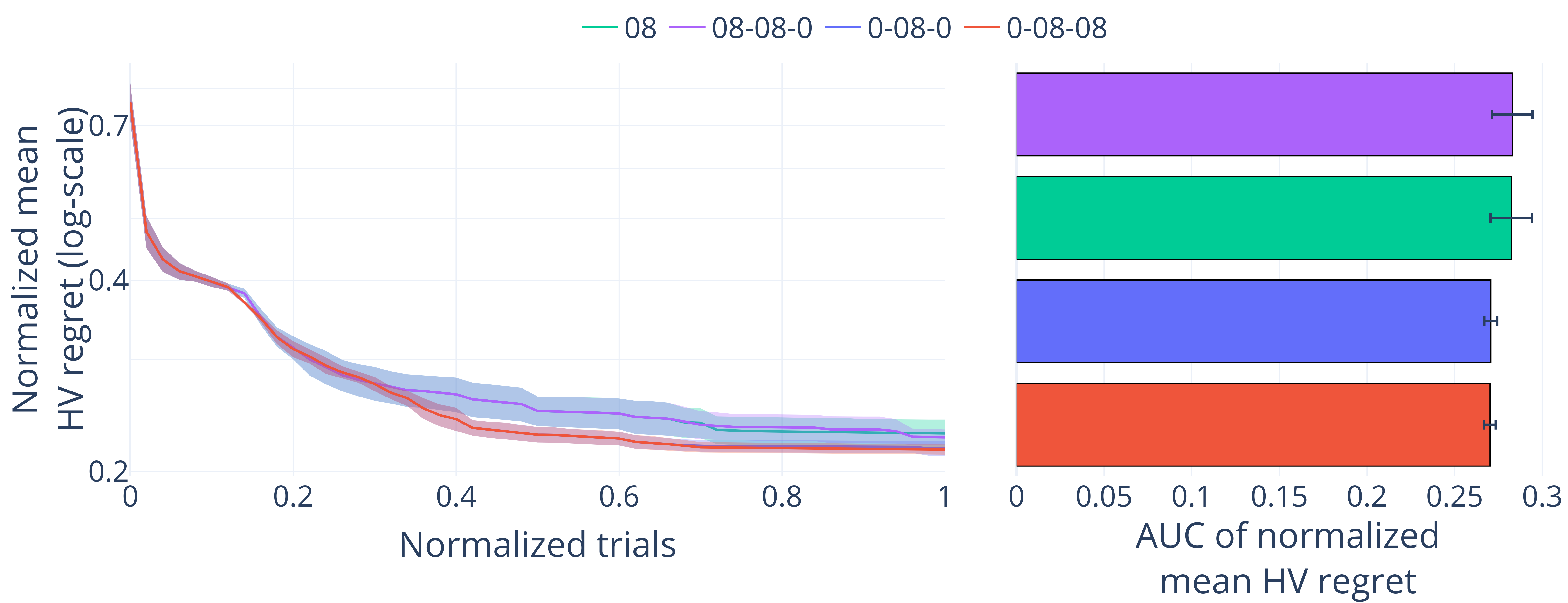}
        \caption{HV regret.}
    \end{subfigure}
    \caption{Ablation results for different thresholds. 0 means no reduction (HPI-ParEGO in blue).}
\label{fig:threshold}
\end{figure}

Figure~\ref{fig:threshold} shows that the two schedules with an initialization phase perform best, while the schedules that reduce the space from the start fall behind.
Allowing enough trials is thus important to estimate HPI well.
In contrast, the consolidation phase has no noticeable effect.
However, for IGD+ regret, the difference of the initialization schedule to the symmetric schedule was statistically significant (see Figure~\ref{fig:ablation}).
We therefore keep the reduction until the end, yielding a simpler schedule with a single switch and exploiting the reduced space for longer. 

\newpage
\subsection{Random Chance for HPI}
\begin{figure}[h!]
    \centering
    \begin{subfigure}[h]{\textwidth}
        \includegraphics[width=\columnwidth]{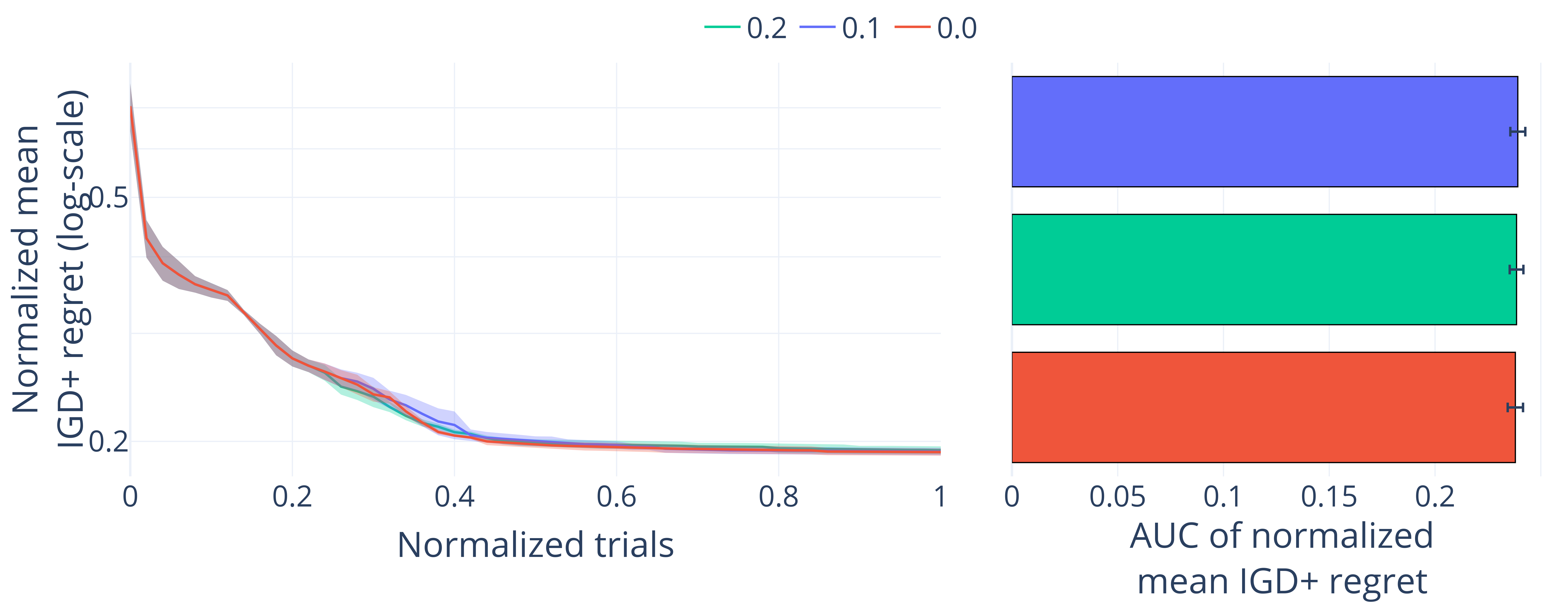}
        \caption{IGD+ regret.}
    \end{subfigure}%
    
    \begin{subfigure}[h]{\textwidth}
        \includegraphics[width=\columnwidth]{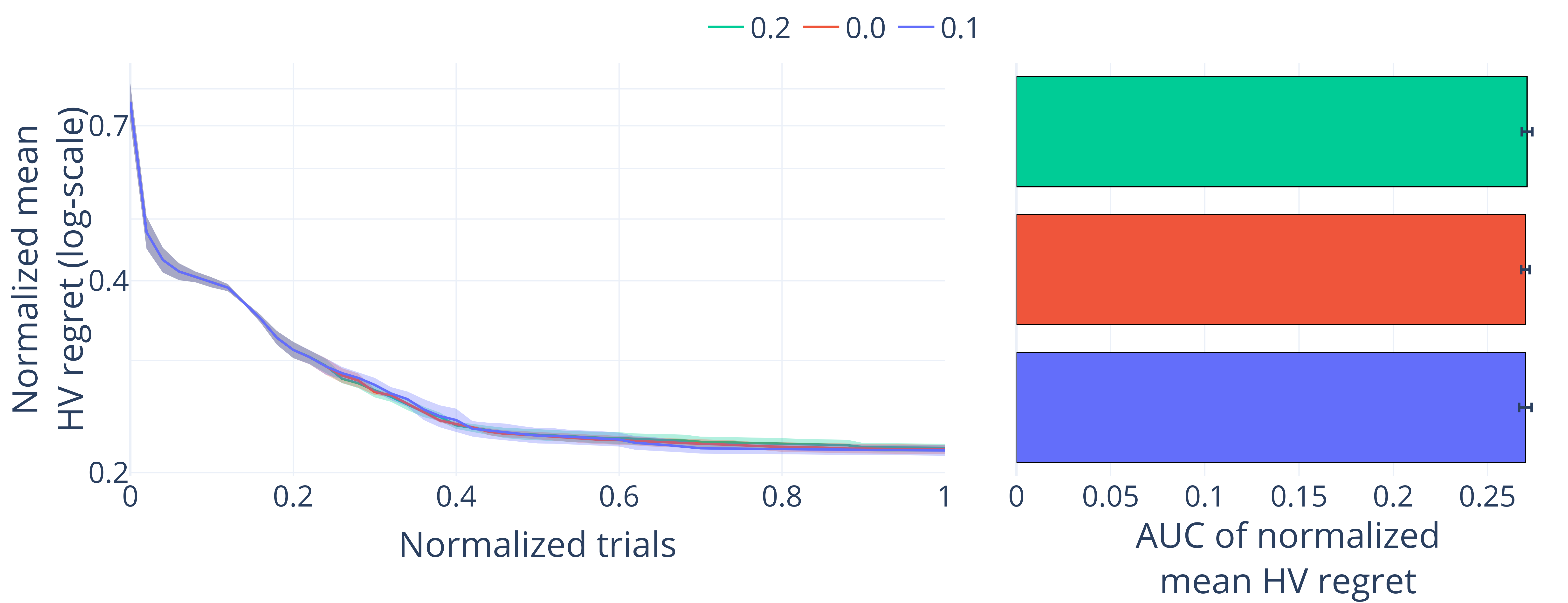}
        \caption{HV regret.}
    \end{subfigure}
    \caption{Ablation results for different random chances $r$ (HPI-ParEGO in blue).}
\label{fig:rnd_prob}
\end{figure}

We also considered different values for the random chance $r$ that a random configuration of the original configuration space will be evaluated. We tested: 0.0, 0.1, 0.2. The results are shown in Figure \ref{fig:rnd_prob}. It can be seen that sometimes using random, full configurations has almost no effect. However, to maintain some exploration, we evaluate a full configuration with a 0.1 probability.

\newpage
\subsection{Number of Iterations for a Weight Update}
\begin{figure}[h!]
    \centering
    \begin{subfigure}[h]{\textwidth}
        \includegraphics[width=\columnwidth]{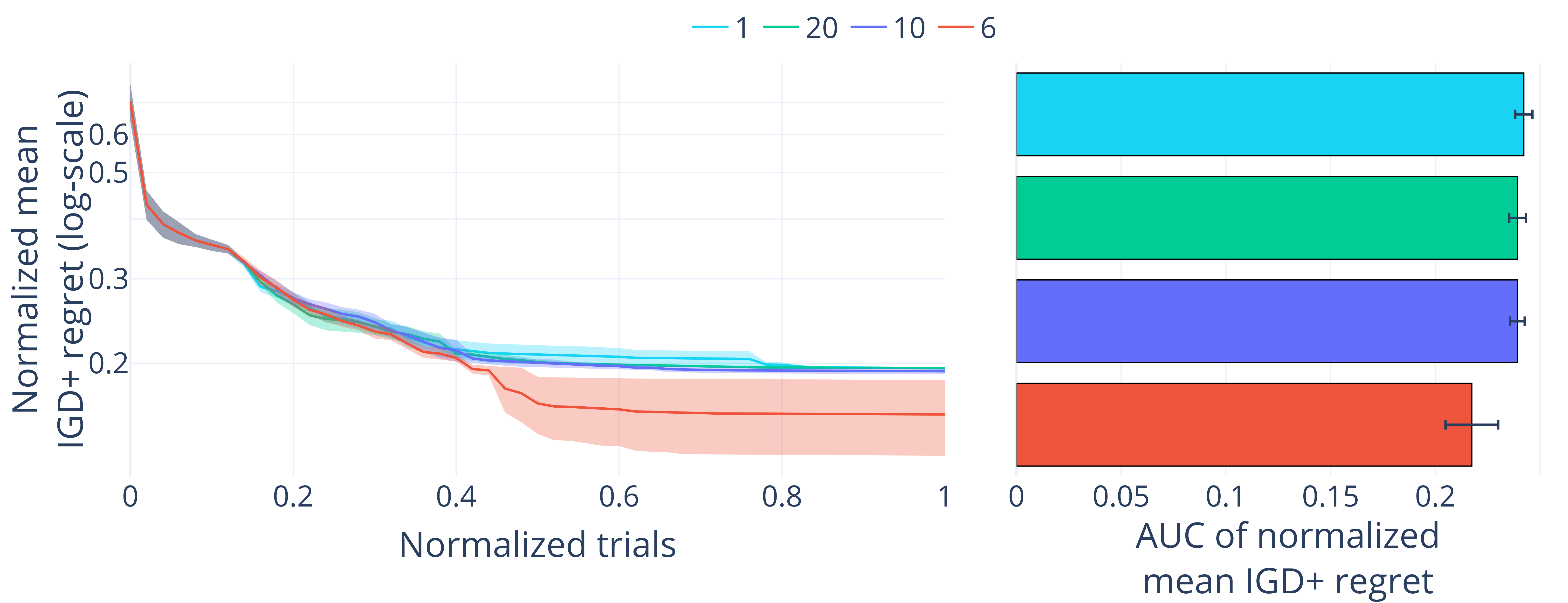}
        \caption{IGD+ regret.}
    \end{subfigure}%
    
    \begin{subfigure}[h]{\textwidth}
        \includegraphics[width=\columnwidth]{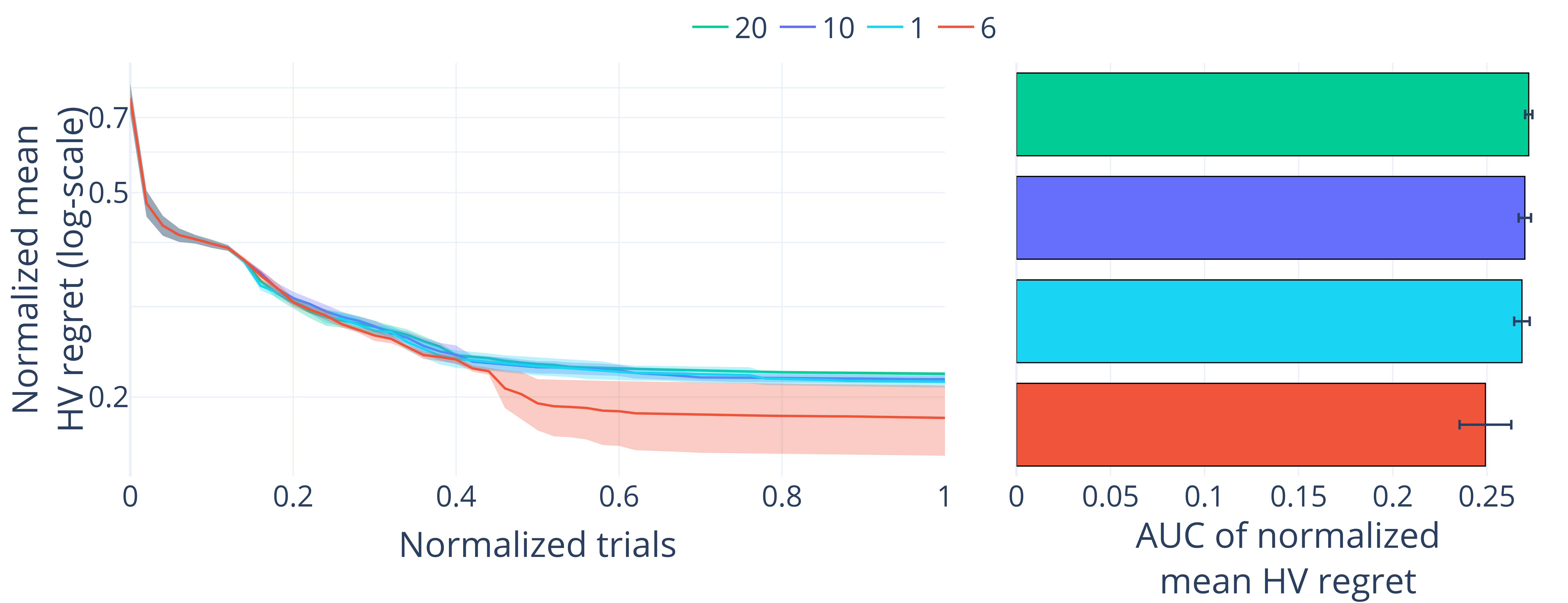}
        \caption{HV regret.}
    \end{subfigure}
    \caption{Ablation results for different numbers of iterations for weight updates $u$ (HPI-ParEGO in blue).}
\label{fig:reweigh}
\end{figure}

\begin{figure}[h!]
    \centering
    \begin{subfigure}[h]{0.5\textwidth}
        \includegraphics[width=\columnwidth]{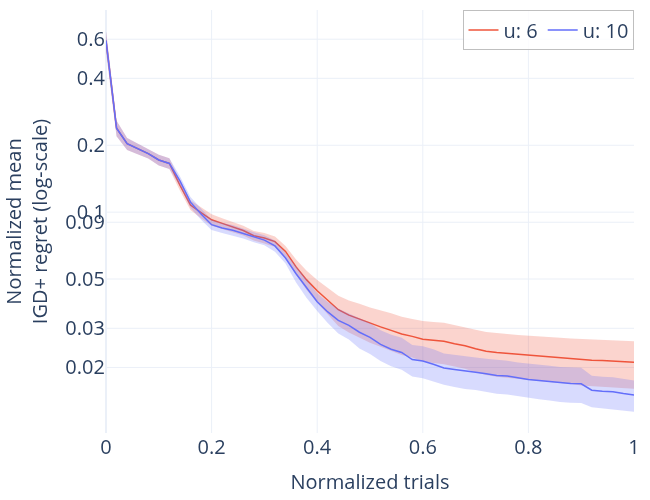}
        \caption{IGD+ regret.}
    \end{subfigure}%
    ~
    \begin{subfigure}[h]{0.5\textwidth}
        \includegraphics[width=\columnwidth]{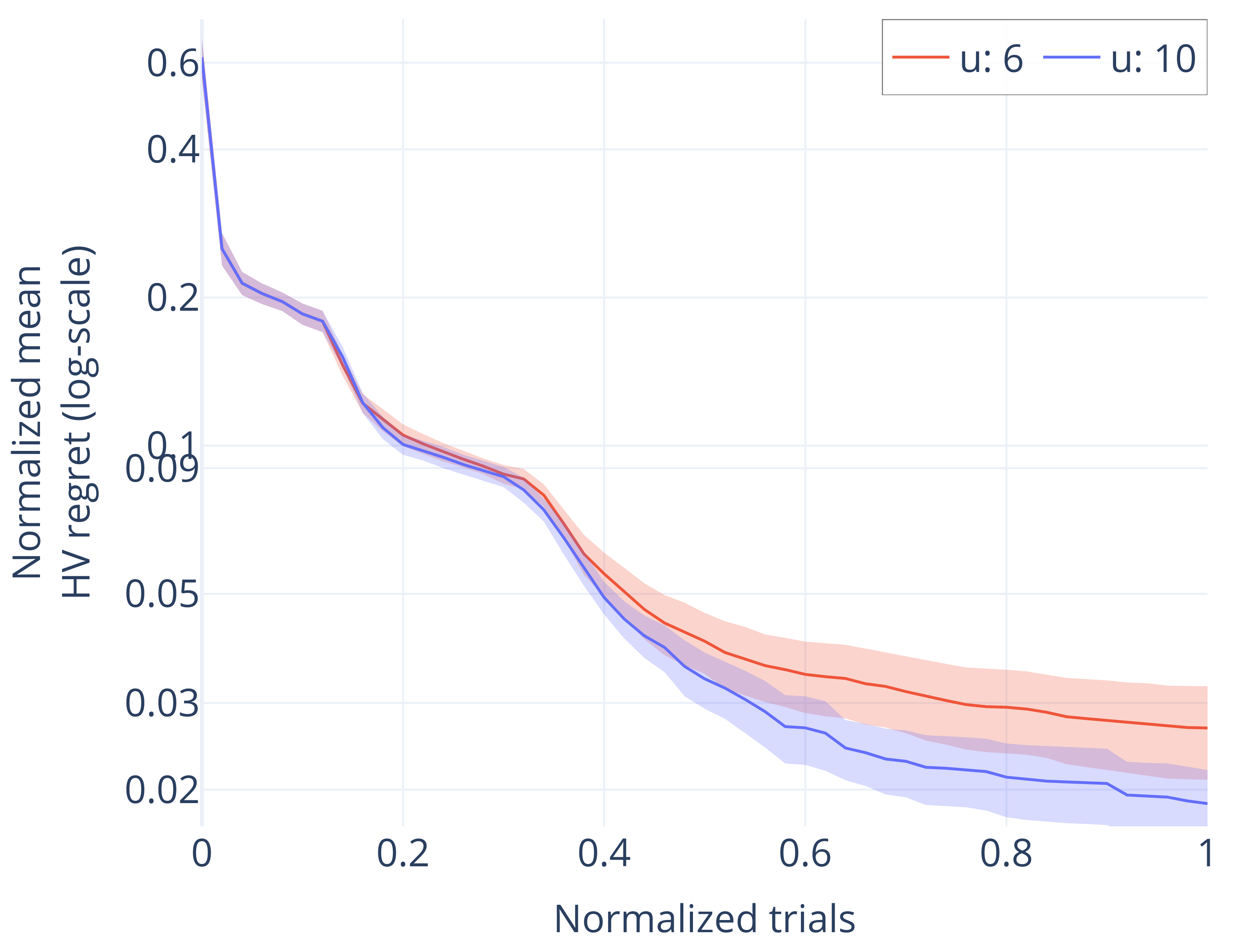}
        \caption{HV regret.}
    \end{subfigure}
    \caption{Results for $u=6$ vs. $u=10$ (HPI-ParEGO) on the selected \texttt{LCBench} tasks.}
\label{fig:u_6}
\end{figure}

The intuition behind increasing the number of iterations for a weight update, compared to the original ParEGO algorithm, is that the optimization process can focus on a scalarization for more iterations and thereby improve the current scalarization. This also allows for leveraging the reduced configuration space several times before it is reset. We tested different values: 1, 6, 10, and 20. Figure \ref{fig:reweigh} shows 6 iterations as the best value, although not significantly (see Figure~\ref{fig:ablation} in Section~\ref{sec:ablation_study}). Therefore, we also ran that variant on the 10 selected \texttt{LCBench} tasks of Section~\ref{sub:yahpo}. 
Figure~\ref{fig:u_6} shows that the advantage of $u=6$ does not hold on the \texttt{LCBench} evaluation data, which has more tasks and seeds.

\newpage
\newpage
\subsection{Interactions}
HyperSHAP at order 1 accounts for interactions by distributing the gain from an interaction among the participating hyperparameters.
However, it only returns importance values for single hyperparameters, not for coalitions.
It can thus happen that from a strongly interacting pair, only one hyperparameter is selected to be tuned.
We therefore tested whether selecting coalitions of interacting hyperparameters directly would improve performance.
For that, we change the order to 2 in HyperSHAP and the tunability index to k-SII. 
When selecting the hyperparameters, we now select coalitions, which can include one or two hyperparameters. 
We again add the selected coalitions until the threshold is satisfied.

\begin{figure}[h!]
    \centering
    \begin{subfigure}[h]{\textwidth}
        \includegraphics[width=\columnwidth]{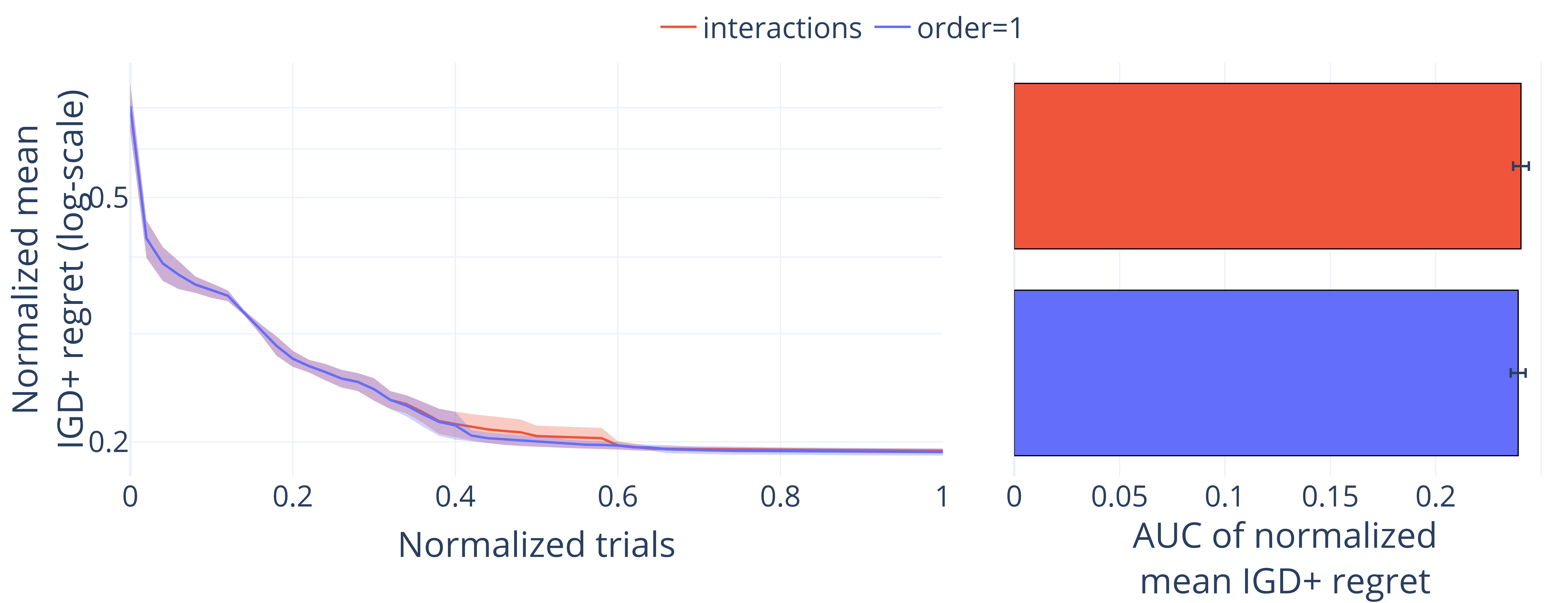}
        \caption{IGD+ regret.}
    \end{subfigure}%
    
    \begin{subfigure}[h]{\textwidth}
        \includegraphics[width=\columnwidth]{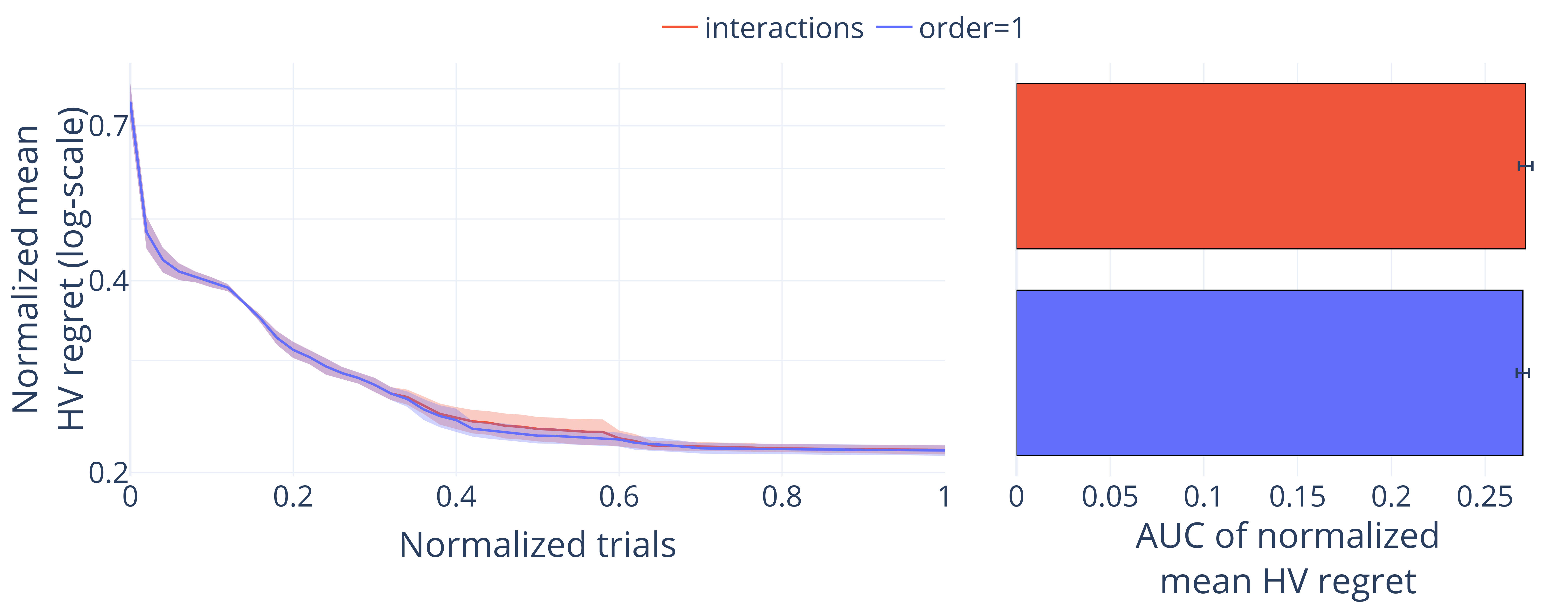}
        \caption{HV regret.}
    \end{subfigure}
    \caption{Ablation results for considering interactions (HPI-ParEGO in blue).}
\label{fig:interactions}
\end{figure}

This approach might work well if the surrogate were trained on several thousand configurations necessary to capture interactions accurately~\citep{wever-aaai26a}.
In our setting, which exploits HPI early in the optimization, the number of trials is insufficient. 
Figure \ref{fig:interactions} shows that including interactions has no significant effect. 
We thus keep the order of 1.

\newpage
\subsection{Surrogate Model for HyperSHAP}
\begin{figure}[h!]
    \centering
    \begin{subfigure}[h]{\textwidth}
        \includegraphics[width=\columnwidth]{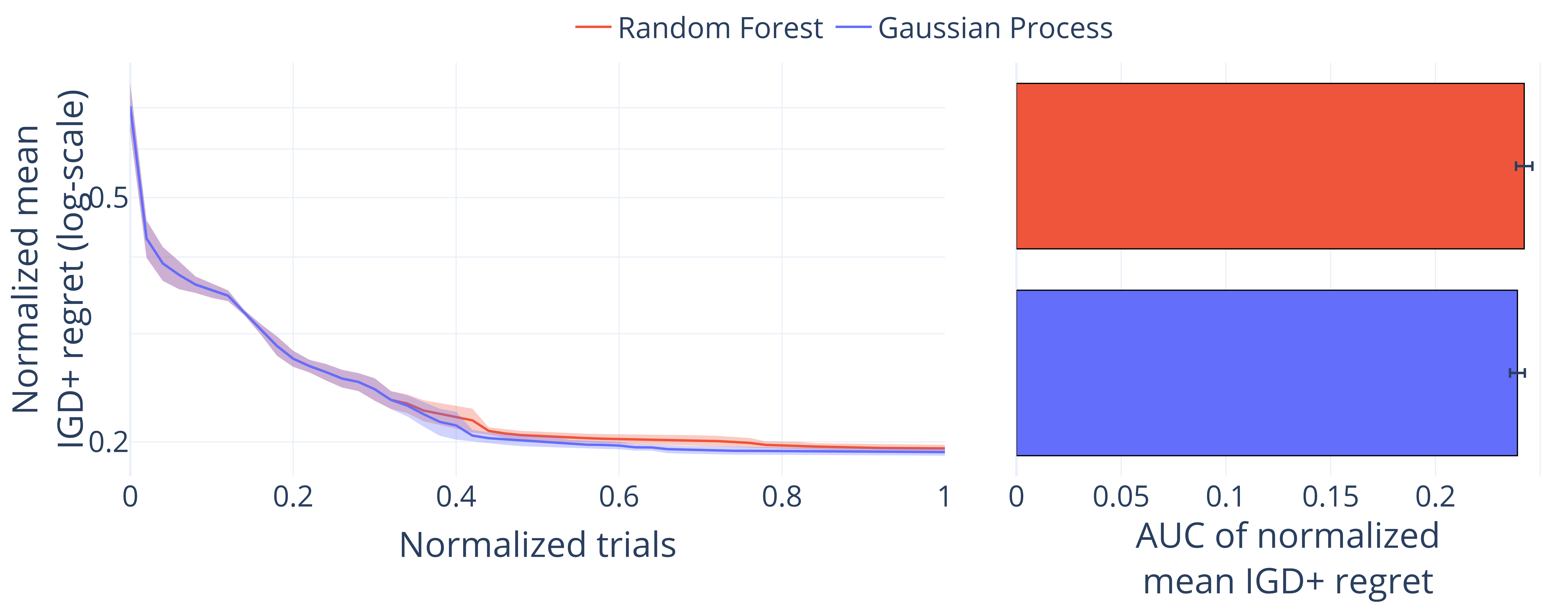}
        \caption{IGD+ regret.}
    \end{subfigure}%
    
    \begin{subfigure}[h]{\textwidth}
        \includegraphics[width=\columnwidth]{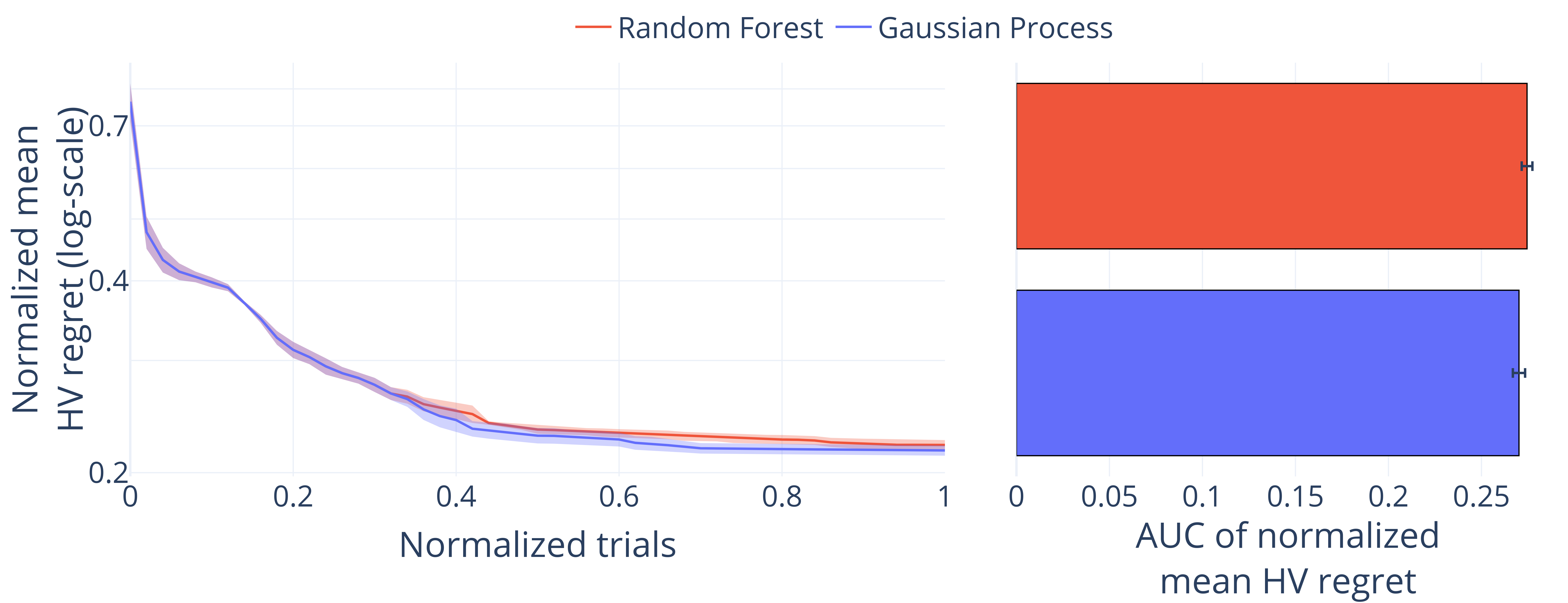}
        \caption{HV regret.}
    \end{subfigure}
    \caption{Ablation results for the surrogate model (HPI-ParEGO in blue).}
\label{fig:surrogate}
\end{figure}

In this experiment, we compare the random forest SMAC uses for optimization with a Gaussian Process to assess tunability using HyperSHAP.
Figure \ref{fig:surrogate} shows slightly better performance of the Gaussian Process in predicting improvement potential around the incumbent. Figure~\ref{fig:ablation} also showed it to be statistically significantly better.

\newpage
\section{Further Results on PyMOO}\label{app:pymoo}
\begin{figure*}[h!]
    \centering
    \begin{subfigure}[t]{0.5\textwidth}
      \centering
      \includegraphics[width=\columnwidth]{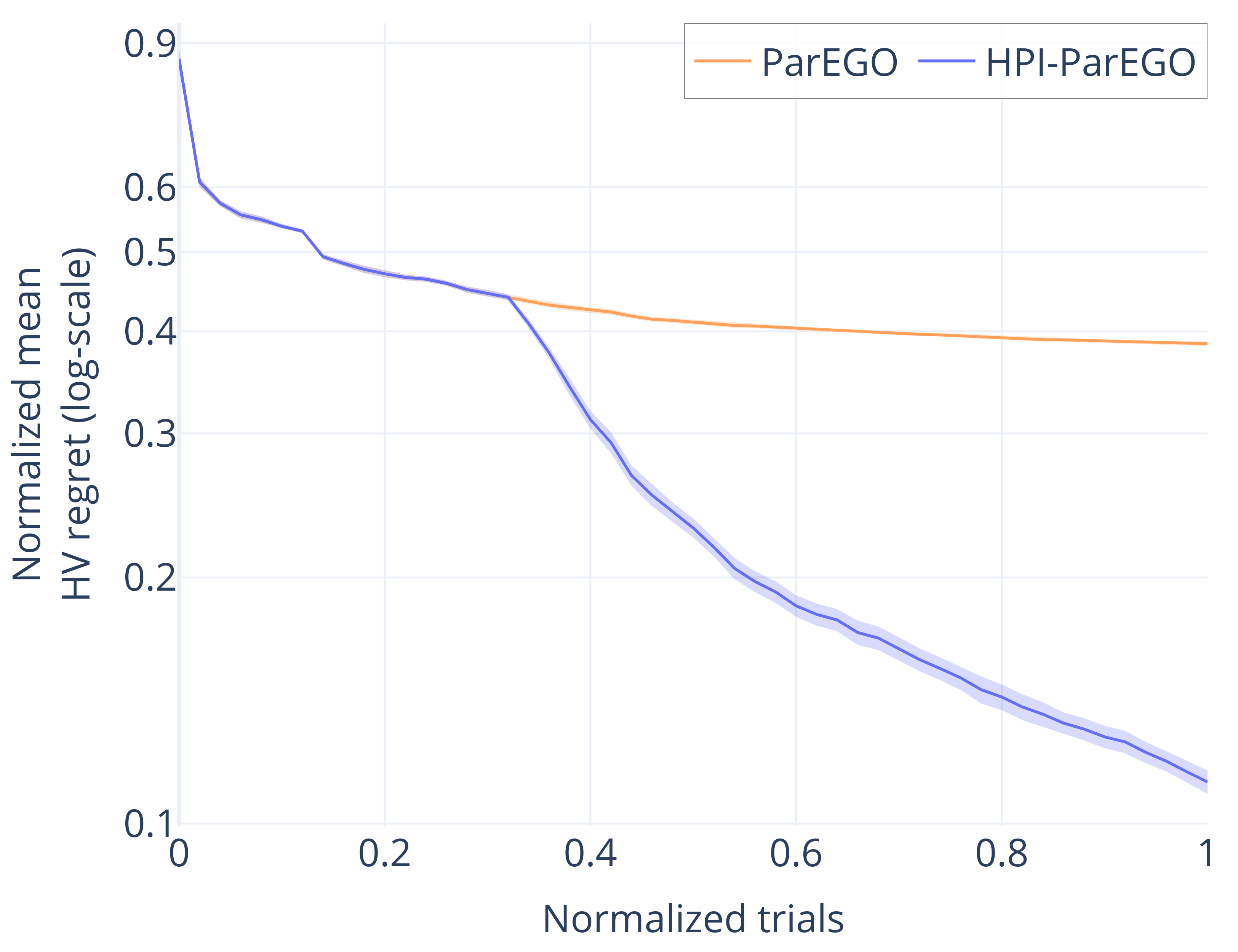}
      \caption{The HPI-ParEGO optimizer compared to the ParEGO optimizer.}
    \end{subfigure}%
    ~ 
    \begin{subfigure}[t]{0.5\textwidth}
          \centering
          \includegraphics[width=\columnwidth]{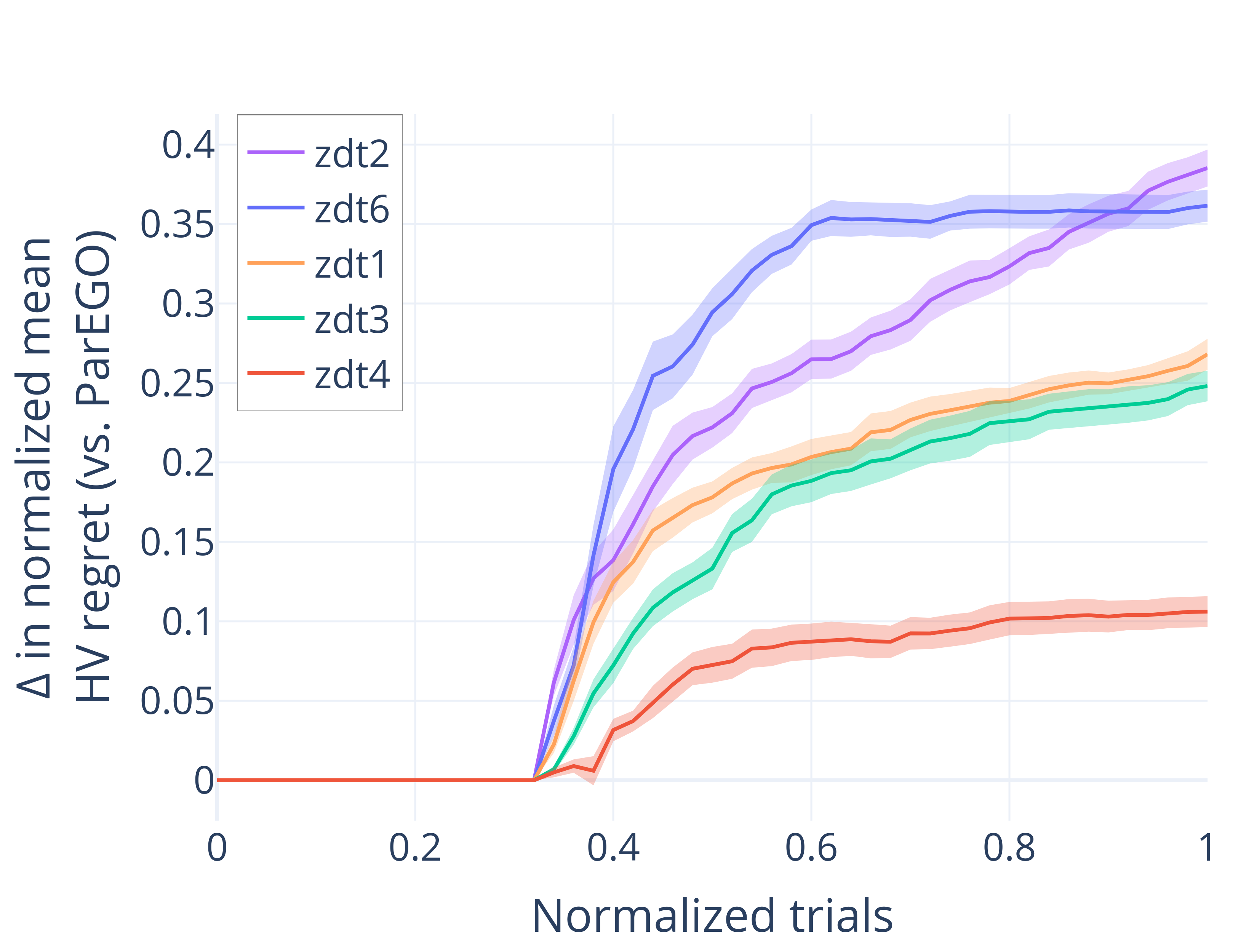}
          \caption{Difference in mean HV regret with standard error between ParEGO and HPI-ParEGO. Values $> 0$ indicate a better performance of HPI-ParEGO.}
    \end{subfigure}
    \caption{Results on the \pymoo tasks.}
    \label{fig:pymoo_hv}
\end{figure*}

Figure~\ref{fig:pymoo_hv} shows the comparison of HPI-ParEGO against the ParEGO optimizer as well as the task difference on the PyMOO tasks for the HV regret indicator.
Similar to the IGD+ results, HPI-ParEGO clearly outperforms ParEGO.

\begin{figure*}[h!]
    \centering
    \begin{subfigure}[t]{0.5\textwidth}
      \centering
      \includegraphics[width=\columnwidth]{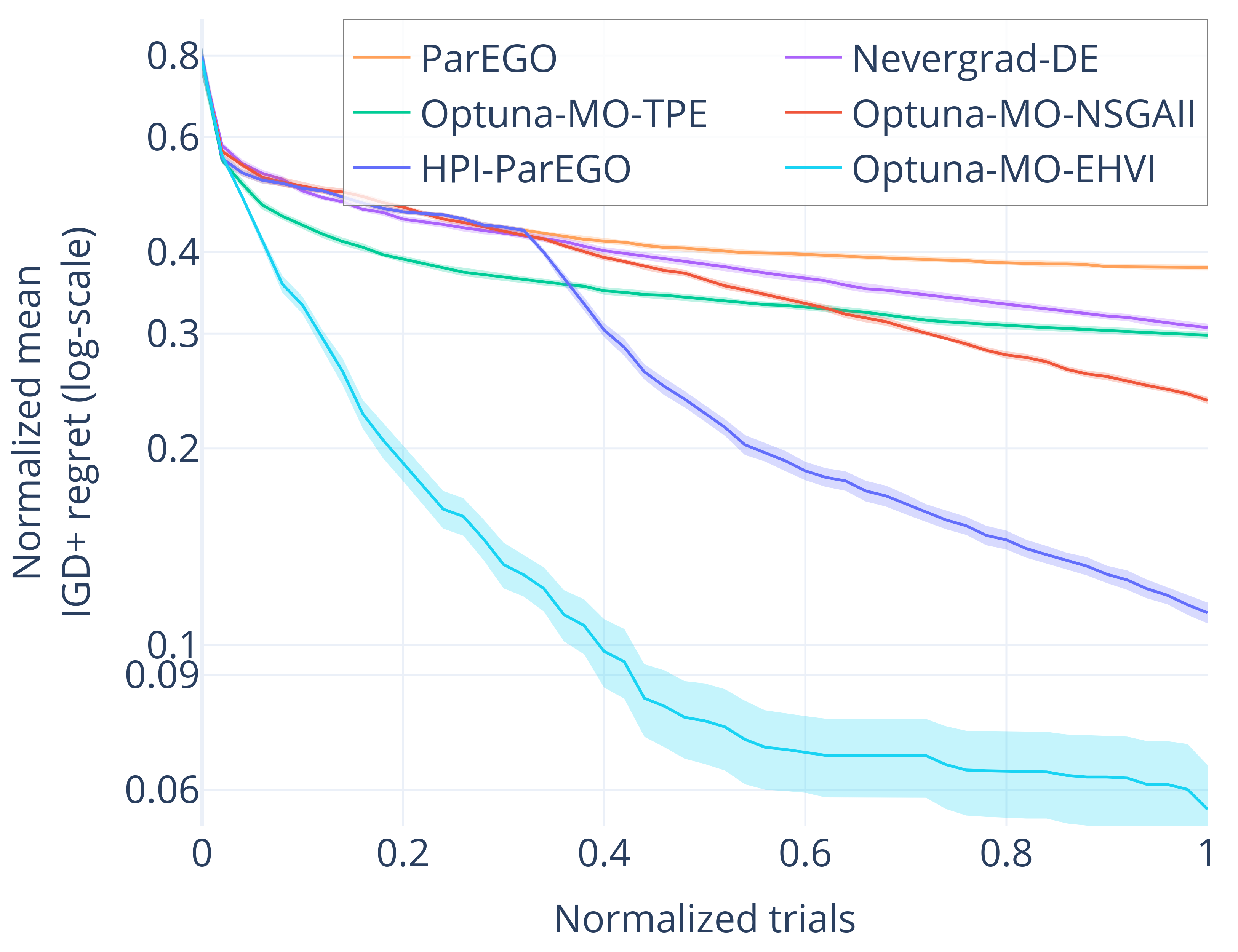}
      \caption{IGD+ regret.}
    \end{subfigure}%
    ~ 
    \begin{subfigure}[t]{0.5\textwidth}
          \centering
          \includegraphics[width=\columnwidth]{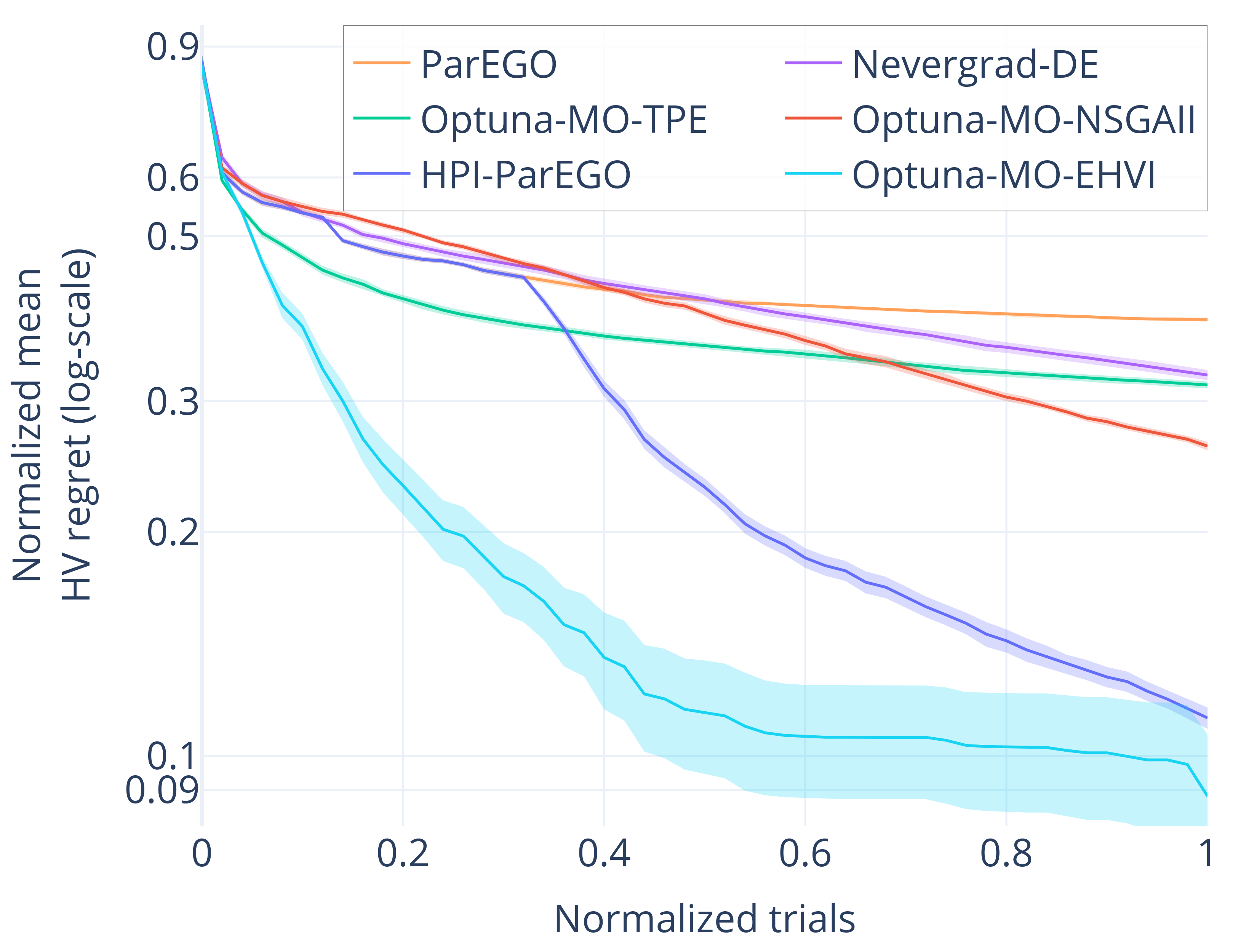}
          \caption{HV regret.}
    \end{subfigure}
    \caption{Results on the \pymoo tasks for the HPI-ParEGO optimizer compared to all baselines.}
    \label{fig:pymoo_all_baselines}
\end{figure*}

Figure~\ref{fig:pymoo_all_baselines} shows the results of HPI-ParEGO against all baselines on the \pymoo tasks.
EHVI performs best, likely because the simple, smooth ZDT problems allow its HV-based acquisition function to efficiently exploit the Pareto front by directly optimizing the HV performance measure and, for some tasks, the ZDT problems yield simple solutions (one objective at 0, the other at 1). 
As the \yahpo results show, this advantage does not translate to more complex real-world problems.
Following EHVI, MO-TPE leads initially, but as soon as the reduction begins, HPI-ParEGO improves drastically, almost reaching EHVI on HV regret in the end.

\newpage
\section{Scalability}\label{app:dim}
\begin{figure*}[h!]
    \centering
    \begin{subfigure}[h]{\textwidth}
      \centering
      \includegraphics[width=\columnwidth]{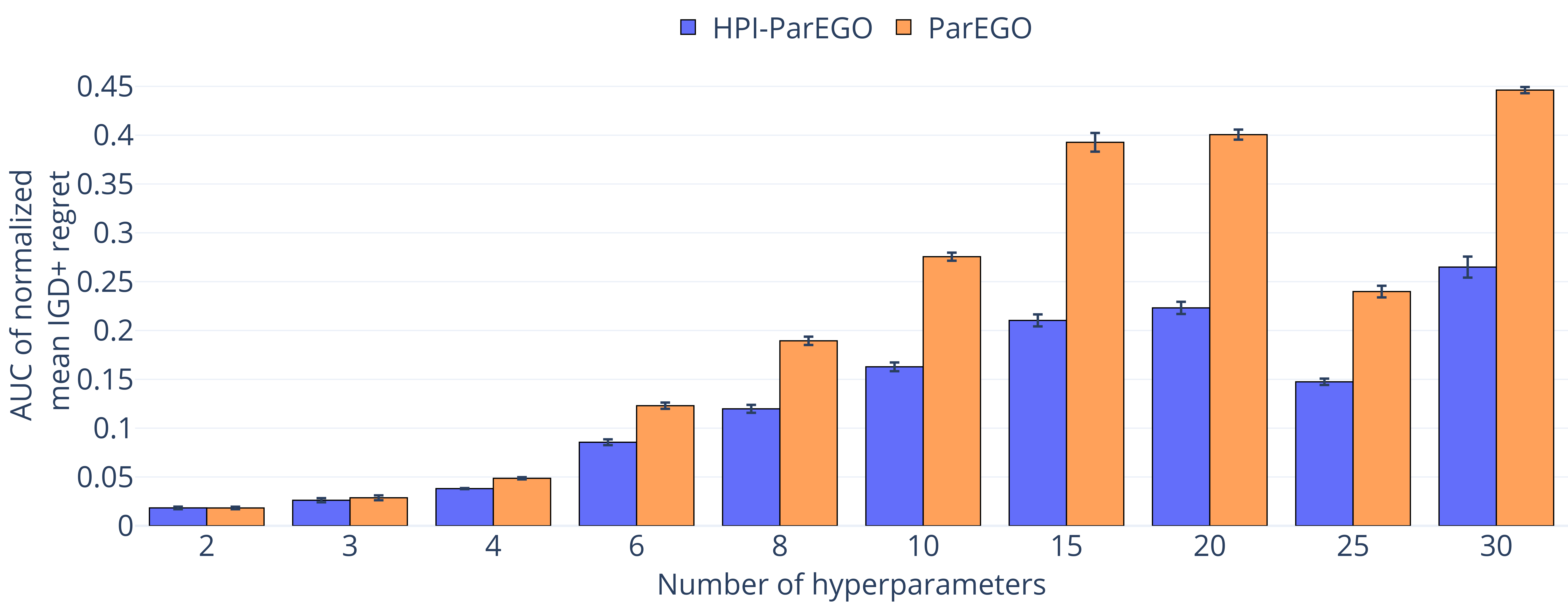}
      \caption{AUC of mean IGD+ regret.}
    \end{subfigure}%
    
    \begin{subfigure}[h]{\textwidth}
          \centering
          \includegraphics[width=\columnwidth]{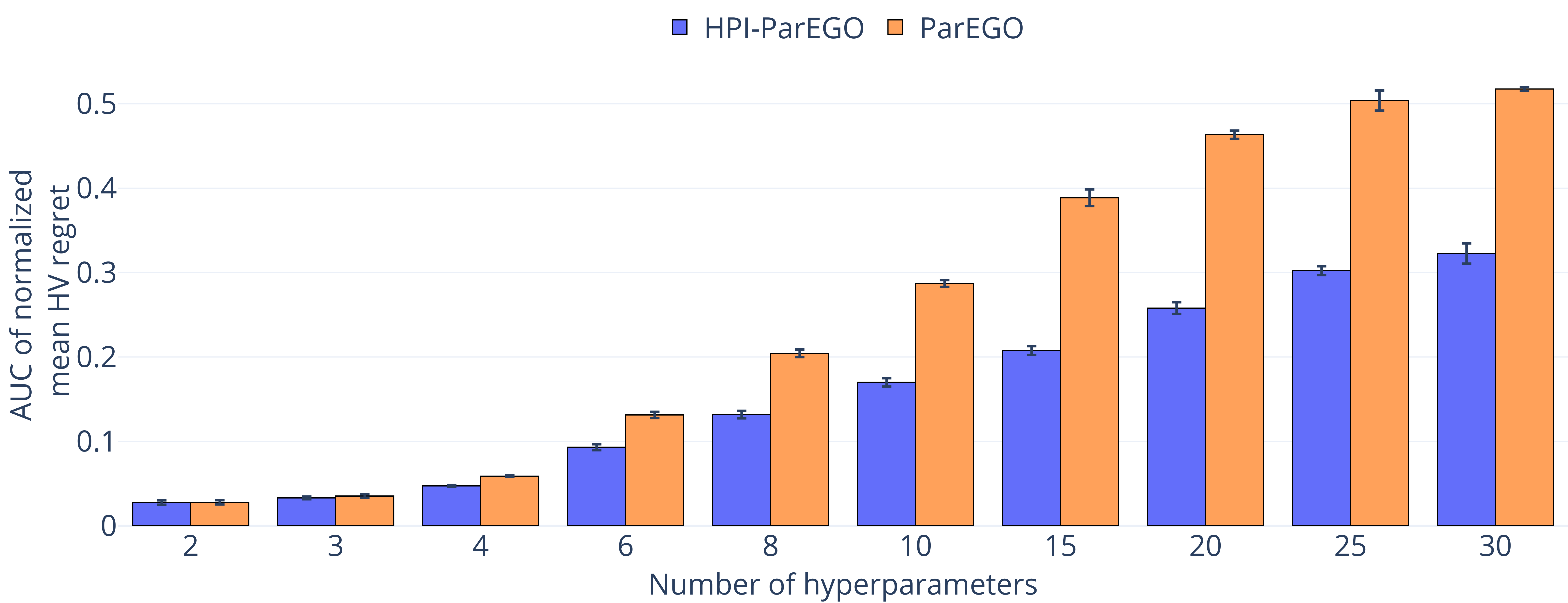}
          \caption{AUC of mean HV regret.}
    \end{subfigure}
    \caption{Results of HPI-ParEGO vs ParEGO for different dimensionalities on the \pymoo task \texttt{zdt2}.}
    \label{fig:dim}
\end{figure*}

In this experiment, we want to determine how our optimizer behaves across different dimensionalities.
We compare ParEGO against HPI-ParEGO on the \pymoo task \texttt{zdt2}, which has 30 dimensions. In this task, the first objective $f_1(\mathbf{\lambda}) = x_1$ depends solely on $x_1$, while the second objective depends on all other variables through a sum $g(\mathbf{\lambda}) = 1 + \frac{9}{(n-1)} \cdot \sum_{i=2}^{n} x_i$. Because each variable $x_2, \ldots, x_{30}$ contributes only a fraction to this sum, their individual influence is small relative to $x_1$, making \texttt{zdt2} a natural testbed for scalability under varying hyperparameter importance.
We repeat the experiment and include more irrelevant hyperparameters until we reach all 30 with this schedule: 2, 4, 6, 8, 10, 15, 20, 25, 30.
Figure~\ref{fig:dim} shows that for the first dimensions, the two optimizers perform similarly. As dimensionality scales up, the advantage of HPI-based selection becomes evident.

\newpage
\section{Additional Results for the Selected \yahpo Tasks}\label{app:task_diff}
\begin{figure*}[h!]
    \centering
    \begin{subfigure}[t]{0.5\textwidth}
      \includegraphics[width=\columnwidth]{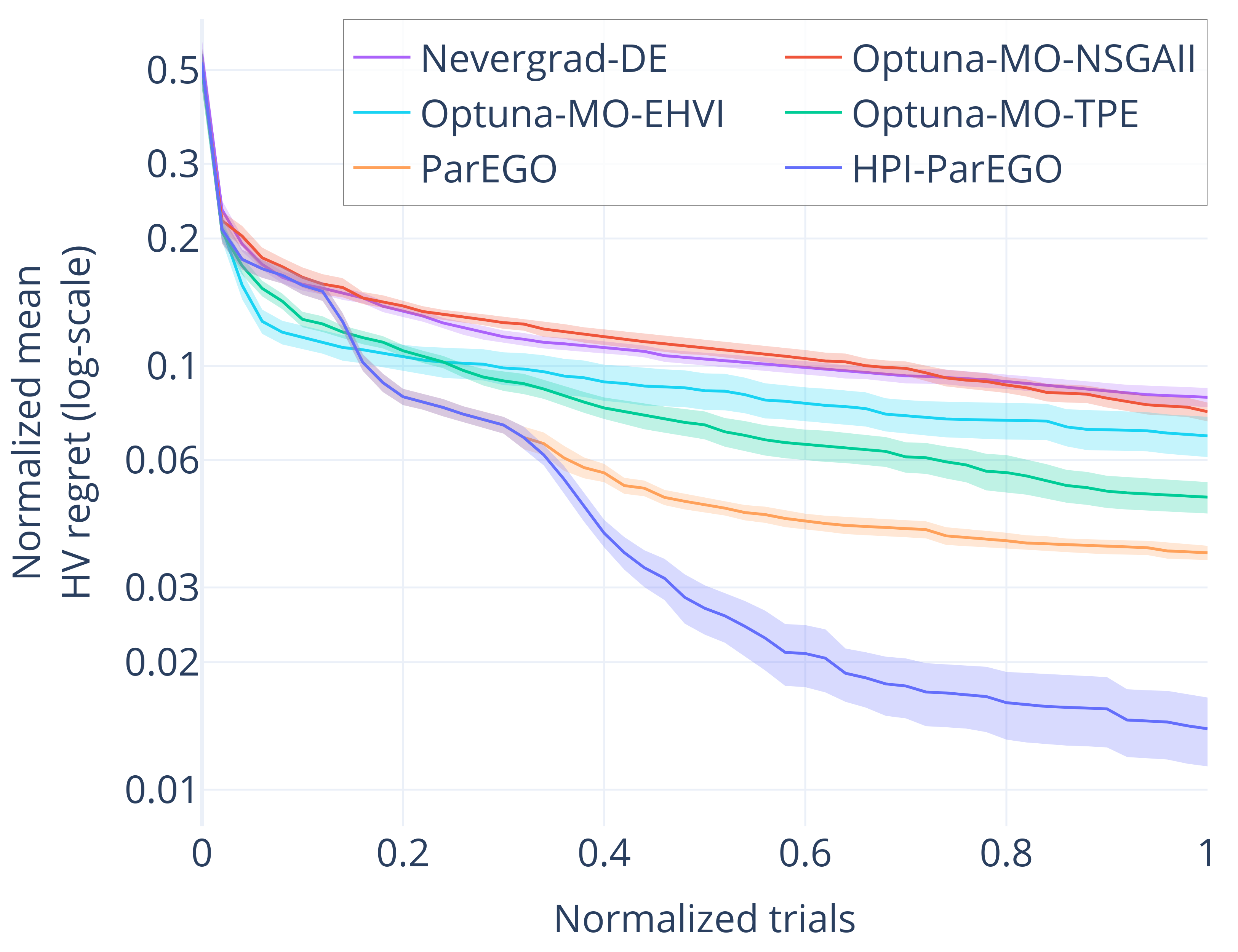}
      \caption{Results on the selected \texttt{LCBench} tasks.}
    \end{subfigure}%
    ~ 
    \begin{subfigure}[t]{0.5\textwidth}
      \centering
      \includegraphics[width=\columnwidth]{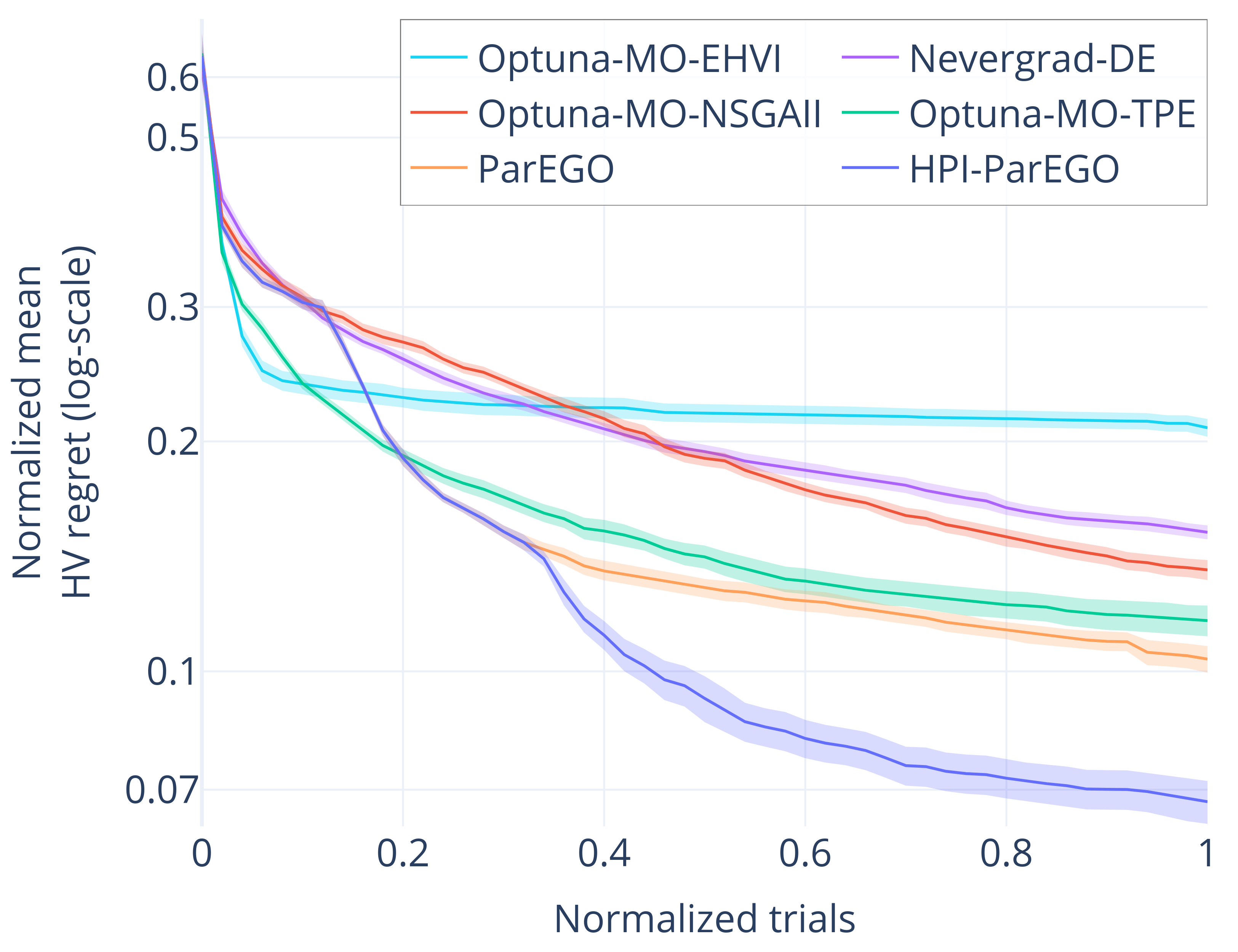}
      \caption{Results on the selected \texttt{rbv2\_ranger} tasks.}
    \end{subfigure}
    \caption{Results of HPI-ParEGO compared to the baseline optimizers measured in HV regret.}
\label{fig:yahpo_hv}
\end{figure*}

\begin{figure}[h]
    \centering
    \begin{subfigure}[h]{\textwidth}
    \includegraphics[width=\columnwidth]{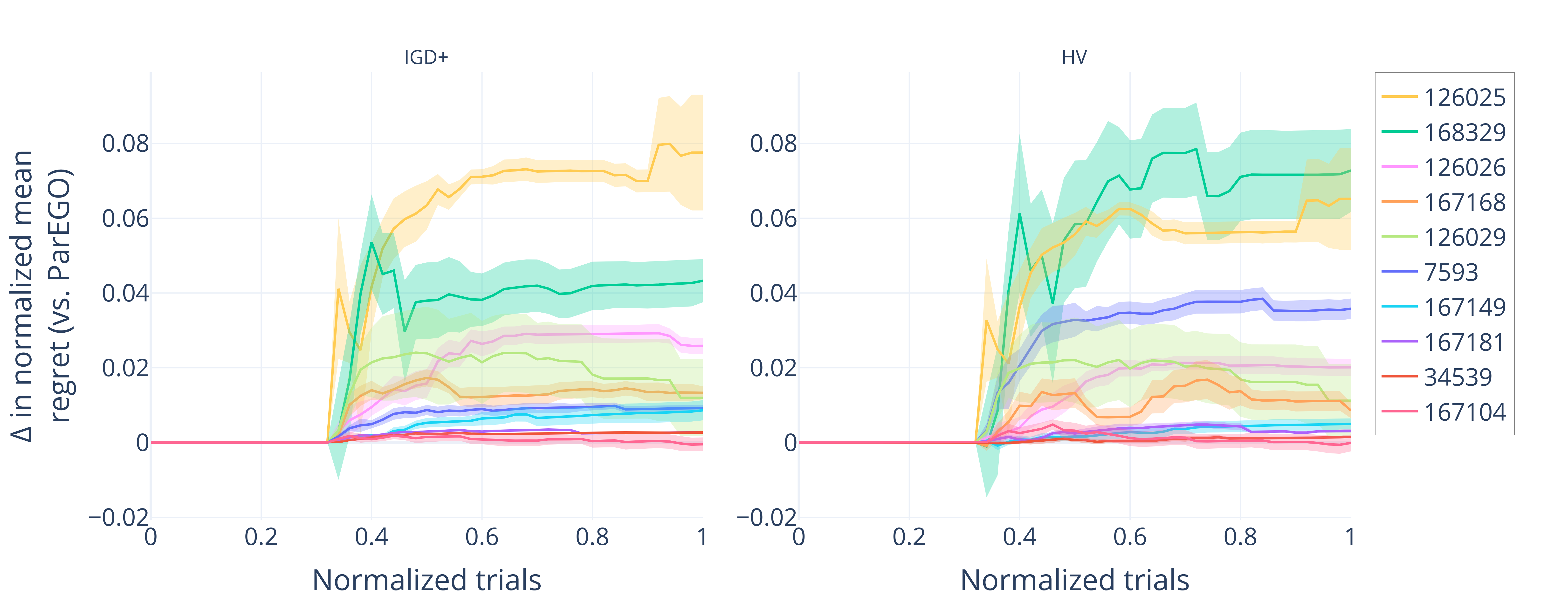}
    \caption{Difference in mean regret with standard error for the \texttt{LCBench} tasks.}
    \end{subfigure}%
    
    \begin{subfigure}[h]{\textwidth}
      \includegraphics[width=\columnwidth]{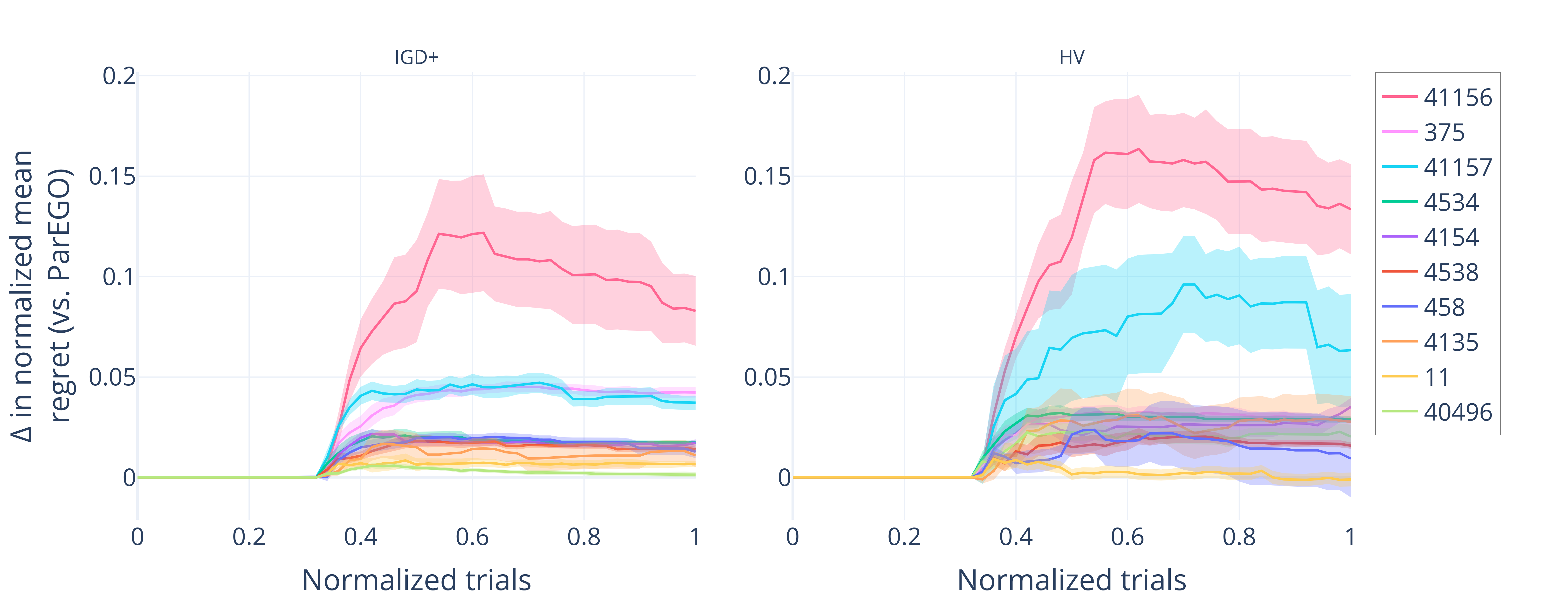}
      \caption{Difference in mean regret with standard error for the \texttt{rbv2\_ranger} tasks.}
    \end{subfigure}
    \caption{Difference between ParEGO and HPI-ParEGO on the selected \yahpo tasks. Values $> 0$ indicate a better performance of HPI-ParEGO.}
\label{fig:task_diff_yahpo}
\end{figure}

Figure~\ref{fig:yahpo_hv} shows the comparison of HPI-ParEGO against the baseline optimizers for the HV regret indicator.
The results are very similar to the IGD+ indicator.
Figure \ref{fig:task_diff_yahpo} compares HPI-ParEGO and the ParEGO baseline across the different scenarios from \yahpo for HV regret and IGD+ regret. The difference between the two metrics is minimal and only visible in a few tasks. The results confirm the benefits of incorporating HPI, as HPI-ParEGO consistently outperforms all baselines in both scenarios.

\newpage
\section{Results for the Randomly Selected \yahpo Tasks}\label{app:yahpo_random_tasks}
\begin{figure*}[h!]
    \centering
    \begin{subfigure}[t]{0.5\textwidth}
      \includegraphics[width=\columnwidth]{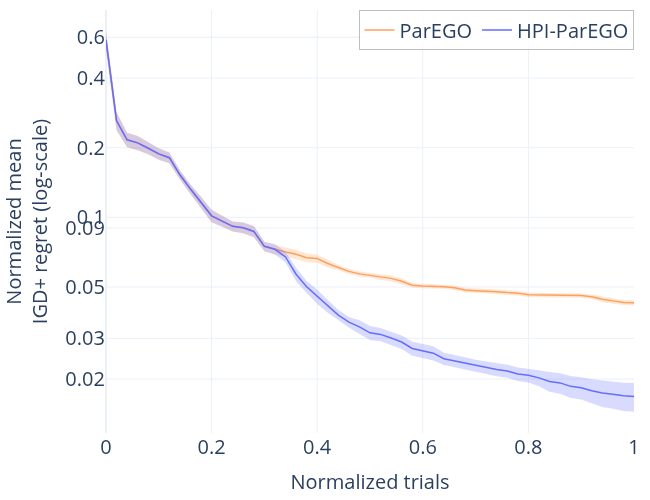}
      \caption{Results on the randomly selected \texttt{LCBench} tasks.}
    \end{subfigure}%
    ~ 
    \begin{subfigure}[t]{0.5\textwidth}
      \centering
      \includegraphics[width=\columnwidth]{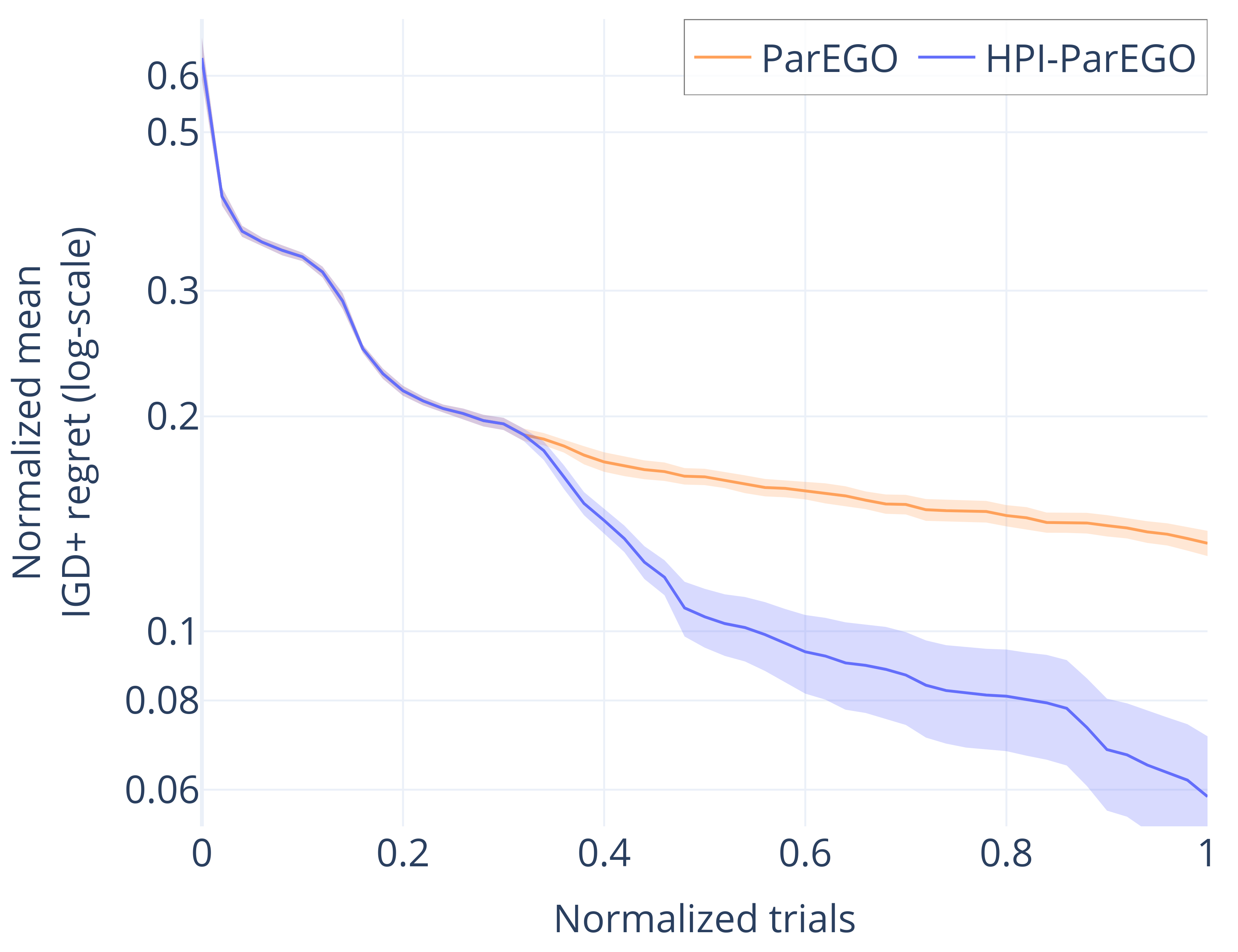}
      \caption{Results on the randomly selected \texttt{rbv2\_ranger} tasks.}
    \end{subfigure}
    \caption{Results of HPI-ParEGO compared to ParEGO measured in IGD+ regret.}
\label{fig:yahpo_rnd_igd}
\end{figure*}

\begin{figure*}[h!]
    \centering
    \begin{subfigure}[t]{0.5\textwidth}
      \includegraphics[width=\columnwidth]{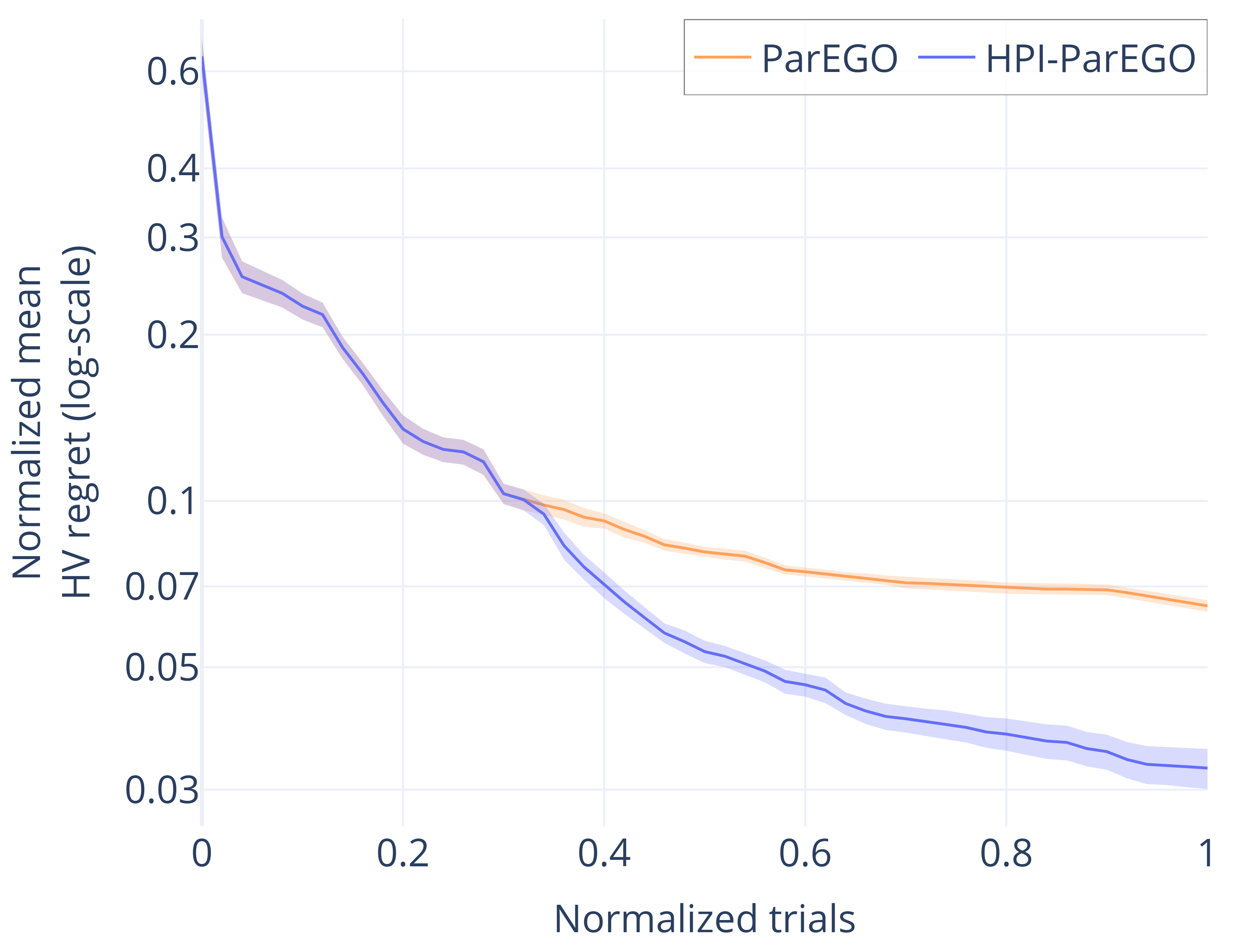}
      \caption{Results on the randomly selected \texttt{LCBench} tasks.}
    \end{subfigure}%
    ~ 
    \begin{subfigure}[t]{0.5\textwidth}
      \centering
      \includegraphics[width=\columnwidth]{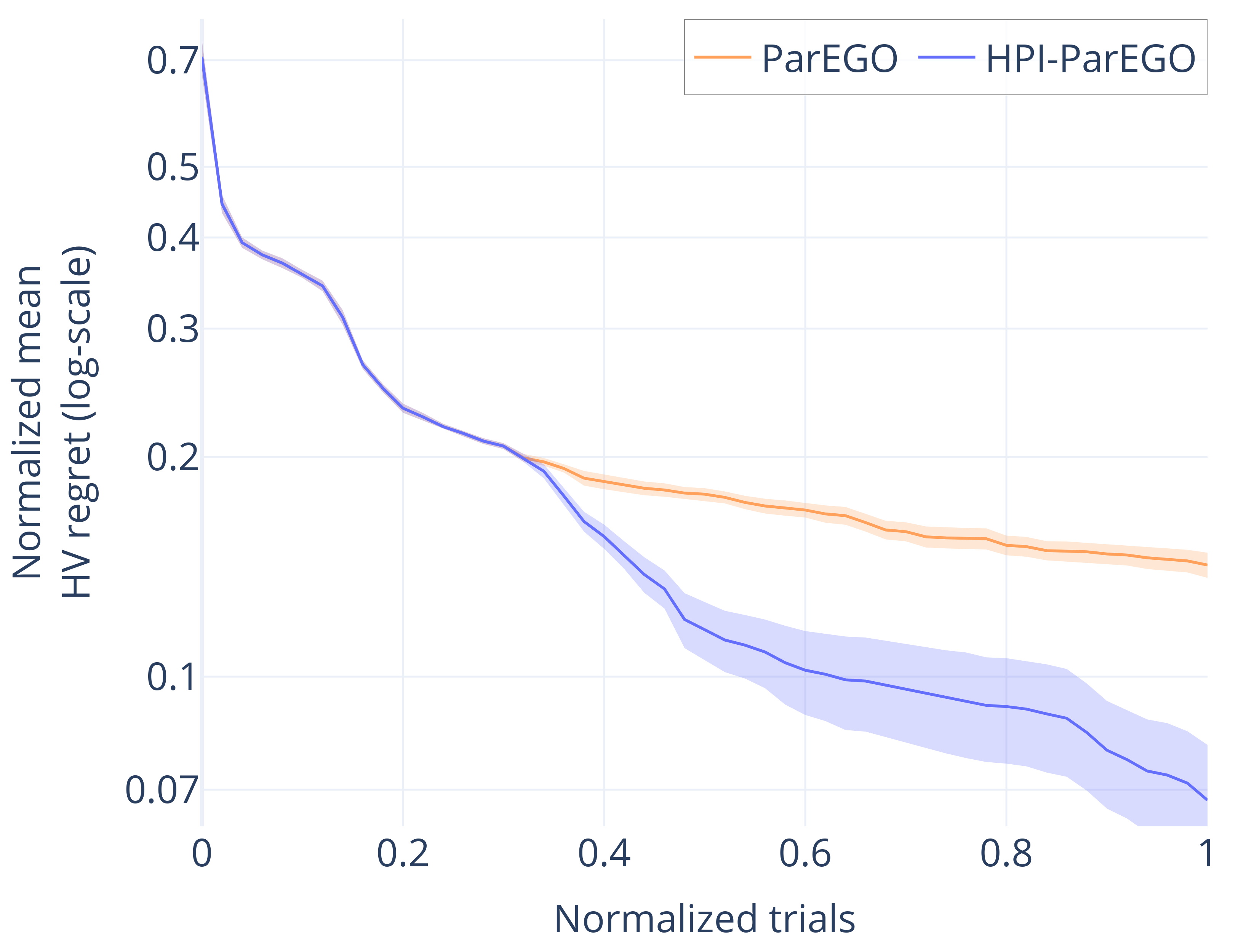}
      \caption{Results on the randomly selected \texttt{rbv2\_ranger} tasks.}
    \end{subfigure}
    \caption{Results of HPI-ParEGO compared to ParEGO measured in HV regret.}
\label{fig:yahpo_rnd_hv}
\end{figure*}

Since our task selection best reflects the effect we exploit in the optimization, we still want to show that HPI-ParEGO also works on a randomly selected set of tasks.
Figure~\ref{fig:yahpo_rnd_igd} shows the comparison of HPI-ParEGO against ParEGO on the 10 randomly selected \texttt{LCBench} tasks and the 10 randomly selected \texttt{rbv2\_ranger} tasks for the IGD+ indicator.
Figure \ref{fig:yahpo_rnd_hv} shows the same for HV regret.
The advantage of the HPI-ParEGO optimizer is also evident on these tasks, which are selected independently of the method.

\newpage
\section{Results for qNParEGO}\label{app:qnparego}
\begin{figure*}[h!]
    \centering
    \begin{subfigure}[t]{0.5\textwidth}
      \includegraphics[width=\columnwidth]{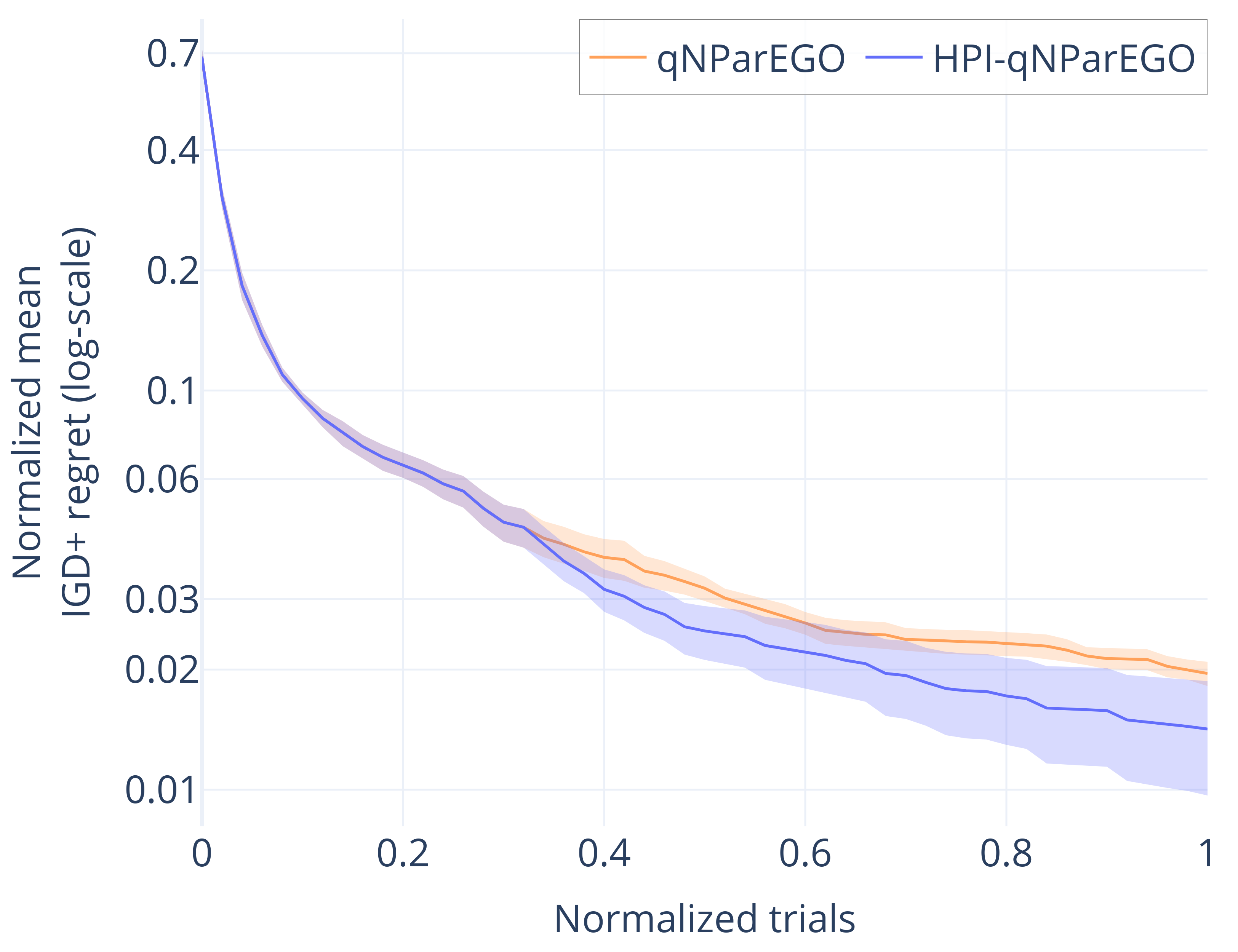}
      \caption{Results measured in IGD+ regret.}
    \end{subfigure}%
    ~ 
    \begin{subfigure}[t]{0.5\textwidth}
      \centering
      \includegraphics[width=\columnwidth]{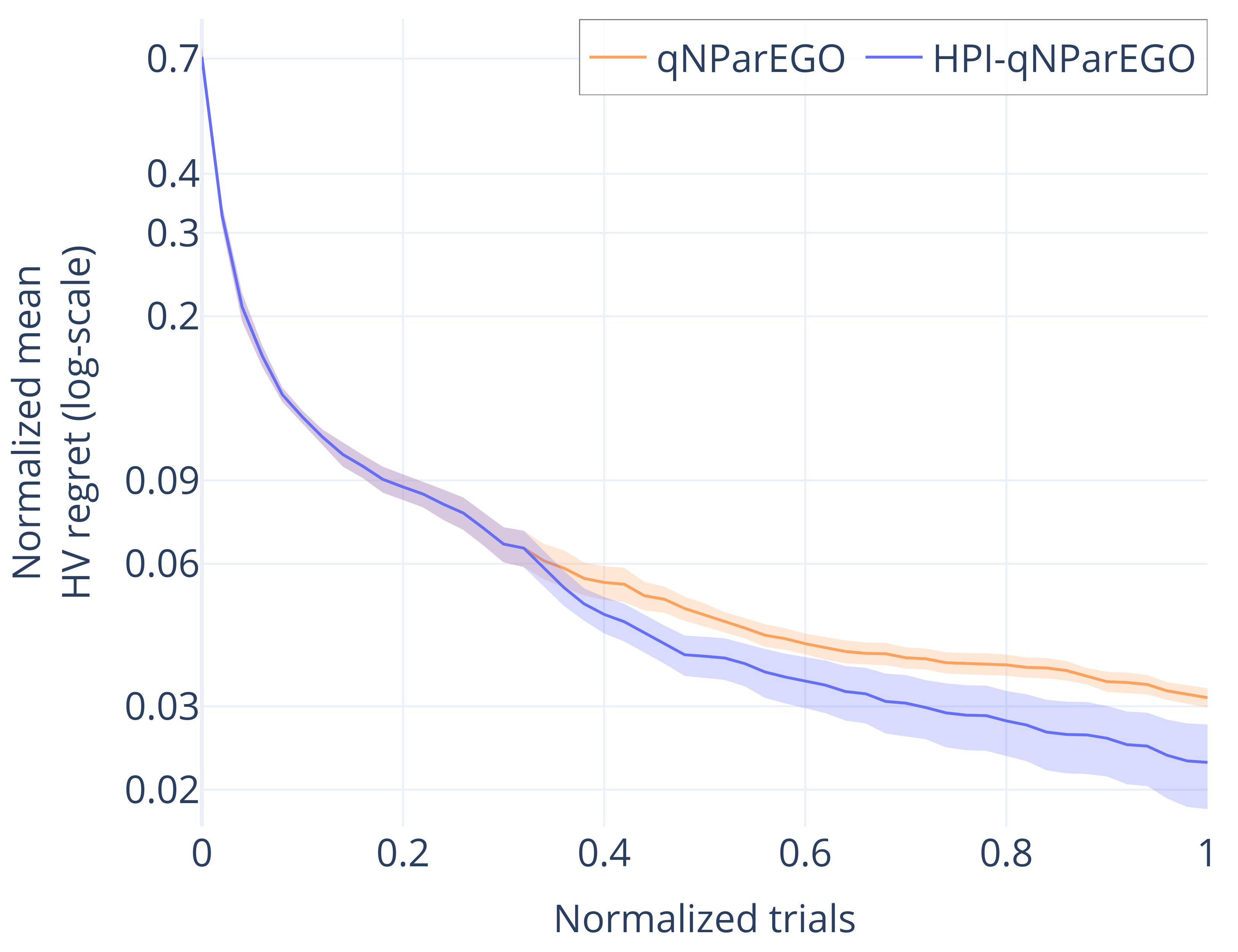}
      \caption{Results measured in HV regret.}
    \end{subfigure}
    \caption{Results of HPI-qNParEGO compared to qNParEGO.}
\label{fig:qnparego}
\end{figure*}

We apply our algorithm to qNParEGO~\citep{daulton-neurips20a} to demonstrate that it also works with another scalarization-based optimizer. For that, we use the BoTorch implementation~\citep{balandat-neurips20a} and set $q=2$. We use the same 10 selected \texttt{LCBench} tasks as in Section~\ref{sub:yahpo} with 10 seeds. Figure~\ref{fig:qnparego} shows that HPI-qNParEGO performs better than qNParEGO, albeit with a smaller gap than HPI-ParEGO to ParEGO.

\newpage
\section{Ground Truth Importance}\label{app:ground_truth}
\begin{figure}[h!]
    \centering
    \begin{subfigure}[h]{\textwidth}
        \includegraphics[width=\columnwidth]{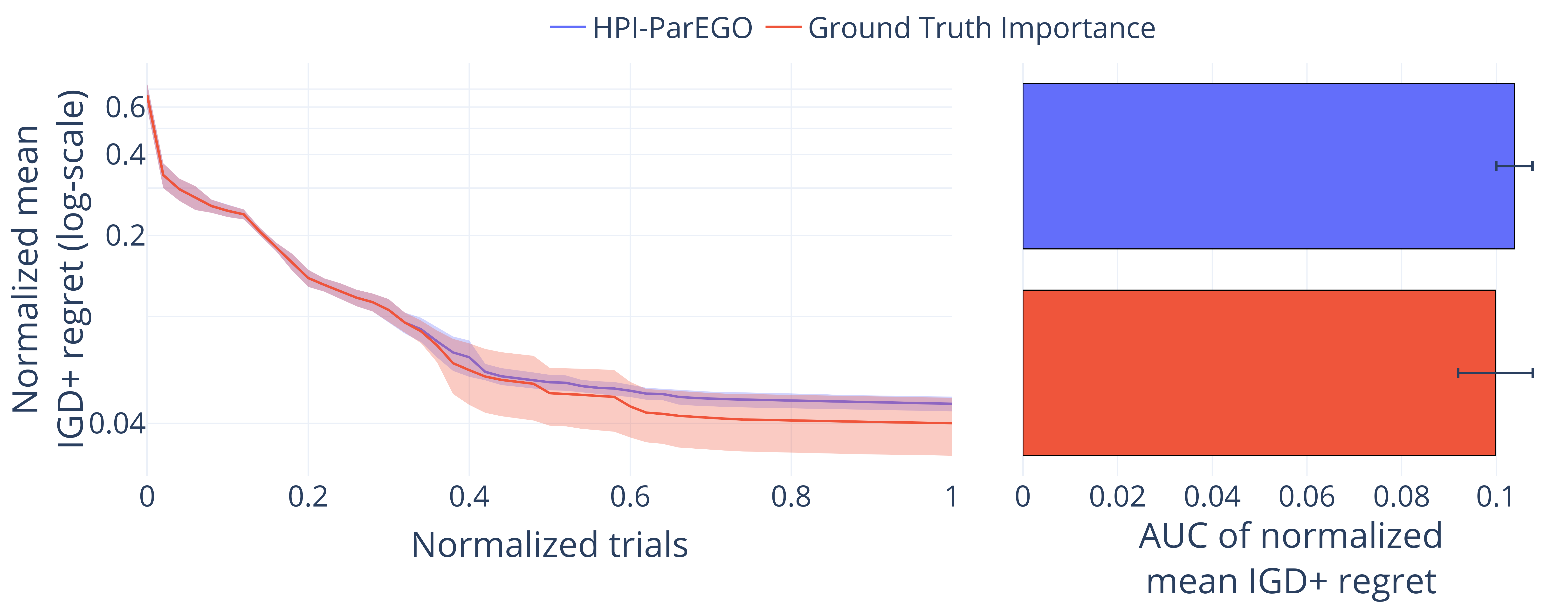}
        \caption{IGD+ regret.}
    \end{subfigure}%
    
    \begin{subfigure}[h]{\textwidth}
        \includegraphics[width=\columnwidth]{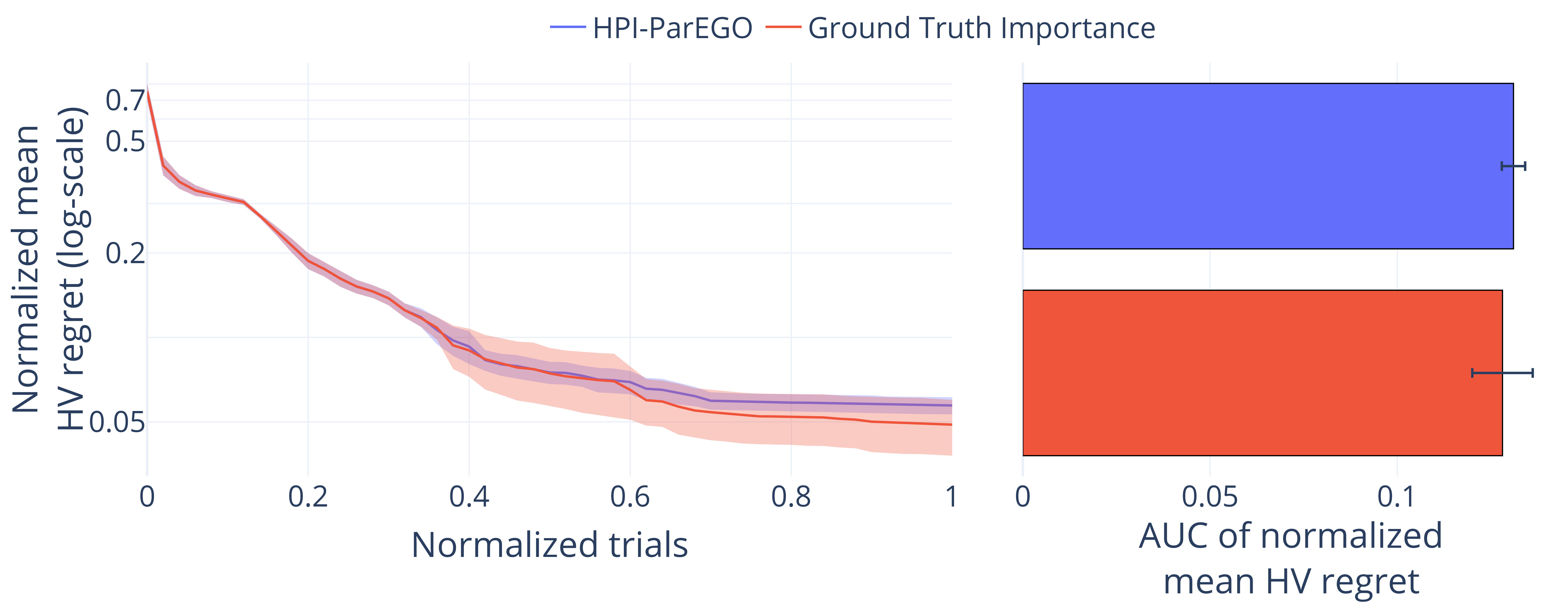}
        \caption{HV regret.}
    \end{subfigure}
    \caption{Results for HPI-ParEGO with ground truth importance against HPI-ParEGO.}
\label{fig:gt}
\end{figure}

Since HPI-ParEGO estimates HPI with a surrogate model that was trained on only a few hundred configurations, its estimates might be inaccurate. To assess this, we ran HPI-ParEGO on the ablation tasks with ground-truth importance, which we calculated by replacing the surrogate model used by HyperSHAP with the benchmark surrogate model, i.e., the YAHPO-Gym surrogate model, using the true objectives. We normalize them with the observed bounds and scalarize them with the current ParEGO weights. Figure~\ref{fig:gt} shows that HPI-ParEGO performs almost as well as the ground-truth variant. This indicates that the HPI estimation is very good, even with only a few hundred configurations.

\end{document}